\documentclass{article}

\PassOptionsToPackage{numbers,compress}{natbib}
\usepackage[eandd, final]{neurips_2026}

\usepackage{algorithm}
\usepackage{algpseudocode}
\usepackage[utf8]{inputenc} 
\usepackage[T1]{fontenc}    
\usepackage{hyperref}       
\usepackage{url}            
\usepackage{booktabs}       
\usepackage{amsfonts}       
\usepackage{nicefrac}       
\usepackage{microtype}      
\usepackage{xcolor}         
\usepackage{listings}       

\usepackage{tabularx}
\usepackage{placeins}
\usepackage{ragged2e}
\usepackage{paralist} 
\usepackage{csquotes} 
\usepackage{graphicx} 
\usepackage{adjustbox}
\usepackage{multirow}
\usepackage{array}
\usepackage{makecell}
\usepackage{siunitx}
\graphicspath{{figures/}} 
\usepackage{amsmath, amssymb, amsthm} 
\usepackage{doi}
\newcommand{\R}{\mathbb{R}}
\newcommand{\Nbb}{\mathbb{N}}

\newcommand{\x}{\textbf{x}}

\newtheorem{definition}{Definition}

\definecolor{repoCodeBg}{HTML}{F7FAFC}
\definecolor{repoCodeFrame}{HTML}{B7C7D8}
\definecolor{repoCodeCommand}{HTML}{005CC5}
\definecolor{repoCodeScript}{HTML}{6F42C1}
\definecolor{repoCodeOption}{HTML}{B31D28}
\definecolor{repoCodeValue}{HTML}{22863A}
\definecolor{repoCodeComment}{HTML}{6A737D}
\lstdefinestyle{repoShell}{
  language=bash,
  basicstyle=\ttfamily\small\setlength{\baselineskip}{1.15\baselineskip},
  backgroundcolor=\color{repoCodeBg},
  frame=lines,
  framesep=4pt,
  framerule=0.3pt,
  rulecolor=\color{repoCodeFrame},
  columns=fullflexible,
  keepspaces=true,
  aboveskip=0.6em,
  belowskip=0.6em,
  showstringspaces=false,
  breaklines=true,
  breakatwhitespace=true,
  alsoletter={-_.},
  keywordstyle=\color{repoCodeCommand}\bfseries,
  commentstyle=\color{repoCodeComment}\itshape,
  stringstyle=\color{repoCodeValue},
  morekeywords={uv,python},
  emph=[1]{run_training.py,run_testing.py,run_analysis.py},
  emphstyle=[1]{\color{repoCodeScript}},
  emph=[2]{--data-files,--model,--method,--max-epochs,--max-hp-trials-per-model,--logdir,--perturbation-scenarios,--full-coverage},
  emphstyle=[2]{\color{repoCodeOption}},
  emph=[3]{ETTh1,DLinear,baseline,missing_data,noise,runs,sync,run},
  emphstyle=[3]{\color{repoCodeValue}},
  xleftmargin=0pt,
  xrightmargin=0pt
}

\newcommand{\DatasetLineBreak}[2]{\begin{tabular}[t]{@{}l@{}}#1\\#2\end{tabular}}

\newcommand{\DatasetLineBreakTinyCentered}[2]{\begin{tabular}[c]{@{}c@{}}\tiny #1\\[-0.35ex]\tiny #2\end{tabular}}
\newcommand{\BeijingTiantan}{Beijing Air Tiantan}
\newcommand{\BeijingTiantanShort}{\DatasetLineBreak{Beijing Air}{Tiantan}}

\newcommand{\BeijingTiantanTinyCentered}{\DatasetLineBreakTinyCentered{Beijing Air}{Tiantan}}
\newcommand{\PenmanshielWT}{Penmanshiel WT08}
\newcommand{\PenmanshielWTShort}{\DatasetLineBreak{Penmanshiel}{WT08}}

\newcommand{\PenmanshielWTTinyCentered}{\DatasetLineBreakTinyCentered{Penmanshiel}{WT08}}
\newcommand{\SensorFaultBenchRepo}{\url{https://github.com/awindmann/SensorFault-Bench}}
\newcommand{\BeijingTiantanZenodo}{\url{https://doi.org/10.5281/zenodo.23797045}}
\newcommand{\PenmanshielZenodo}{\url{https://doi.org/10.5281/zenodo.23797057}}

\title{Benchmarking Sensor-Fault Robustness in Forecasting}

\author{%
  \textbf{Alexander Windmann$^{1}$ \quad
  Philipp Wittenberg$^{1}$ \quad
  Gianluca Manca$^{2}$} \\
  \textbf{Marcel Dix$^{3}$ \quad
  Jens U. Brandt$^{4,5}$ \quad
  Oliver Niggemann$^{1}$} \\
  {\small
  $^{1}$Helmut Schmidt University \hspace{0.35em}
  $^{2}$Ruhr University Bochum \hspace{0.35em}
  $^{3}$ABB Corporate Research}\\
  {\small
  $^{4}$TH Köln \hspace{0.35em}
  $^{5}$Leiden University}
}

\begin{document}

\maketitle

\begin{abstract}
  Cyber-physical system (CPS) forecasting models depend on sensor streams with noisy, biased, missing, or temporally misaligned readings, yet standard forecasting evaluation often selects models by nominal error without showing whether they remain robust under such faults.
  We introduce SensorFault-Bench, a CPS-grounded protocol that evaluates forecasting architectures and robustness-improvement methods through controlled synthetic stress tests of isolated input faults, together with an operational taxonomy of the methods.
  Across four real-world datasets and eight scored scenarios governed by a standardized severity model, it reports worst-scenario degradation, clean mean squared error (MSE), and worst-scenario fault-time MSE, separating relative robustness from absolute error.
  A disjoint fault-transfer split lets explicit fault-training methods train on adjacent fault families while reported scores use separate benchmark scenarios.
  Empirically, forecasting architectures favored by clean MSE can degrade sharply under faults, and clean-MSE rankings can disagree with worst-scenario fault-time error rankings.
  Chronos-2, the evaluated zero-shot foundation-model representative, matches or trails the last-value naive forecaster in clean MSE on the two single-target datasets and has the largest worst-scenario degradation on ETTh1 and Traffic, where all channels are forecast targets.
  For the evaluated robustness-improvement method set, paired deltas show selective degradation reductions: projected gradient descent adversarial training and randomized training lead where value faults dominate observed degradation, and fault augmentation leads on Traffic, where availability faults dominate.
  High degradation reveals fault sensitivity that clean MSE misses, while low degradation supports robustness only within the evaluated scenarios.
  The benchmark therefore screens forecasters before use in a specific deployment, which requires validation against its recorded faults.
  SensorFault-Bench provides open-source code, documented data access, and reproduction and extension guides, so new datasets, architectures, and robustness-improvement methods can be evaluated under the same protocol.
\end{abstract}

\begin{center}
\small
\textbf{Code and benchmark repository:} \SensorFaultBenchRepo
\end{center}

\section{Introduction}
In cyber-physical systems (CPS), forecasting models are deployed on sensor streams whose failures are technical faults of the measurement pipeline, not only statistical nuisances.
Sensor wear, communication outages, maintenance events, synchronization failures, and preprocessing mistakes can corrupt the history seen by the forecaster without changing the underlying physical process or the forecasting target.
A forecaster that looks strong under nominal evaluation can therefore become operationally unreliable once the sensing pipeline degrades.
Yet standard forecasting evaluation still emphasizes held-out error under nominal conditions, split design, and metric choice \citep{hewamalageForecastEvaluationData2023,songDeepLearningbasedTime2024}.

These failures are not rare edge cases.
Sensor time series are routinely affected by missing values, outliers, noise, and drift \citep{tehSensorDataQuality2020,jesusSurveyDataQuality2017}, arising from sensor wear, maintenance events, synchronization failures, communication problems, or preprocessing mistakes.
Many of these disturbances are structured rather than isolated: a channel can stay biased, remain stuck at its last valid reading, or lose local temporal alignment while still looking plausible.
For industrial and safety-relevant settings, low nominal error alone is therefore insufficient: forecasters must degrade gracefully under documented fault modes that practitioners diagnose, repair, or tolerate.

Forecasting robustness work has addressed parts of this problem, but mostly in different settings.
One line studies adversarial perturbations in forecasting \citep{liu2023robust,yoonRobustProbabilisticTime2022}, while another studies learning from anomaly-contaminated training data \citep{chengRobustTSFTheoryDesign2024}.
Recent forecasting testbeds define controlled stress tests for spike and level-shift corruptions \citep{kimLocalGeometryAttention2026}, unavailable variables at inference \citep{chauhanMultiVariateTimeSeries2022}, or synthetic signal and noise processes \citep{janssenBenchmarkingMLTSFFrequency2026}.
Severity-controlled CPS forecasting benchmarks already evaluate architectures on real CPS datasets with formal robustness scores \citep{windmannQuantifyingRobustnessBenchmarking2025}, and industrial time-series work evaluates gradual data-quality perturbations for classification \citep{Dix.etal_2023}.
Together, these studies cover stress tests, architecture comparisons, and industrial data quality, but none defines a shared forecasting protocol with a suite of sensor faults, paired comparisons of robustness-improvement methods, disjoint transfer faults, and separate reporting of clean error, degradation, and fault-time error.

\begin{figure*}[t]
  \centering
\includegraphics[
  width=0.99\textwidth, trim=0 0 1.2em 0, clip]{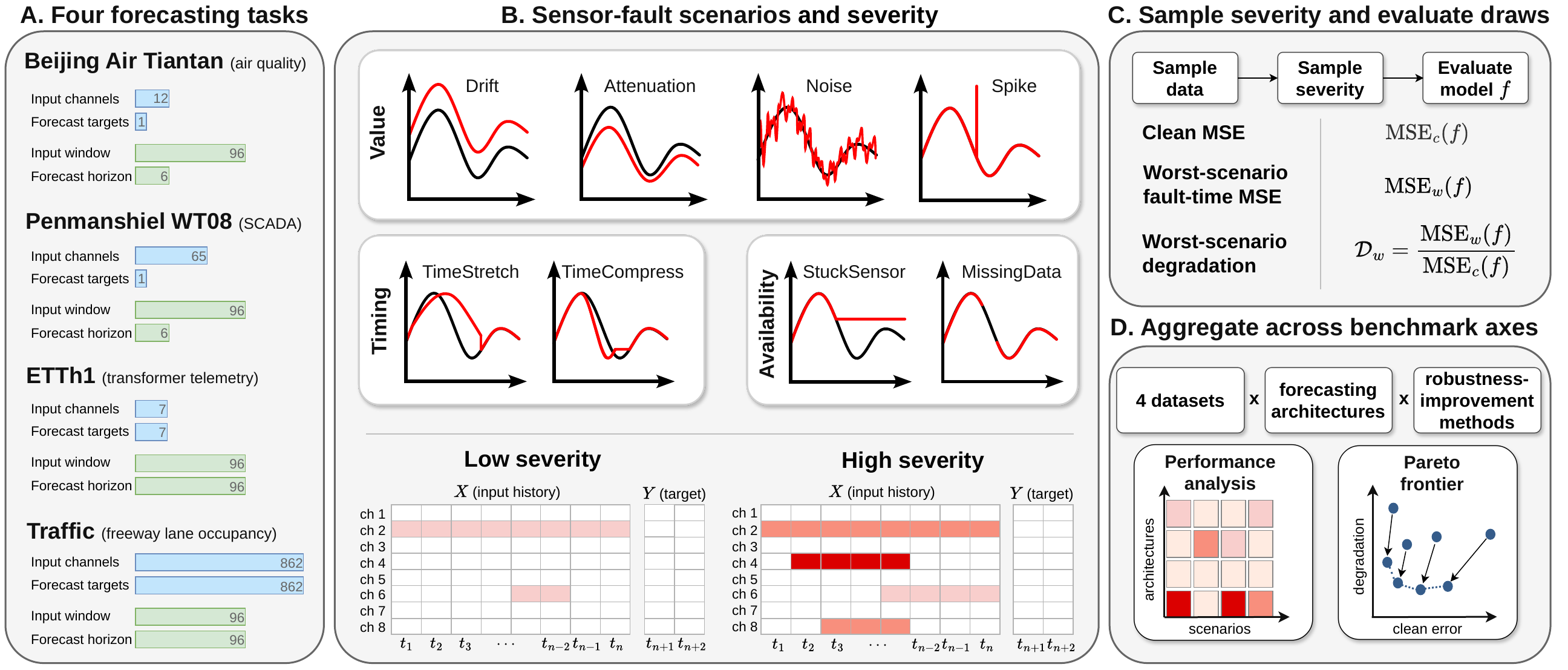}
  \caption{Benchmark overview for SensorFault-Bench.
  The benchmark perturbs observed input histories while keeping forecast targets fixed.
  The scored set $\mathcal{P}$ contains eight value, timing, and availability scenarios over one severity axis.
  Evaluation estimates degradation per scenario and reports worst-scenario degradation with clean MSE and worst-scenario fault-time MSE.
  Panels A--C show task scope, scenario suite, and sampling-to-score flow.
  Panel D compares clean error with degradation after aggregating results across datasets, architectures, robustness-improvement methods, and scenarios.}
  \label{fig:scenarios}
\end{figure*}

We address that gap with SensorFault-Bench, a benchmark for robustness evaluation in forecasting under structured sensor faults.
SensorFault-Bench turns recurring sensing failures documented in studies of sensor and industrial data quality into eight scored sensor-fault scenarios \citep{hubauerAnalysisDataQuality2013,jesusSurveyDataQuality2017,tehSensorDataQuality2020}.
These studies motivate the value, timing, and availability families shown in Figure~\ref{fig:scenarios}.
The shared severity model fixes common stress-test levels for cross-dataset comparison, leaving deployment frequencies and operating limits to system-specific calibration.
Across four real-world forecasting tasks, including two curated datasets derived from larger source archives, the benchmark evaluates those eight scenarios under the shared severity model.
It reports worst-scenario degradation with clean mean squared error (MSE) and worst-scenario fault-time MSE, keeping relative robustness and absolute forecasting error distinct.
It also defines $\mathcal{P}_{\mathrm{trans}}$, seven training-only transfer fault families adjacent to but disjoint from the scored set $\mathcal{P}$.
This design separates two roles that single corrupted-error summaries can conflate: all models are scored on $\mathcal{P}$, while explicit fault-training methods are checked for transfer from $\mathcal{P}_{\mathrm{trans}}$ rather than by replaying scored scenarios.

Using this framework, we compare forecasting architectures and robustness-improvement methods spanning stochastic and adversarial perturbation training, robust objectives, adaptive normalization, ensemble redundancy, and post-hoc randomized smoothing.
Pairing each method with the baseline forecasting model of the same architecture turns claims about method robustness into comparisons on the same datasets and scored fault scenarios.
The operational taxonomy organizes these benchmarked families by stage of intervention, required model access, and target of intervention.
Appendix~\ref{sec:taxonomy} situates the evaluated method set within a literature-grounded coverage and extension map.
The study supports screening forecasters before deployment under CPS sensor faults, with relative degradation, clean error, absolute fault-time error, scenario families, and transfer behavior each answering a distinct question.

Our contributions are fourfold: (i) a four-dataset sensor-fault benchmark for forecasting whose scored scenario families are grounded in recurring CPS sensing failures and defined as standardized input-history stress tests, (ii) a disjoint fault-transfer split that evaluates explicit fault-training methods on scored scenarios disjoint from their training faults, (iii) a three-quantity reporting protocol that pairs worst-scenario degradation with clean MSE and worst-scenario fault-time MSE and evaluates robustness-improvement methods through method-baseline deltas in the same quantities, and (iv) a comparative study showing structured trade-offs across architectures, scored fault scenarios, and the evaluated method set.

\section{Related work}
\label{sec:related_work}
\begin{table*}[t]
  \centering
  \small
  \setlength{\tabcolsep}{1.0pt}
  \renewcommand{\arraystretch}{0.95}
  \newcommand{\poshdr}[2]{\begin{tabular}[b]{@{}c@{}}#1\\#2\end{tabular}}
  \newcommand{\worktype}[2]{\multirow[t]{#1}{=}{\begin{tabular}[t]{@{}l@{}}#2\end{tabular}}}
  \caption{Selected related work by protocol properties relative to SensorFault-Bench.
  Symbols $\checkmark$, $\circ$, and -- denote central, narrower or adjacent, and absent or incidental coverage within each work's own task and stress setting, not results re-scored under this benchmark.}
  \vspace{7pt}
  \label{tab:benchmark_positioning}
  \begin{tabularx}{\textwidth}{@{}>{\RaggedRight\arraybackslash}p{0.0925\textwidth}>{\RaggedRight\arraybackslash}p{0.3325\textwidth}*{7}{>{\centering\arraybackslash}X}@{}}
    \toprule
	    Category &
	    Source &
	    \poshdr{Forecast}{task} &
	    \poshdr{CPS}{focus} &
	    \poshdr{Fault}{scope} &
	    \poshdr{Model}{scope} &
	    \poshdr{Method}{scope} &
	    \poshdr{Transfer}{faults} &
	    \poshdr{Measure}{split} \\
    \midrule
    Nominal &
    Monash Forecasting Repository \citep{godahewaMonashTimeSeries2021} &
    $\checkmark$ & $\circ$ & -- & $\checkmark$ & -- & -- & -- \\
    &
    TFB \citep{qiuTFBComprehensiveFair2024} &
    $\checkmark$ & $\circ$ & -- & $\checkmark$ & -- & -- & -- \\
      \addlinespace[3pt]
	    \worktype{4}{Method\\paper} &
	    \citet{liu2023robust} &
	    $\checkmark$ & $\circ$ & $\circ$ & $\circ$ & $\circ$ & -- & $\circ$ \\
	    &
	    \citet{yoonRobustProbabilisticTime2022} &
	    $\checkmark$ & $\circ$ & $\circ$ & -- & $\circ$ & -- & $\circ$ \\
	    &
	    RobustTSF \citep{chengRobustTSFTheoryDesign2024} &
	    $\checkmark$ & $\circ$ & $\circ$ & $\circ$ & $\circ$ & -- & $\circ$ \\
	    &
	    \citet{brandtFaultsFeaturesPretraining2025} &
	    $\circ$ & $\checkmark$ & $\checkmark$ & $\circ$ & $\circ$ & $\circ$ & $\circ$ \\
 \addlinespace[3pt]
    \worktype{3}{Forecast\\stress} &
    TSRBench \citep{kimLocalGeometryAttention2026} &
    $\checkmark$ & $\circ$ & $\circ$ & $\checkmark$ & $\circ$ & -- & $\circ$ \\
    &
    Variable Subset Forecast \citep{chauhanMultiVariateTimeSeries2022} &
    $\checkmark$ & $\circ$ & $\circ$ & $\checkmark$ & $\checkmark$ & -- & $\checkmark$ \\
	    &
	    \citet{janssenBenchmarkingMLTSFFrequency2026} &
	    $\checkmark$ & -- & $\circ$ & $\checkmark$ & -- & -- & $\circ$ \\
 \addlinespace[3pt]
    Industrial &
    \citet{Dix.etal_2023} &
    -- & $\checkmark$ & $\checkmark$ & $\checkmark$ & -- & -- & $\checkmark$ \\
    &
    \citet{windmannQuantifyingRobustnessBenchmarking2025} &
    $\checkmark$ & $\checkmark$ & $\checkmark$ & $\checkmark$ & -- & -- & $\circ$ \\
 \addlinespace[3pt]
    This\,work &
    SensorFault-Bench &
    $\checkmark$ & $\checkmark$ & $\checkmark$ & $\checkmark$ & $\checkmark$ & $\checkmark$ & $\checkmark$ \\
    \bottomrule
  \end{tabularx}
  \vspace{-6pt}
\end{table*}
Nominal forecasting benchmarks make model comparison reusable: the Monash Forecasting Repository provides a multi-dataset forecasting archive with baseline results \citep{godahewaMonashTimeSeries2021}, and TFB standardizes comparison across dataset domains, method families, evaluation strategies, and metrics \citep{qiuTFBComprehensiveFair2024}.
Robust forecasting and stress-test studies cover adjacent settings: adversarial input perturbations \citep{liu2023robust,yoonRobustProbabilisticTime2022}, training series contaminated by anomalies \citep{chengRobustTSFTheoryDesign2024}, spikes and level shifts \citep{kimLocalGeometryAttention2026}, unavailable variables at inference \citep{chauhanMultiVariateTimeSeries2022}, and synthetic signal and noise processes \citep{janssenBenchmarkingMLTSFFrequency2026}.
Industrial and CPS studies include gradual perturbation stress tests for industrial time series classification \citep{Dix.etal_2023}, CPS forecasting benchmarks with controlled severity \citep{windmannQuantifyingRobustnessBenchmarking2025}, and representation pretraining aimed at sensor failures in vehicle dynamics prediction and control \citep{brandtFaultsFeaturesPretraining2025}.
Together, these lines of work motivate the comparison dimensions, but no single work covers the full setting evaluated here: forecasting under CPS sensor faults with a broad comparison of architectures, paired evaluation of robustness-improvement methods, disjoint transfer faults, and separate measures of clean error, error under stress, and degradation.

Table~\ref{tab:benchmark_positioning} tracks this combined setting across selected related work.
The CPS focus column records whether CPS or industrial sensor streams are central to a work, not the breadth of its datasets or domains.
For fault scope, checks require a central suite with several families of sensor, data quality, or CPS faults, whereas circles indicate narrower fault families or related stress settings.
For method scope, checks require method-baseline comparisons under a shared stress setting across at least two base architectures or families, whereas circles indicate narrower comparisons of a proposed method, wrapper, preprocessing step, or adaptation, such as those on a single model.
For measure split, checks require separate quantities for clean or reference conditions, stress conditions, and degradation or sensitivity, whereas circles indicate two of these categories.

\section{Benchmark overview}
\noindent SensorFault-Bench evaluates four real-world forecasting tasks under the protocol defined in Section~\ref{sec:sensor_fault_protocol}, which injects sensor faults into model inputs.
Each selected model is tested on the same eight value, timing, and availability scenarios, with severity sampled uniformly over $[0,1]$ within each scenario.
Explicit fault-training methods may draw from the disjoint training-only transfer set $\mathcal{P}_{\mathrm{trans}}$ (Appendix~\ref{sec:test_scenarios}), but all reported scores use only the scenario set $\mathcal{P}$.
Forecasting architectures and robustness-improvement methods are summarized by worst-scenario degradation, clean MSE, and worst-scenario fault-time MSE.

\subsection{Datasets and forecasting settings}

The four tasks cover urban air quality, wind turbine supervisory control and data acquisition (SCADA), electricity transformer telemetry, and freeway lane occupancy.
We curate \BeijingTiantan{} from the UCI Beijing Multi-Site Air Quality archive \citep{Chen_2017} and \PenmanshielWT{} from one turbine in the SCADA release of the Penmanshiel wind farm \citep{Plumley.Takeuchi_2025}.
ETTh1 and Traffic are public forecasting benchmarks: ETTh1 is an Electricity Transformer Temperature dataset \citep{zhouInformerEfficientTransformer2021}, while Traffic contains 862 hourly road-occupancy series measured by freeway sensors in the San Francisco Bay Area and drawn from California Department of Transportation data \citep{Lai.etal_2018}.
All models receive the full multivariate history of 96 input hours.
\BeijingTiantan{} and \PenmanshielWT{} forecast one target over a 6-hour horizon, and ETTh1 and Traffic forecast all channels over a 96-hour horizon.
This split is descriptive because horizon length, target count, channel count, and affected-channel scope shift together.
Appendix~\ref{sec:dataset_asset_derivation} gives derivation details for the curated datasets.
Table~\ref{tab:benchmark_overview} summarizes the resulting settings used throughout every reported comparison.
Appendix~\ref{app:sample_level_sensor_fault_examples} visualizes representative input and forecast windows from each dataset.

\begin{table*}[tp]
  \centering
  \small
  \setlength{\tabcolsep}{2.0pt}
  \renewcommand{\arraystretch}{1.18}
  \caption{Four forecasting tasks under the shared perturbation and scoring protocol.
  Columns list target variables, lengths of the retained hourly series, numbers of observed input channels and forecast targets, input window lengths, and forecast horizons.}
  \vspace{7pt}
  \label{tab:benchmark_overview}
  \begin{tabularx}{\textwidth}{@{}l l >{\RaggedRight\arraybackslash}X c c c c c@{}}
    \toprule
    & & & Series & Input & Forecast & Input & Forecast \\
    Data & Domain & Target & length & channels & targets & window & horizon \\
    \midrule
    \BeijingTiantan{} & air quality & PM2.5 & 24{,}038 & 12 & 1 & 96 & 6 \\
    \PenmanshielWT{} & wind turbine SCADA & Power (kW) & 25{,}867 & 65 & 1 & 96 & 6 \\
    ETTh1 & transformer telemetry & all channels & 17{,}420 & 7 & 7 & 96 & 96 \\
    Traffic & freeway lane occupancy & all channels & 17{,}544 & 862 & 862 & 96 & 96 \\
    \bottomrule
  \end{tabularx}
\end{table*}

\subsection{Compared forecasting architectures and robustness-improvement methods}

We use \emph{forecasting architecture} for a base forecaster evaluated without an added robustness-improvement protocol, and \emph{robustness-improvement method} for a training, adaptation, aggregation, or wrapper protocol evaluated against the baseline forecasting model of the same architecture.
The architecture comparison spans DLinear as a linear forecaster \citep{zengAreTransformersEffective2023}, gated recurrent unit (GRU) as a recurrent model \citep{choLearningPhraseRepresentations2014}, ModernTCN as a convolutional model \citep{donghaoModernTCNModernPure2023}, PatchTST as a patch-based transformer \citep{nieTimeSeriesWorth2022}, TSMixer as an all-MLP model \citep{chen2023tsmixer}, SeasonalNaive as a simple reference forecaster recommended in forecast evaluation practice \citep{hewamalageForecastEvaluationData2023,hyndmanForecastingPrinciplesPractice2021}, and Chronos-2 as the zero-shot foundation model \citep{ansariChronos2UnivariateUniversal2025}.
SeasonalNaive forecasts by repeating the most recent block whose length is the selected seasonal period.
A one-step period gives the last-value naive forecaster, and, on hourly data, a 24-step period gives the daily seasonal naive forecaster \citep{hyndmanForecastingPrinciplesPractice2021}.
The clean-validation selector chooses the one-step period for \BeijingTiantan{} and \PenmanshielWT{} and the 24-hour period for ETTh1 and Traffic.
Robustness-improvement methods are benchmarked on the five architectures trained on each dataset, so SeasonalNaive and Chronos-2 enter only the architecture comparison.

The evaluated method set includes ensemble aggregation for time series \citep{barrowEvaluationNeuralNetwork2010}, projected gradient descent (PGD) adversarial training \citep{madry2018towards}, randomized training for probabilistic forecasting \citep{yoonRobustProbabilisticTime2022}, randomized smoothing of a fixed predictor with alpha-trimmed aggregation for regression \citep{rekavandiCertifiedAdversarialRobustness2024}, adaptive robust loss \citep{barronGeneralAdaptiveRobust2019}, Reversible Instance Normalization (RevIN) \citep{Kim.etal_2022}, and fault augmentation, a data augmentation baseline that trains on $\mathcal{P}_{\mathrm{trans}}$ to test transfer to the scored scenarios.
Of all evaluated architectures and robustness-improvement methods, only fault augmentation uses $\mathcal{P}_{\mathrm{trans}}$ during training.
The architecture baselines, model selection, and other methods do not train on those transfer families.
RevIN is evaluated on four paired architectures because PatchTST already applies per-window instance normalization and de-normalization.
The operational taxonomy organizes this non-exhaustive comparison, with broader coverage and extension paths recorded in Appendix~\ref{sec:taxonomy}.

\section{Protocol for structured sensor faults}
\label{sec:sensor_fault_protocol}

Sensor faults such as noise, drift, stuck readings, and missing data are recurring data-quality problems in sensor streams that monitor physical processes~\citep{jesusSurveyDataQuality2017,tehSensorDataQuality2020}.
SensorFault-Bench simulates these faults one at a time in the observed input history, while the target remains the future process state rather than a stuck, flat, missing, or misaligned trace.
Comparing clean and corrupted histories against the same target attributes the change in forecasting error to the injected input fault.
The selection criteria in Appendix~\ref{sec:selection_criteria} therefore exclude target faults, process-level shifts, and cascading or closed-loop effects from scoring.
Section~\ref{sec:discussion} furthermore discusses composed faults among the limitations.
Scenario degradation compares perturbed and clean test MSE within each benchmark scenario, so worst-scenario degradation measures performance retention under the evaluated faults.
Clean MSE and worst-scenario fault-time MSE keep nominal accuracy and absolute fault-time error visible.

\subsection{Benchmark scenarios}
\label{sec:benchmark_scenarios}
\begin{table}[t]
  \centering
  \small
  \setlength{\tabcolsep}{1.0pt}
  \renewcommand{\arraystretch}{1.12}
  \caption{Compact protocol specification for the evaluated benchmark~$\mathcal{P}$.
  Severity $s\in[0,1]$ determines the scenario-specific quantity shown in the table, with endpoints at $s=0$ and $s=1$.
  Scenarios scoped to channels perturb a subset of continuous channels that grows with severity, up to half of the eligible channels.
  Appendix~\ref{sec:implementation_details_scenarios} gives formal definitions and implementation details.}
  \label{tab:perturbation_parameters}

  \newcolumntype{Y}{>{\raggedright\arraybackslash}X}
  \begin{tabularx}{\columnwidth}{@{}
    l
    @{\hspace{0.65em}}
    l
    >{\hsize=.9\hsize\raggedright\arraybackslash}X
    >{\hsize=1.3\hsize\raggedright\arraybackslash}X
    >{\hsize=.8\hsize\raggedright\arraybackslash}X
    c
    @{\hspace{0.45em}}
    c
    @{}}
    \toprule
	    Class &
	    Scenario &
	    Channels &
	    Window &
	    Quantity &
	    $s=0$ &
    $s=1$ \\
    \midrule
    \multirow[t]{4}{*}{Value}
      & Drift        & continuous subset & full window & offset (std) & 0 & 0.75 \\
      & Attenuation  & continuous subset & full window & ampl. factor & 1 & 0.25 \\
      & Noise        & continuous subset & full window & Gaussian std & 0 & 1 \\
      & Spike        & continuous subset & one step per channel & magnitude (std) & 0 & 7.5 \\
    \addlinespace
    \multirow[t]{2}{*}{Timing}
      & TimeStretch  & continuous subset & shared window ($\lceil n/2\rceil$) & resampling factor & 1 & 5 \\
      & TimeCompress & continuous subset & shared window ($\lceil n/2\rceil$) & resampling factor & 1 & 0.1 \\
    \addlinespace
    \multirow[t]{2}{*}{Availability}
      & StuckSensor  & continuous subset & per-channel window ($\lceil\theta(n\!-\!1)\rceil$) & frozen frac. & 0 & 1 \\
      & MissingData  & all channels & shared gap ($\lceil\theta(n\!-\!1)\rceil$) & gap frac. & 0 & 0.5 \\
    \bottomrule
  \end{tabularx}
\end{table}

The scored set $\mathcal{P}$ contains eight benchmark scenarios grouped into value, timing, and availability faults (Figure~\ref{fig:scenarios}, Table~\ref{tab:perturbation_parameters}).
A benchmark scenario is one evaluated sensor fault, such as MissingData or Noise, with severity and application rules fixed by the protocol.
\textbf{Value} faults alter recorded amplitudes through Drift, Attenuation, Noise, and Spike.
\textbf{Timing} faults alter the effective sampling through TimeStretch and through TimeCompress, which combines compression with endpoint holding by repeating the final observed value.
\textbf{Availability} faults remove information through StuckSensor and MissingData.
The sample-level examples in Appendix~\ref{app:sample_level_sensor_fault_examples} show how these scenarios alter real input windows.
We examine the effect of fault direction by reversing Drift and Spike in Appendix~\ref{subsec:fault_direction_sensitivity}.
The timing analysis in Appendix~\ref{subsec:timecompress_dependence} replaces TimeCompress with a local warp that preserves both window endpoints.

The seven transfer fault families in $\mathcal{P}_{\mathrm{trans}}$ act on continuous channels and serve only for training (Appendix~\ref{sec:test_scenarios}).
Their role is to support fault-aware training protocols, including future methods, that test transfer from adjacent fault families to $\mathcal{P}$ without training on scored scenarios.
In this paper, only fault augmentation uses them.
Architecture baselines, model selection, wrappers, and all other robustness-improvement methods use neither transfer-family augmentation nor scored-scenario replay, keeping the transfer set separate from scoring.

\subsection{Severity and channel perturbation design}
\label{sec:severity_channel_design}

Each benchmark scenario uses a severity variable $s \in [0,1]$ from a benign endpoint to the strongest tested condition.
Evaluation samples severity uniformly, so parameter endpoints and, where applicable, the channel-count rule define a tested severity profile rather than a single evaluation condition.
Fault-family provenance and endpoint calibration are separate.
Reported or injected value and availability fault scales inform some endpoint ranges where applicable \citep{Balaban2009_IEEESensors_SensorFaultsAerospace,Ni2009_TOSN_SensorNetworkDataFaultTypes,Sharma2010_TOSN_SensorFaultPrevalence,Hansen2022_Water_DriftDetectionProcessTanks}, and corruption-benchmark practice motivates deliberately strong upper endpoints for unusual or long-tail shifts \citep{Hendrycks.etal_2022}.
The final endpoints for normalized magnitude and timing resampling and the cap on affected channels are standardized settings for stress testing, chosen for comparability across datasets and tractability rather than estimated from field prevalence.
These settings and the severity sampling distribution are explicit parameters (Appendix~\ref{subsec:perturbation_general_setup}), so they can also be calibrated to the recorded faults of a specific system.
Different sensor technologies and physical fault causes can produce recurring effects in recorded streams, such as noise, fixed readings, and missing data \citep{jesusSurveyDataQuality2017,tehSensorDataQuality2020}, so simulating these effects under common rules makes forecasting robustness comparable across datasets.
For scenarios scoped to channels, increasing $s$ also enlarges the subset of affected continuous channels, up to half of the eligible channels, so the score averages over both corruption strength and affected-channel count.
Appendix~\ref{sec:implementation_details_scenarios} specifies window placement, channel-count rules, and training-statistic standardization.
Appendix~\ref{subsec:channel_count_sensitivity} tests the channel-count rule by fixing the fraction of selected channels.

\paragraph{Design rationale.}
\label{sec:design_rationale}
The \textbf{Value} class is grounded in studies of sensor faults and process monitoring that report noisy readings, spikes, offset or gain errors, loss of accuracy, and drift \citep{Balaban2009_IEEESensors_SensorFaultsAerospace,Ni2009_TOSN_SensorNetworkDataFaultTypes,Sharma2010_TOSN_SensorFaultPrevalence,Hansen2022_Water_DriftDetectionProcessTanks,Luca2023_AppliedSciences_DOSensorFaultDiagnosis}.
The \textbf{Timing} class is grounded in studies of channel alignment during sampling, timing in SCADA event logs, synchronization delays, delayed records, timestamp pathologies, poor timestamping, and delay variability \citep{Funck2014_Measurement_TimeSynchronousSampling,Andrade2022_IEEEAccess_SCADAAlarmAnomalyDetection,Kalappa2006_IEEE1588_TimeSynchronizationDataQuality,Schmetz2022_ProcediaCIRP_TimeSynchronizationProblem}.
The \textbf{Availability} class is grounded in studies of sensor networks, data quality, and data acquisition from programmable logic controllers that report missing values, stuck or fixed readings, and data loss during transfer \citep{Ni2009_TOSN_SensorNetworkDataFaultTypes,Sharma2010_TOSN_SensorFaultPrevalence,hubauerAnalysisDataQuality2013,jesusSurveyDataQuality2017,tehSensorDataQuality2020,Hijazi2024_Heliyon_PLCDataLossSynchronization}.
Industrial and CPS robustness studies provide complementary precedent for controlled perturbation evaluation and severity-controlled forecasting benchmarks \citep{Dix.etal_2023,windmannQuantifyingRobustnessBenchmarking2025}.
Appendix~\ref{sec:test_scenarios} links each scenario to documented sensor faults and separates reported scales from standardized benchmark settings.
Appendix~\ref{sec:selection_criteria} gives the selection rationale for excluding model-dependent adversarial attacks, generic augmentations, generative perturbations, and system-specific simulators from the scored scenarios.

\section{Robustness score and evaluation}
\label{sec:robustness_score}

Common-corruption benchmarks separate corrupted performance from relative degradation \citep{hendrycksBenchmarkingNeuralNetwork2019,Michaelis.etal_2019}, while certified-robustness work reports performance as a function of perturbation radius \citep{cohenCertifiedAdversarialRobustness2019}.
SensorFault-Bench keeps this separation and reports worst-scenario degradation together with clean MSE and worst-scenario fault-time MSE, so relative robustness and absolute error stay distinct.

After training on $\mathcal{D}_{\mathrm{train}}$, models are selected using only clean $\mathcal{D}_{\mathrm{val}}$ and evaluated on a shared set of test windows from $\mathcal{D}_{\mathrm{test}}$.
A clean pass establishes each model's clean error before each scenario draws severity and auxiliary randomness per window.
Aggregating these errors yields clean MSE, worst-scenario degradation, and worst-scenario fault-time MSE, with method effects computed as differences from the paired baseline forecasting model on the same dataset and architecture.
Figure~\ref{fig:scenarios} gives an overview of this process, while Algorithm~\ref{alg:benchmark_loop} in Appendix~\ref{ch:robustness_score} formalizes the evaluation loop.

\subsection{Degradation per scenario}
\label{sec:uniform_sampling_degradation}

The benchmark evaluates each selected model $f$ under every scenario $p\in\mathcal{P}$, with severity sampled uniformly over $[0,1]$ so the score averages the full severity profile rather than a single endpoint.
Averaging over this profile, which couples intensity with the number of affected channels where present, captures both the onset and the magnitude of error increases, so models with similar error at $s=1$ remain distinguishable when their deterioration begins at different severities.
Testing only the strongest condition would miss this distinction.
The score compares expected perturbed test MSE under this sampling distribution with expected clean test MSE on the same held-out test-sample distribution.
The estimator in Appendix~\ref{ch:robustness_score} uses one shared sample of test windows for clean and perturbed evaluations and draws fresh severity and perturbation randomness per scenario-sample pair.
Let $X'_p$ denote the perturbed input after applying scenario $p$ to $X$ under the perturbation model (Appendix~\ref{sec:perturbation_model}), and let $\mathrm{MSE}_{p}(f)$ and $\mathrm{MSE}_c(f)$ denote perturbed and clean expected test MSE.
Assuming $\mathrm{MSE}_c(f)>0$, the corresponding scenario degradation score is
\begin{equation}
\label{eq:dp}
  \mathcal{D}_{p}(f)=\frac{\mathrm{MSE}_{p}(f)}{\mathrm{MSE}_c(f)}.
\end{equation}
A value of $\mathcal{D}_{p}(f)=1$ indicates no expected error increase under scenario $p$, while larger or smaller values indicate higher or lower perturbed-input MSE.
For example, $\mathcal{D}_{p}(f)=2$ means the perturbed expected MSE is twice the clean expected MSE.
Appendix~\ref{subsec:severity_weighting_sensitivity} reweights severity toward mild faults.

\subsection{Worst-scenario score}
\label{sec:aggregate_metrics}

For any evaluated model $f$, relative robustness is measured by worst-scenario degradation:
\begin{equation}
  \label{eq:dmax}
  \mathcal{D}_w(f)=\max_{p\in\mathcal{P}}\mathcal{D}_{p}(f).
\end{equation}
The maximum is taken over the eight scenario degradations, which are already averaged over severity.
Lower $\mathcal{D}_w$ therefore means better relative robustness at the model's worst benchmark scenario.
The maximum prevents low mean degradation across scenarios from masking a severe failure in one scenario.
This follows worst-case evaluation under distribution shift \citep{liEvaluatingModelPerformance2021, Subbaswamy.etal_2021} and aligns with uniform-performance and distributionally robust viewpoints \citep{Duchi.Namkoong_2021, Dziugaite.etal_2020}.
Appendix~\ref{app:metric_mapping} evaluates scenario means as alternatives.

The accompanying absolute-error measures keep relative score separate from error scale: $\mathrm{MSE}_c(f)$ records clean error, while $\mathrm{MSE}_w(f)=\max_{p\in\mathcal{P}}\mathrm{MSE}_{p}(f)$ records fault-time error at the worst scenario.
Because $\mathrm{MSE}_c(f)$ is fixed with respect to $p$, the scenario maximizing $\mathcal{D}_{p}(f)$ also maximizes $\mathrm{MSE}_{p}(f)$.
Thus $\mathrm{MSE}_w(f)$ is evaluated at the same model-specific worst scenario as $\mathcal{D}_w(f)$, and that scenario can differ across architectures or methods.
It is an absolute fault-time error measure at each model's own worst benchmark scenario, not an error comparison under one common scenario.
\citet{taoriMeasuringRobustnessNatural2020} show in image classification that apparent robustness gains under distribution shift can reflect higher standard accuracy.
Likewise, lower $\mathrm{MSE}_w$ can reflect lower $\mathrm{MSE}_c$ rather than lower fault sensitivity, so the relative-robustness score remains $\mathcal{D}_w$.

\subsection{Evaluation of robustness-improvement methods}

Method evaluation is paired at the dataset-architecture level.
For a given dataset and forecasting architecture, $f_b$ denotes the evaluated no-intervention forecasting model and $f'$ the evaluated variant produced by applying one robustness-improvement method to that same architecture.
The method deltas are direct differences, for example $\Delta\mathcal{D}_w(f',f_b)=\mathcal{D}_w(f')-\mathcal{D}_w(f_b)$, with negative values favoring the robustness-improved variant.
The corresponding $\Delta\mathrm{MSE}_c$ and $\Delta\mathrm{MSE}_w$ keep clean-error and worst-scenario fault-time changes explicit.
For method evaluation, $\Delta\mathcal{D}_w$ is the clean-controlled robustness comparison, separating changes in fault-induced degradation from changes in clean error.
The MSE deltas report the matching nominal and worst-scenario fault-time error changes, with $\Delta\mathrm{MSE}_w$ comparing the two models at their respective worst scenarios.
Within each dataset, method summaries average these pairwise deltas over the available paired forecasting architectures, and Appendix~\ref{ch:robustness_score} gives the arithmetic definitions.

\subsection{Selection and evaluation protocol}

All four datasets use temporal splits with 0.6 train, 0.2 validation, and 0.2 test proportions.
The default configuration (Table~\ref{tab:experimental_protocol}) uses up to 40 seeded random-subgrid candidates per dataset--architecture--method configuration, capped at 200 epochs with validation-based early stopping.
This keeps search effort comparable because hyperparameter choices and tuning budgets can materially affect empirical comparisons \citep{Sivaprasad2020_OptimizerBenchmarkingHPO,Bouthillier2021_AccountingVarianceBenchmarks}.
Within this budget, candidate settings are drawn at random from a hyperparameter grid, and validation-based early stopping ends each training run \citep{bergstraRandomSearchHyperParameter2012,precheltEarlyStoppingWhen2012}.
Winner selection uses only clean validation information and excludes benchmark-fault validation performance and test-set metrics.
Baseline forecasting models and end-to-end trained methods use the clean validation objective, while post-hoc methods use the wrapper's own clean validation MSE with the same deterministic tiebreak.
This clean-selection rule asks which forecasting architectures and robustness-improvement methods remain robust before the realized sensor-fault family and severity are known.
Appendix~\ref{sec:selector_pressure} tests sensitivity to the clean-selection rule, Appendix Table~\ref{tab:experimental_protocol} summarizes the experimental settings, including budgets and seeds, and the repository guide in Appendix~\ref{app:repo_guide} maps the protocol to its implementation and extension interface.

\section{Results}
\label{sec:results}

The benchmark yields four practical findings: clean MSE can miss sharp degradation under sensor faults, Chronos-2 has either high clean error or low robustness on each dataset, degradation concentrates in different scenario families across dataset settings, and no evaluated robustness-improvement method transfers consistently across architectures, datasets, and reported quantities.
Table~\ref{tab:main_results} gives the dataset-level synthesis, and Figure~\ref{fig:baseline_architecture_heatmap} separates scenario structure.
Both come from one training and search seed, whereas Appendix~\ref{app:extended_results} adds results over two further seeds and bootstrap intervals.

\subsection{Clean MSE can miss sharp degradation under sensor faults}
\label{sec:results_architectures}

\begin{table*}[t]
  \centering
  \small
  \setlength{\tabcolsep}{1.0pt}
  \renewcommand{\arraystretch}{1.00}
  \caption{Per-dataset architecture scores and robustness-improvement method-baseline deltas.
  Architecture columns report $\mathcal{D}_w$ (Eq. \eqref{eq:dmax}), $\mathrm{MSE}_c$, and $\mathrm{MSE}_w$ at each model's own worst scenario, and method columns report mean paired deltas.
  Lower values are better, negative deltas indicate improvement, and bold marks each block's best score.
  Seas.\! Naive denotes SeasonalNaive, ModTCN ModernTCN, TSMix TSMixer, and Chr2 Chronos-2.
  Rand.\! Train denotes randomized training, Rand.\! Smooth randomized smoothing, ARL adaptive robust loss, and Fault Aug. fault augmentation.}
  \vspace{7pt}
  \label{tab:main_results}
  \begin{adjustbox}{max width=\textwidth}
  \begin{tabular}{@{}cl*{7}{S[
    table-format=1.3,
    table-number-alignment=center,
    mode=text,
    detect-weight=true
  ]}@{\hspace{6pt}}l*{7}{S[
    table-format=-1.3,
    table-number-alignment=center,
    mode=text,
    detect-weight=true
  ]}@{}}
    \toprule
    \multicolumn{9}{c}{\textbf{Baseline architecture comparison}} &
    \multicolumn{8}{c}{\textbf{Robustness-improvement method comparison}} \\
    \cmidrule(r{6pt}){1-9} \cmidrule{10-17}
    Data & Measure
      & {\makecell{Seas.\\Naive}} & {DLinear} & {GRU} & {\makecell{Mod\\TCN}} & {TSMix} & {\makecell{Patch\\TST}} & {Chr2}
      & Measure
      & {PGD} & {Ensemble} & {\makecell{Rand.\\Train}} & {\makecell{Rand.\\Smooth}} & {ARL} & {RevIN} & {\makecell{Fault\\Aug.}} \\
    \midrule
    \multirow{3}{*}{\rotatebox[origin=c]{90}{\BeijingTiantanTinyCentered{}}}
      & $\mathcal{D}_w$ & 1.353 & 1.628 & 1.251 & 1.502 & \bfseries 1.153 & 1.376 & 1.230
      & $\Delta\!\mathcal{D}_w$ & -0.136 & 0.007 & \bfseries -0.146 & 0.001 & -0.024 & 0.053 & -0.039 \\
      & $\mathrm{MSE}_c$ & 0.353 & 0.329 & \bfseries 0.281 & 0.301 & 0.359 & 0.338 & 0.349
      & $\Delta\!\mathrm{MSE}_c$ & 0.004 & \bfseries -0.009 & 0.026 & 0.002 & 0.001 & 0.005 & 0.005 \\
      & $\mathrm{MSE}_w$ & 0.477 & 0.536 & \bfseries 0.352 & 0.452 & 0.414 & 0.465 & 0.429
      & $\Delta\!\mathrm{MSE}_w$ & \bfseries -0.038 & -0.009 & -0.018 & 0.003 & -0.005 & 0.025 & -0.007 \\
    \addlinespace
    \multirow{3}{*}{\rotatebox[origin=c]{90}{\PenmanshielWTTinyCentered{}}}
      & $\mathcal{D}_w$ & 1.320 & 1.259 & \bfseries 1.064 & 1.183 & 1.256 & 1.202 & 1.160
      & $\Delta\!\mathcal{D}_w$ & \bfseries -0.059 & -0.023 & -0.036 & -0.002 & 0.144 & 0.110 & -0.019 \\
      & $\mathrm{MSE}_c$ & 0.389 & 0.340 & 0.346 & \bfseries 0.336 & 0.353 & 0.354 & 0.401
      & $\Delta\!\mathrm{MSE}_c$ & -0.005 & \bfseries -0.007 & -0.003 & 0.003 & 0.040 & -0.001 & -0.001 \\
      & $\mathrm{MSE}_w$ & 0.514 & 0.428 & \bfseries 0.368 & 0.397 & 0.443 & 0.426 & 0.466
      & $\Delta\!\mathrm{MSE}_w$ & \bfseries -0.026 & -0.016 & -0.016 & 0.003 & 0.102 & 0.037 & -0.008 \\
    \addlinespace
    \multirow{3}{*}{\rotatebox[origin=c]{90}{\tiny ETTh1}}
      & $\mathcal{D}_w$ & 1.288 & 1.251 & \bfseries 1.064 & 1.216 & 1.169 & 1.312 & 1.448
      & $\Delta\!\mathcal{D}_w$ & 0.000 & \bfseries -0.005 & -0.004 & 0.000 & 0.001 & 0.124 & -0.001 \\
      & $\mathrm{MSE}_c$ & 0.634 & 0.438 & 0.664 & 0.462 & 0.475 & \bfseries 0.431 & 0.496
      & $\Delta\!\mathrm{MSE}_c$ & -0.005 & -0.005 & 0.009 & -0.001 & 0.033 & \bfseries -0.045 & 0.003 \\
      & $\mathrm{MSE}_w$ & 0.817 & \bfseries 0.548 & 0.707 & 0.562 & 0.555 & 0.566 & 0.718
      & $\Delta\!\mathrm{MSE}_w$ & -0.005 & \bfseries -0.007 & 0.008 & -0.001 & 0.037 & 0.010 & 0.002 \\
    \addlinespace
    \multirow{3}{*}{\rotatebox[origin=c]{90}{\tiny Traffic}}
      & $\mathcal{D}_w$ & 1.210 & \bfseries 1.177 & 1.284 & 1.265 & 1.290 & 1.676 & 1.911
      & $\Delta\!\mathcal{D}_w$ & 0.023 & -0.019 & 0.011 & -0.021 & -0.007 & 0.267 & \bfseries -0.025 \\
      & $\mathrm{MSE}_c$ & 1.270 & 0.684 & 0.726 & 0.575 & 0.631 & \bfseries 0.453 & 0.602
      & $\Delta\!\mathrm{MSE}_c$ & -0.020 & -0.017 & -0.007 & 0.011 & 0.075 & \bfseries -0.063 & -0.013 \\
      & $\mathrm{MSE}_w$ & 1.537 & 0.806 & 0.932 & \bfseries 0.728 & 0.814 & 0.759 & 1.150
      & $\Delta\!\mathrm{MSE}_w$ & -0.016 & \bfseries -0.037 & -0.003 & 0.008 & 0.100 & 0.071 & -0.026 \\
    \bottomrule
  \end{tabular}
  \end{adjustbox}
  \vspace{-8pt}
\end{table*}

Clean-MSE rankings can hide sharp degradation and higher worst-scenario fault-time error under sensor faults.
On \PenmanshielWT{}, ModernTCN and DLinear slightly beat GRU on clean MSE, but GRU has the lowest $\mathcal{D}_w$ and $\mathrm{MSE}_w$.
On Traffic, PatchTST has the lowest $\mathrm{MSE}_c$ but the second-largest $\mathcal{D}_w$ behind Chronos-2, while ModernTCN has lower $\mathcal{D}_w$ and $\mathrm{MSE}_w$.
On \BeijingTiantan{}, GRU has the lowest clean and worst-scenario fault-time MSE despite TSMixer's lower degradation.
On ETTh1, GRU, PatchTST, and DLinear lead $\mathcal{D}_w$, $\mathrm{MSE}_c$, and $\mathrm{MSE}_w$, respectively, so no single model dominates.
The model favored by clean MSE can therefore be sensitive to faults, while another model has the lowest error under faults.
Low degradation does not imply low fault-time error either, so none of the three quantities can stand in for the others.

Chronos-2, the evaluated zero-shot foundation model, has either high clean error or low robustness on each dataset.
Its clean MSE matches the last-value naive forecaster on \BeijingTiantan{} and trails it on \PenmanshielWT{}, even though its $\mathcal{D}_w$, which is normalized by its own clean MSE, is the second lowest on both datasets.
On the all-channel datasets, Chronos-2 has the largest $\mathcal{D}_w$ on both ETTh1 and Traffic (Appendix~\ref{app:per_dataset_arch}), so Traffic's acceptable clean MSE would be misleading without the degradation score and the reported fault-time error.

The architectures with the lowest clean MSE, worst-scenario degradation, and worst-scenario fault-time MSE persist across three training and search seeds (Appendix~\ref{subsec:training_seed_sensitivity}), with Chronos-2 and SeasonalNaive held fixed.
Exact rankings can change, but for 18 of the 20 combinations of retrained architecture and dataset the range of $\mathcal{D}_w$ across seeds is at most 0.118, and the largest ranges occur for PatchTST on \BeijingTiantan{} and TSMixer on \PenmanshielWT{}.

\subsection{Scenario difficulty is structured by dataset setting}
\label{sec:results_scenarios}

\begin{figure*}[!htb]
  \centering
  \includegraphics[width=\textwidth]{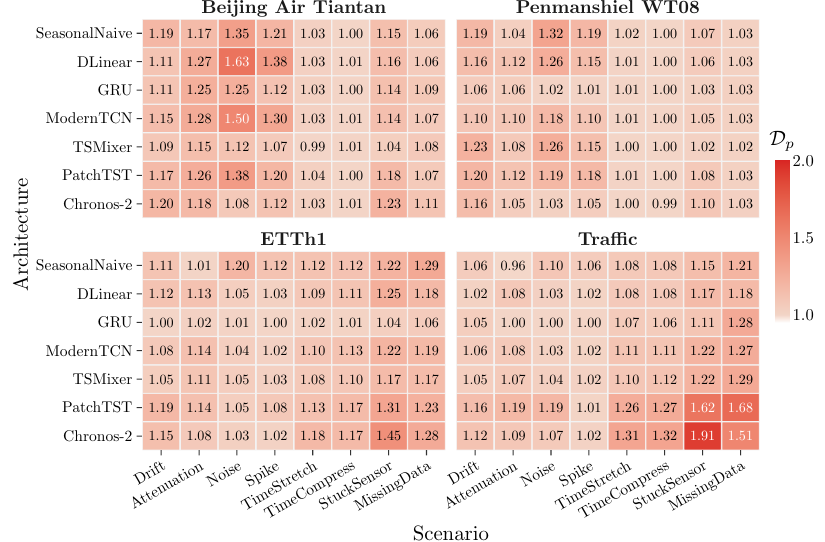}
  \caption{Degradation $\mathcal{D}_{p}(f)$ (Eq.~\eqref{eq:dp}) per benchmark scenario for the baseline forecasting models.
  Rows follow Table~\ref{tab:main_results}, columns group scenario classes, and darker cells indicate larger degradation within each dataset and architecture.}
  \label{fig:baseline_architecture_heatmap}
\end{figure*}

Figure~\ref{fig:baseline_architecture_heatmap} separates the architecture results by benchmark scenario.
The largest degradation concentrates in value perturbations, especially noise, for \BeijingTiantan{} and \PenmanshielWT{}, while ETTh1 and Traffic shift the strongest penalties toward availability faults.
This contrast remains descriptive because forecast horizon, target multiplicity, channel count, and affected-channel scope change together across the four datasets.
In Appendix~\ref{subsec:etth1_task_ablation}, an ablation within ETTh1 varies target scope and forecast horizon independently while keeping all seven input channels, the 96-step history, and the perturbation rules fixed.
Restricting the targets to oil temperature changes the worst-scenario class from availability to value faults for all seven architectures at horizon 96 and for six at horizon 6.
DLinear is the exception at horizon 6, where Attenuation is the worst scenario under both target scopes.
Shortening the horizon preserves the worst-scenario class for six of seven architectures when forecasting all channels and for all seven when forecasting oil temperature alone.
Thus, target scope changes the worst-scenario class more consistently than horizon within ETTh1.

\subsection{Robustness-improvement methods do not transfer uniformly}
\label{sec:results_methods}

Table~\ref{tab:main_results} reports paired mean deltas, with 95\% percentile bootstrap intervals over method-baseline architecture pairs in Appendix Table~\ref{tab:appendix_method_delta_ci}.
These intervals summarize variation across these pairs, not retraining or selector variability.
Viewed with the scenario heatmap, the deltas show scenario-specific robustness improvements for the evaluated method set.
On the single-target datasets, value faults, especially Noise, produce the largest observed degradation, and PGD adversarial training and randomized training reduce degradation most.
On Traffic, availability faults dominate, and fault augmentation reduces degradation, whereas PGD lowers absolute error but increases degradation.
Ensemble aggregation is the most consistent error reducer.
RevIN gives the cautionary contrast: it lowers clean MSE on ETTh1 and Traffic but increases degradation in the same paired comparisons.

Across three training and search seeds, the main degradation contrasts persist for PGD and RevIN, while gains from fault augmentation depend on the dataset (Appendix~\ref{subsec:training_seed_sensitivity}).
PGD reduces degradation on both single-target datasets and increases it on Traffic.
RevIN increases both mean worst-scenario degradation and mean worst-scenario fault-time MSE on every dataset and seed.
Fault augmentation consistently reduces degradation on Traffic, but its effect changes sign across seeds on \BeijingTiantan{} and \PenmanshielWT{}.

\section{Discussion}
\label{sec:discussion}

SensorFault-Bench exposes a weakness in nominal forecasting evaluation: models favored by clean MSE can degrade sharply under sensor faults.
Worst-scenario degradation measures relative robustness, while clean MSE and worst-scenario fault-time MSE retain nominal and absolute fault-time error.
As in shifted accuracy, lower fault-time MSE can reflect lower clean MSE rather than lower relative fault sensitivity \citep{taoriMeasuringRobustnessNatural2020}.
Chronos-2, the evaluated zero-shot foundation-model representative, matches or trails the last-value naive forecaster on the single-target datasets and has the largest worst-scenario degradation on ETTh1 and Traffic.
A separate cloud-forecasting study reports a similar pattern: \citet{tonerPerformanceZeroShotTime2025} find evaluated time-series foundation models trailing a seasonal naive forecaster and producing inaccurate or chaotic forecasts.

Robustness-improvement method comparisons use deltas against the paired baseline forecasting model before aggregation.
Persistent degradation reductions across datasets and architectures provide stronger transfer evidence than gains concentrated in one forecaster or dataset.
Thus, no evaluated method dominates across datasets, architectures, scenario classes, or reported quantities.

\paragraph{Limitations and responsible use.}
SensorFault-Bench supports stress testing before deployment by injecting simulated faults grounded in documented CPS sensing failures into real input histories while keeping forecasting targets fixed.
High degradation reveals sensitivity to the defined corruption, while low degradation supports robustness only within the evaluated scenarios.
Transfer to faults observed in deployment requires separate validation against recorded fault episodes from the intended system, for example in field studies that compare model rankings under simulated and recorded faults.
The common severity scale enables comparison across datasets, but equal normalized severity need not represent equal physical fault magnitude across sensor technologies.
The shared scenarios may omit faults specific to a technology and do not estimate deployment fault frequencies, so their scores cannot certify safety or determine operating limits.

Evaluating one scenario at a time isolates its effect on forecasting error, leaving composed faults and their interactions for further study.
Persistent faults in target sensors, process shocks, and closed-loop effects require different target or system models.
The four datasets span distinct CPS forecasting settings but not all domains, sensor modalities, or control regimes.
The paired robustness-improvement method panel is architecture-controlled but small, with bounded tuning for compute-matched rather than exhaustive comparisons.
Residual analysis by scenario would complement the aggregate MSE measures by revealing how forecast errors are distributed over time.

\section{Conclusion}
\label{sec:conclusion}

SensorFault-Bench stress-tests forecasting under simulated CPS sensor faults.
Across four tasks and eight scenarios, its three quantities separate nominal accuracy, fault sensitivity, and fault-time error.
Forecasting architectures favored by clean MSE can degrade sharply under these faults.
Robustness-improvement methods mitigate such degradation only selectively, with no evaluated method transferring consistently across architectures, datasets, and reported quantities.
PGD adversarial training and randomized training reduce degradation most where value faults dominate, whereas fault augmentation reduces it most on Traffic, where availability faults dominate.
Zero-shot Chronos-2 has either high clean error or low robustness on each dataset.
Fixed scenarios and clean selection rules keep extensions to new methods, datasets, and architectures comparable.

\begin{ack}

AW and ON acknowledge support from the project SmartShip,
PW from the project SHM, and GM from the projects RIVA and
ProMoDi, all funded by dtec.bw --
Digitalization and Technology Research Center of the
Bundeswehr, which we gratefully acknowledge. dtec.bw is
funded by the European Union -- NextGenerationEU.
GM and MD are supported by the German
Federal Ministry of Research, Technology and Space
(BMFTR, formerly BMBF) through the
project EXPLAIN under grant 01IS22030. GM is also
supported by the German Federal Ministry for Economic Affairs
and Energy (BMWE, formerly BMWK) through the project kaMeL under
grant 20X2206H.
JUB is supported by the project ShapeFuture under
grant 101139996 of the European call \mbox{HORIZON-KDT-JU-2023-2-RIA}
and by the BMFTR through the
project TH~Köln -- \mbox{Künstliche} Intelligenz plus (THK-KIplus)
within the funding measure Forschung an Fachhochschulen~--
\mbox{KI-Nachwuchs@FH~2-2021} under grant 13FH007KI2.

GM was employed by ABB during part of the
36 months before submission.
The other authors declare no competing interests.
\end{ack}


\bibliographystyle{plainnat}
\bibliography{references}


\clearpage
\appendix

\section*{Appendix}

The appendices are organized by evidence role.
Appendices~\ref{app:dataset_task_details} and~\ref{app:perturbation_definitions} specify the benchmark construction, covering dataset derivation, target choices, scenario-selection criteria, and perturbation mechanics.
Appendix~\ref{app:sample_level_sensor_fault_examples} visualizes representative perturbed input and forecast windows.
Appendices~\ref{ch:robustness_score} and~\ref{sec:benchmark_design_validation} define the robustness quantities and audit sensitivity to evaluation data seeds, selectors, channel counts, severity weighting, fault direction, and TimeCompress.
Appendices~\ref{app:full_experimental_details} and~\ref{app:repo_guide} record the experimental configuration and repository interface for reproduction and extension.
Appendices~\ref{app:extended_results} and~\ref{sec:taxonomy} collect extended results and the coverage map that situates the evaluated method set.

\section{Dataset and task details}
\label{app:dataset_task_details}

The benchmark fixes the four dataset-specific forecasting tasks listed in Table~\ref{tab:benchmark_overview}.
Every model receives the full multivariate input history, but each dataset definition fixes the forecast targets used for validation, scoring, and reported errors.
ETTh1 and Traffic use all-channel targets, whereas \BeijingTiantan{} and \PenmanshielWT{} are curated single-target datasets derived from larger source archives.
The single-target forecast targets are PM2.5 and Power (kW), respectively.

The four datasets cover complementary CPS forecasting settings: urban air quality, wind turbine SCADA, electricity transformer telemetry, and freeway lane occupancy.
They define two task regimes under one perturbation and scoring protocol.
\BeijingTiantan{} and \PenmanshielWT{} use a 96-step input window to forecast one target series for 6 steps, with 12 and 65 observed input channels, respectively.
ETTh1 and Traffic use the same 96-step input window but forecast all channels over a 96-step horizon, yielding 7 and 862 target series, respectively.
These choices vary domain, input dimensionality, target multiplicity, forecast horizon, and affected-channel scope while keeping the full architecture, method, and scenario evaluation tractable.
Additional sources would need to add a domain or task-structure axis while preserving the derivation and scoring protocol.

\subsection{Dataset derivation}
\label{sec:dataset_asset_derivation}

The two single-target tasks are curated from larger source archives.
Table~\ref{tab:dataset_asset_derivation} summarizes the derivation rules.
Each source archive is screened once across all candidate stations or turbines under the same missingness policy: a strict target-gap rule plus a looser rule for the other input channels.
The retained station or turbine and segment are then fixed before model training.
Keeping one retained station or turbine per source archive is a tractability and validity choice rather than a performance-based search.
Targets from several stations or turbines would require synchronizing timestamps and aligning valid intervals across sites for every architecture, method, and scenario, which risks either substantial loss of history or leakage through alignment or imputation choices.
The retained histories have hourly series lengths comparable to the public benchmarks in Table~\ref{tab:benchmark_overview}: 24{,}038 and 25{,}867 for the curated datasets versus 17{,}420 and 17{,}544 for ETTh1 and Traffic.
After the retained segment and surviving channel set are fixed, remaining missing values are forward-filled only within that segment.
The curation balances data quality with fidelity to the source archive: long target outages and unusable gaps in the other input channels are excluded before modeling, while bounded source gaps remain part of the documented CPS data-quality setting.
Before forward fill, the retained target spans still contain gaps up to 24 h for Tiantan PM2.5 and 13 h 20 min for WT08 Power (kW).
Under the chronological split, target values missing in the source account for 1.34\% of \BeijingTiantan{} and 0.22\% of \PenmanshielWT{} forecast values across the available test windows, with 96 input steps and 6 forecast steps.
These missing target positions are fixed before model training and are not adjusted per model, so the task keeps the gap structure of the source while observed target values dominate test scoring.
After forward fill, leading timestamps with unresolved missing values are trimmed.
The preprocessing uses no backfill, interpolation, or model-dependent dataset selection.

\begin{table}[!htb]
  \centering
  \small
  \setlength{\tabcolsep}{4.5pt}
  \renewcommand{\arraystretch}{1.16}
  \caption{Derivation rules for the two curated single-target datasets.
  The station or turbine is fixed after applying the same screening policy to all candidates, with stricter screening of target gaps than of gaps in other input channels and no selection by model performance.}
  \label{tab:dataset_asset_derivation}
  \begin{tabularx}{\textwidth}{@{}
  >{\RaggedRight\arraybackslash}p{0.16\textwidth}
  >{\RaggedRight\arraybackslash}p{0.17\textwidth}
  X
  X
  @{}}
    \toprule
    Curated dataset & Retained window & Selection rule & Fill and preprocessing \\
    \midrule
    \BeijingTiantanShort{} &
    2014-06-03 10:00\newline to 2017-02-28 23:00 &
    Screen all stations under one policy: split at PM2.5 target gaps $>1$ day and at gaps $>3$ days in other input channels, then retain Tiantan because it has the longest surviving segment with acceptable similarity to the cross-station median PM2.5 series. &
    Keep local pollutant and weather inputs, encode \texttt{wd} as one discrete input channel, forward-fill only within the retained segment, trim leading timestamps with unresolved missing values, and use no backfill. \\
    \addlinespace[2pt]
    \PenmanshielWTShort{} &
    2016-08-18 15:00\newline to 2019-08-01 09:00 &
    Screen all turbines under the same preprocessing order: split at Power (kW) target gaps $>1$ day, derive each turbine's longest segment with a continuous target, drop the separate group of input channels with long gaps, and retain WT08 because it gives the longest hourly span while remaining close to the site-median Power (kW) series. &
    Remove SCADA totals, contractual indices, and controller target fields that risk leakage, forward-fill only within the retained segment, trim leading timestamps with unresolved missing values, aggregate complete hours as means of six 10-minute values, and keep the continuous channels that survive the gap screen on the retained segment. \\
    \bottomrule
  \end{tabularx}
\end{table}

These rules keep target-gap screening, screening of the other input channels, and leakage removal as separate preprocessing decisions.
Target-gap screening defines the candidate spans with a continuous target, while screening of the other input channels prevents their long outages from being forward-filled merely because the target remains within its gap threshold.
This matters for \BeijingTiantan{} because the PM2.5 target has no retained gap longer than 1 day, but the other input channels are still subject to the 3-day rule.
For \PenmanshielWT{}, SCADA totals, contractual indices, and controller target fields that risk leakage are removed before both screens.
For WT08, retained channels have no gap longer than 20 h, every dropped channel has a gap longer than 35 days, and the selected 21-day cutoff lies inside that empty interval.
Hourly WT08 aggregation averages six complete 10-minute values, so Power (kW) remains a period-average quantity on the retained span.
Both curated datasets are available as versioned Zenodo records under CC BY-SA 4.0:
\begin{center}
  \small
  \begin{tabular}{@{}l>{\RaggedRight\arraybackslash}p{0.62\textwidth}@{}}
    \BeijingTiantan{} & \BeijingTiantanZenodo{} \\
    \addlinespace[3pt]
    \PenmanshielWT{} & \PenmanshielZenodo{} \\
  \end{tabular}
\end{center}

All four datasets use chronological splits with 60\% training, 20\% validation, and 20\% test data.
Channels are standardized with training-set statistics only.
Benchmark perturbations act on the standardized input window while leaving forecast targets clean.
For \BeijingTiantan{}, the discrete wind-direction code \texttt{wd} is excluded from continuous-channel perturbations.
The remaining three datasets contain only continuous channels.

\FloatBarrier
\subsection{ETTh1 target scope and horizon ablation}
\label{subsec:etth1_task_ablation}

Within ETTh1, target scope changes the worst-scenario class more consistently than forecast horizon.
The four task settings combine all seven output channels or oil temperature alone with horizons of 96 or 6 steps.
Seven input channels and a 96-step history remain fixed, together with the eight scenarios and the rules for affected channels.
Each task uses one training and search seed, clean validation selection, and $K=10{,}000$ sampled test windows.

Table~\ref{tab:etth_task_results} shows that at horizon 96, changing from all seven targets to oil temperature moves the worst scenario of all seven architectures from the Availability class to Value.
At horizon 6, six architectures make this change, while for DLinear Attenuation is the worst scenario in both settings.
Changing the horizon alters one class when all channels are targets and none under oil temperature alone.
MSE is computed over each task's own targets and horizon, so raw errors across target scopes refer to different loss objects.

\begin{table}[!htb]
  \centering
  \small
  \setlength{\tabcolsep}{2.4pt}
  \renewcommand{\arraystretch}{1.10}
  \caption{Ablation of ETTh1 target scope and horizon.
  All four settings use seven input channels and a 96-step history.
  Lower values are better within each task, and bold marks the best unrounded score for each quantity and setting.
  MSE averages over the selected targets and forecast horizon, so different target scopes define different loss objects.}
  \label{tab:etth_task_results}
  \begin{tabular*}{\textwidth}{@{\extracolsep{\fill}}l*{2}{*{3}{S[table-format=1.3,detect-weight=true,mode=text]}l}@{}}
    \toprule
     & \multicolumn{4}{c}{Horizon 96} & \multicolumn{4}{c}{Horizon 6} \\
    \cmidrule(lr){2-5}\cmidrule(lr){6-9}
    Architecture & {$\mathrm{MSE}_c$} & {$\mathcal{D}_w$} & {$\mathrm{MSE}_w$} & Worst scenario & {$\mathrm{MSE}_c$} & {$\mathcal{D}_w$} & {$\mathrm{MSE}_w$} & Worst scenario \\
    \midrule
    \multicolumn{9}{@{}l}{\textit{All seven targets}} \\
    SeasonalNaive & 0.634 & 1.288 & 0.817 & MissingData & 0.465 & 1.642 & 0.764 & MissingData \\
    DLinear & 0.438 & 1.251 & \bfseries 0.548 & StuckSensor & 0.261 & 1.273 & 0.333 & Attenuation \\
    GRU & 0.664 & \bfseries 1.064 & 0.707 & MissingData & 0.354 & \bfseries 1.191 & 0.422 & MissingData \\
    ModernTCN & 0.462 & 1.216 & 0.562 & StuckSensor & 0.260 & 1.348 & 0.351 & MissingData \\
    TSMixer & 0.475 & 1.169 & 0.555 & StuckSensor & \bfseries 0.242 & 1.315 & \bfseries 0.319 & MissingData \\
    PatchTST & \bfseries 0.431 & 1.312 & 0.566 & StuckSensor & 0.243 & 1.345 & 0.327 & StuckSensor \\
    Chronos-2 & 0.496 & 1.448 & 0.718 & StuckSensor & 0.263 & 1.443 & 0.379 & StuckSensor \\
    \addlinespace[4pt]
    \multicolumn{9}{@{}l}{\textit{Oil temperature only}} \\
    SeasonalNaive & 0.132 & 1.973 & 0.261 & Noise & 0.021 & 7.498 & 0.160 & Noise \\
    DLinear & 0.106 & 1.770 & 0.188 & Attenuation & 0.018 & 6.189 & 0.112 & Attenuation \\
    GRU & 0.142 & \bfseries 1.404 & 0.200 & Attenuation & 0.022 & \bfseries 5.598 & 0.122 & Noise \\
    ModernTCN & \bfseries 0.103 & 1.822 & 0.188 & Noise & 0.019 & 6.076 & 0.115 & Noise \\
    TSMixer & 0.106 & 1.723 & \bfseries 0.182 & Attenuation & 0.018 & 6.041 & \bfseries 0.109 & Attenuation \\
    PatchTST & 0.111 & 1.757 & 0.195 & Attenuation & \bfseries 0.018 & 6.220 & 0.112 & Attenuation \\
    Chronos-2 & 0.119 & 1.689 & 0.200 & Attenuation & 0.019 & 5.971 & 0.112 & Attenuation \\
    \bottomrule
  \end{tabular*}
\end{table}

\FloatBarrier

\section{Detailed perturbation definitions}
\label{app:perturbation_definitions}

The scored benchmark protocol is defined by the evaluated scenario set, shared perturbation model, and uniform-severity sampling protocol.
The evaluated benchmark scenario set $\mathcal{P}$ is distinct from the training-only transfer fault families $\mathcal{P}_{\mathrm{trans}}$, which are available only to explicit fault-perturbation training protocols.

Real-world CPS time series are routinely affected by sensor faults.
A robust model should \enquote{maintain its level of performance under any circumstances} \citep{ISOIEC2021_TR24029_1} or \enquote{function correctly in the presence of invalid inputs or stressful environmental conditions} \citep{IEEEStandardGlossary1990}.
Modeling these failures as structured perturbations therefore yields CPS-grounded benchmark scenarios for forecasting robustness.

\subsection{Selection criteria}
\label{sec:selection_criteria}

The benchmark targets CPS time series collected through sensing pipelines.
The \textbf{Value} class is grounded in studies of sensor faults and process monitoring that report corrupted recorded values, including noisy readings, spikes, offset or gain errors, loss of accuracy, and sensor drift \citep{Balaban2009_IEEESensors_SensorFaultsAerospace,Ni2009_TOSN_SensorNetworkDataFaultTypes,Sharma2010_TOSN_SensorFaultPrevalence,Hansen2022_Water_DriftDetectionProcessTanks,Luca2023_AppliedSciences_DOSensorFaultDiagnosis}.
The \textbf{Timing} class is grounded in studies of channel alignment during sampling, timing in SCADA event logs, and timing of industrial data acquisition, including time-synchronous sampling, synchronization delays, delayed records, timestamp pathologies, poor timestamping, and delay variability \citep{Funck2014_Measurement_TimeSynchronousSampling,Andrade2022_IEEEAccess_SCADAAlarmAnomalyDetection,Kalappa2006_IEEE1588_TimeSynchronizationDataQuality,Schmetz2022_ProcediaCIRP_TimeSynchronizationProblem}.
The \textbf{Availability} class is grounded in studies of sensor networks, data quality, and data acquisition from programmable logic controllers that report missing values, stuck or fixed readings, and data loss during transfer \citep{Ni2009_TOSN_SensorNetworkDataFaultTypes,Sharma2010_TOSN_SensorFaultPrevalence,hubauerAnalysisDataQuality2013,jesusSurveyDataQuality2017,tehSensorDataQuality2020,Hijazi2024_Heliyon_PLCDataLossSynchronization}.
These input-side issues motivate benchmark scenarios that perturb the observed history available to the forecaster.
Using documented sensor and data-acquisition fault families is also consistent with broader robustness-evaluation calls for benchmarks and simulated environments that stress-test systems under unusual, extreme, or long-tail distribution shifts \citep{Hendrycks.etal_2022}.
The evaluated scenario set is therefore governed by two kinds of criteria.
Scenario-inclusion criteria determine whether a candidate fault mechanism is eligible for scoring:
\begin{enumerate}
    \item \textbf{CPS fault-family provenance.}
    The mechanism must directly abstract a fault family of CPS sensing or data acquisition that is documented in studies of sensor data or taxonomies of sensor failures, or identified by consulting domain experts.
    \item \textbf{Recurring deployment relevance.}
    The supporting evidence must indicate that the underlying fault family recurs in deployed sensing pipelines.
    This criterion requires neither an estimate of fault rates for a specific system nor exact reproduction of each installation.
    \item \textbf{Observed-history impact.}
    The mechanism must be able to change the input history available to the forecaster, without requiring a guaranteed difficulty ordering on every dataset.
\end{enumerate}
Implementation criteria determine whether an eligible mechanism can be scored under a common benchmark protocol:
\begin{enumerate}
    \item \textbf{Replayable input-side evaluation.}
    The benchmark scenario must be generated directly from recorded time-series observations and evaluated against the recorded forecasting target.
    \item \textbf{Model and simulator independence.}
    Scoring must not require access to model internals or a system-specific simulator.
    \item \textbf{Scalar severity profile.}
    One scalar severity parameter $s\in[0,1]$ must control the standardized stress profile, including affected-channel scope where applicable.
    \item \textbf{Scalable evaluation cost.}
    The scenario must be inexpensive enough for large-scale comparison.
\end{enumerate}
The replayability and simulator-independence criteria bound scoring to input-side sensor faults that can be evaluated on recorded histories with the original forecasting target.
They exclude target corruption and target-horizon sensor faults because those would change the quantity being forecast rather than the information available to the forecaster.
They also exclude process-level distribution shifts, cascading faults, and closed-loop effects when scoring them would require counterfactual targets from a digital twin or high-fidelity CPS simulator.
The criterion of a scalar severity profile supports comparability: endpoint selection is informed by reported or injected fault scales where applicable \citep{Balaban2009_IEEESensors_SensorFaultsAerospace,Ni2009_TOSN_SensorNetworkDataFaultTypes,Sharma2010_TOSN_SensorFaultPrevalence,Hansen2022_Water_DriftDetectionProcessTanks}, and the upper endpoint is treated as a stress-test condition for unusual, extreme, or long-tail shifts \citep{Hendrycks.etal_2022}.
Evaluating over $s\in[0,1]$ treats that endpoint as the upper limit of the severity profile rather than as the only tested condition, so robustness is measured over a range of fault intensities instead of depending on one isolated fault magnitude.

Together, these criteria separate scored benchmark scenarios from adjacent robustness tools.
Model-dependent adversarial attacks remain outside scoring because their perturbations are constructed against a model objective rather than from fixed sensor-fault mechanisms \citep{szegedyIntriguingPropertiesNeural2014,spechtGenerationAdversarialExamples2018,liu2023robust}.
Generic augmentation and generative data-synthesis methods remain outside scoring in SensorFault-Bench because the suitability of time-series augmentation depends on data, task, method, and validity choices beyond the selection criteria above \citep{iwanaEmpiricalSurveyData2021,wenTimeSeriesData2021,iglesiasDataAugmentationTechniques2023}.
The evaluated benchmark scenario set therefore contains eight scenarios, each corresponding to a fault mechanism that meets those criteria under the shared severity protocol.

\subsection{Benchmark scenario overview}
\label{sec:test_scenarios}

We group the evaluated perturbation families into Value, Timing, and Availability classes, and distinguish benchmark scenarios that target continuous sensors from scenarios that act on all channels jointly.
Table~\ref{tab:scenario_provenance} links the selected sensor-fault scenarios to the literature and reports recorded scales that guided the standardized construction where available.
Faults can also affect several channels together, as in the episode described by \citet{Ni2009_TOSN_SensorNetworkDataFaultTypes}, where low battery voltage affected every thermistor channel on one buoy.
The affected fraction was therefore 1 within that thermistor group.

\begin{table}[!htb]
  \centering
  \small
  \setlength{\tabcolsep}{3.5pt}
  \renewcommand{\arraystretch}{1.12}
  \caption{Provenance and standardized construction of the eight selected sensor-fault scenarios.
  Reported and injected scales retain their source units, whereas value perturbations are standardized with training statistics.}
  \label{tab:scenario_provenance}
  \begin{tabular}{@{}
    >{\RaggedRight\arraybackslash}p{0.155\linewidth}
    >{\RaggedRight\arraybackslash}p{0.415\linewidth}
    >{\RaggedRight\arraybackslash}p{\dimexpr0.43\linewidth-4\tabcolsep\relax}
    @{}}
    \toprule
    Scenario & Documented fault behavior or scale & Standardized construction \\
    \midrule
    Drift &
    Constant bias, with magnitudes of 1.2--60\% of nominal value compiled from the aerospace literature \citep{Balaban2009_IEEESensors_SensorFaultsAerospace}. &
    Positive constant offset over the history of each selected channel, up to 0.75 standardized units. \\
    \addlinespace
    Attenuation &
    Gain errors, with aerospace scaling factors of 0.3--0.7 \citep{Balaban2009_IEEESensors_SensorFaultsAerospace}. &
    Multiplication of selected channels by a factor decreasing from 1 to 0.25. \\
    \addlinespace
    Noise &
    Gaussian noise injected into real sensor traces at 0.5, 1.5 and 3 times the local standard deviation, over intervals from under 15 minutes to over six hours \citep{Sharma2010_TOSN_SensorFaultPrevalence}. &
    Gaussian noise over the history of each selected channel, with standard deviation up to 1 on standardized inputs. \\
    \addlinespace
    Spike &
    Spikes lasting one time step and isolated outliers in studies of sensor network faults \citep{Sharma2010_TOSN_SensorFaultPrevalence,Ni2009_TOSN_SensorNetworkDataFaultTypes}. &
    One positive impulse per selected channel at a sampled step, up to 7.5 standardized units. \\
    \addlinespace
    TimeStretch &
    Poor timestamping, communication delays and jitter in manufacturing acquisition \citep{Kalappa2006_IEEE1588_TimeSynchronizationDataQuality}. &
    Resampling over half the input window, with a factor increasing from 1 to 5. \\
    \addlinespace
    TimeCompress &
    Delayed entries, synchronization delays and clock pathologies in SCADA event logs \citep{Andrade2022_IEEEAccess_SCADAAlarmAnomalyDetection}. &
    Resampling over half the input window, with a factor decreasing from 1 to 0.1 and endpoint holding. \\
    \addlinespace
    StuckSensor &
    Sensors stuck at one value and anomalous constant readings \citep{Ni2009_TOSN_SensorNetworkDataFaultTypes,Sharma2010_TOSN_SensorFaultPrevalence}. &
    Forward fill on selected continuous channels, with separate windows that can span the whole input window. \\
    \addlinespace
    MissingData &
    Missing values associated with transmission and synchronization delays in an industrial acquisition pipeline \citep{Hijazi2024_Heliyon_PLCDataLossSynchronization}. &
    Forward fill of one synchronized gap across all channels, which can span roughly half of the input window. \\
    \bottomrule
  \end{tabular}
\end{table}

Let $\mathcal{P}$ denote the evaluated benchmark scenario set used for scoring, consisting, in fixed benchmark order, of Drift, Attenuation, Noise, Spike, TimeStretch, TimeCompress, StuckSensor, and MissingData.
Let $\mathcal{P}_{\mathrm{trans}}$ denote the set of training-only transfer fault families over continuous channels, adapted from the sensor failure taxonomy of \citet{brandtFaultsFeaturesPretraining2025} and extended with PacketLoss, consisting of LinearDrift, NonlinearDrift, Scaling, TimeVaryingScaling, TrimmingConstant, TrimmingVarying, and PacketLoss.
These two sets are disjoint in the benchmark protocol: $\mathcal{P}$ is scored at test time, whereas $\mathcal{P}_{\mathrm{trans}}$ is available only to local training protocols that intentionally inject explicit fault families.
Among the evaluated methods, $\mathcal{P}_{\mathrm{trans}}$ is used by the transfer variants of fault augmentation and not by the architecture comparison, post-hoc wrapper selection, or model selection.
Unstructured PGD adversarial training is separate from this taxonomy because it perturbs inputs by generic $L_\infty$ attacks rather than by explicit sensor-fault families.

Figure~\ref{fig:scenarios} summarizes the scored scenario suite and the benchmark score flow.
Table~\ref{tab:holdout_suite} separately summarizes $\mathcal{P}_{\mathrm{trans}}$ because it is not part of the scored benchmark.
Most evaluated scenarios target continuous sensors.
Within the \textbf{Value} class, Drift adds a constant offset and Attenuation multiplies the signal by a factor between 0 and 1, capturing sensor wear that gradually separates the measured value from the true one.
Other common value faults include noisy measurements, modeled as Gaussian Noise, and short-lived outliers, modeled as Spike.
Within the \textbf{Availability} class, StuckSensor freezes a channel for a period of time, while MissingData removes values across all channels, for example during maintenance.
Within the \textbf{Timing} class, TimeStretch and TimeCompress model sampling-rate changes that are not corrected during preprocessing, with TimeCompress also repeating the final observed value.
The evaluated benchmark $\mathcal{P}$ therefore balances realism and efficiency under the selection criteria above.
\begin{table}[!htb]
  \centering
  \small
  \setlength{\tabcolsep}{2.5pt}
  \renewcommand{\arraystretch}{1.16}
  \caption{Training-only transfer fault families $\mathcal{P}_{\mathrm{trans}}$ over continuous channels, adapted from the sensor failure taxonomy of \citet{brandtFaultsFeaturesPretraining2025} and extended with PacketLoss.
  Fault-augmentation variants draw from these families during training, while scoring uses only the disjoint evaluated scenario set $\mathcal{P}$.
  As in Table~\ref{tab:perturbation_parameters}, $s=0$ and $s=1$ denote benign and strongest endpoints with linear severity interpolation.
  Drift offsets and trimming bounds are on the standardized input scale, and lin., quad., damp., and cont.\ abbreviate linear, quadratic, damping, and continuation.}
  \label{tab:holdout_suite}
  \begin{tabularx}{\columnwidth}{@{}
    >{\RaggedRight\arraybackslash}p{0.108\columnwidth}
    >{\RaggedRight\arraybackslash}p{0.189\columnwidth}
    >{\RaggedRight\arraybackslash}X
    >{\RaggedRight\arraybackslash}p{0.197\columnwidth}
    >{\RaggedRight\arraybackslash}p{0.197\columnwidth}
    @{}}
    \toprule
    Fault group & Transfer family & Mechanism & Endpoint $s=0$ & Endpoint $s=1$ \\
    \midrule
    Drift
      & LinearDrift    & additive drift, linear in time & final offset $0$ & final offset $1.0$ \\
      & NonlinearDrift & linear and quadratic drift & lin.\ $0$, quad.\ $0$ & lin.\ $0.5$, quad.\ $0.5$ \\
    \addlinespace
    Scaling
      & Scaling            & uniform gain & multiplier $1.0$ & multiplier $2.0$ \\
      & TimeVaryingScaling & gain ramp, linear in time & final multiplier $1.0$ & final multiplier $2.0$ \\
    \addlinespace
    Trimming
      & TrimmingConstant & symmetric hard clamp & bound $3.0$ & bound $1.0$ \\
      & TrimmingVarying  & damped shrinkage beyond the bound & bound~$3.0$, damp.~$1.0$ & bound~$1.0$, damp.~$0.6$ \\
    \addlinespace
    Availability
      & PacketLoss & forward-filled local bursts, with one anchored start & start~$0$, cont.~$0$ & start~$0.25$, cont.~$0.9$ \\
    \bottomrule
  \end{tabularx}
\end{table}

\subsection{Forecasting setting}
\label{sec:formal_def_cps}

Before defining perturbations, we fix the notation used by the benchmark specification.
A CPS dataset is represented by a multivariate time series $\mathcal{T}\!=\!(\x_i)_{i=1}^N$ with $\x_i\in\R^m$, where $m=m_{\mathrm{cont}}+m_{\mathrm{disc}}$ splits into continuous and discrete channels.
Let $C_{\mathrm{tgt}}\subseteq\{1,\dots,m\}$ denote the designated forecast targets and let $m_{\mathrm{tgt}}=|C_{\mathrm{tgt}}|$.
For the all-channel datasets ETTh1 and Traffic, $C_{\mathrm{tgt}}=\{1,\dots,m\}$.
For the two single-target datasets \BeijingTiantan{} and \PenmanshielWT{}, $m_{\mathrm{tgt}}=1$.
Fix an input length $n$ and forecast horizon $n'$.
Let $\mathcal{I}=\{1,\dots,N-n-n'+1\}$ denote the valid start indices for input-output windows.
Each $I\in\mathcal{I}$ defines a sample $(X,Y)$ with input window $X=(\x_I,\ldots,\x_{I+n-1})\in\R^{n\times m}$ and future target window $Y=((\x_{I+n})_{C_{\mathrm{tgt}}},\ldots,(\x_{I+n+n'-1})_{C_{\mathrm{tgt}}})\in\R^{n'\times m_{\mathrm{tgt}}}$.
We write $X_{i,j}=\x_{I+i-1,j}$ for channel $j$ at input time step $i$.
A forecasting model is a measurable function $f:\R^{n\times m}\to\R^{n'\times m_{\mathrm{tgt}}}$.

The benchmark uses chronological train, validation, and test splits to avoid temporal leakage \citep{lopezdeprado2018advances,hewamalageForecastEvaluationData2023}.
The benchmark partitions $\mathcal{I}$ chronologically into train, validation, and test index sets.
Let $\mathcal{D}_{\mathrm{train}}$, $\mathcal{D}_{\mathrm{val}}$, and $\mathcal{D}_{\mathrm{test}}$ denote the corresponding finite sets of input-output samples.
When a finite split is used as a sampling source, its empirical distribution is uniform over the eligible samples in that split.
Sampling from the empirical test distribution therefore means drawing samples uniformly with replacement from $\mathcal{D}_{\mathrm{test}}$.

The benchmark standardizes each channel using training statistics so that perturbation magnitudes are comparable across sensors and datasets.
Let $\mu_{\mathrm{train}}\in\R^m$ and $\sigma_{\mathrm{train}}\in\R^m$ be the empirical mean and standard deviation of the training input data, and let $\mu_{\mathrm{train},\mathrm{tgt}}\in\R^{m_{\mathrm{tgt}}}$ and $\sigma_{\mathrm{train},\mathrm{tgt}}\in\R^{m_{\mathrm{tgt}}}$ denote the same statistics restricted to the target channels.
Then, we standardize each sample $(X,Y)$ as
\begin{equation*}
    \tilde{X} = \frac{X - \mu_{\mathrm{train}}}{\sigma_{\mathrm{train}}},\quad
    \tilde{Y} = \frac{Y - \mu_{\mathrm{train},\mathrm{tgt}}}{\sigma_{\mathrm{train},\mathrm{tgt}}},
\end{equation*}
where the operations are applied component-wise.
For ease of notation, all data are assumed standardized in what follows.

To quantify benchmark performance, we fix mean squared error on the designated forecast targets:
\begin{equation*}
    \mathrm{MSE}(f(X),Y)
    =
    \frac{1}{n' m_{\mathrm{tgt}}}
    \sum_{t=1}^{n'}
    \sum_{j=1}^{m_{\mathrm{tgt}}}
    \bigl(f(X)_{t,j}-Y_{t,j}\bigr)^2.
\end{equation*}
We assume that the expected target-channel MSE is finite for every evaluated forecasting model.
\BeijingTiantan{} and \PenmanshielWT{} therefore use single-target MSE, whereas ETTh1 and Traffic use all-channel MSE.
Methods may optimize different training objectives on $\mathcal{D}_{\mathrm{train}}$, but validation selection, robustness scoring, and reported fault-time errors are always computed from this target-channel MSE unless stated otherwise.

\subsection{Perturbation model}
\label{sec:perturbation_model}

The benchmark perturbs the observed input window $X$ while keeping the forecasting target $Y$ fixed.
At evaluation time, model weights are fixed, so the benchmark protocol measures robustness to input-side distribution shift rather than online adaptation.
Table~\ref{tab:perturbation_notation} summarizes the core symbols used in the perturbation model and the evaluated benchmark.

\begin{table}[!htb]
  \centering
  \small
  \setlength{\tabcolsep}{2.5pt}
  \renewcommand{\arraystretch}{1.14}
  \caption{Notation for the perturbation model and the evaluated benchmark.}
  \label{tab:perturbation_notation}
  \begin{tabular}{@{}l>{\RaggedRight\arraybackslash}p{0.7\columnwidth}@{}}
    \toprule
    Symbol & Meaning \\
    \midrule
    $\mathcal{P}$ & Evaluated benchmark scenario set used for scoring. \\
    $\mathcal{P}_{\mathrm{trans}}$ & Transfer fault families used only by protocols that train on explicit fault perturbations. \\
    $p\in\mathcal{P}$ & One evaluated benchmark scenario. \\
    $s\in[0,1]$ & Standardized severity level shared by evaluated scenarios. \\
    $\Theta_p$ & Parameter space for benchmark scenario $p$. \\
    $\theta_p^{\min},\theta_p^{\max}$ & Benign and strongest tested endpoint parameters for scenario $p$. \\
    $\theta_p(s)$ & Severity map for the perturbation parameter of scenario $p$. \\
    $k_p(s)$ & Number of affected channels at severity $s$ for scenarios scoped to channels. \\
    $X', X'_p$ & Perturbed input window, with $X'_p$ denoting the result under scenario $p$. \\
    $(M,Z)$ & Mask of affected entries and the remaining random draws of the scenario. \\
    $P_{M,Z\mid p,s}$ & Distribution of perturbation randomness for scenario $p$ at severity $s$. \\
    \bottomrule
  \end{tabular}
\end{table}

The scored benchmark is defined over the evaluated scenario set $\mathcal{P}$.
$\mathcal{P}_{\mathrm{trans}}$ appears only in explicit fault-perturbation training protocols, so the definitions below fix an evaluated benchmark scenario $p\in\mathcal{P}$ and a severity value $s\in[0,1]$.
Each scenario has its own parameter space $\Theta_p$ and severity map $\theta_p$, while all evaluated scenarios share the standardized severity domain $[0,1]$.
During scoring, each $p\in\mathcal{P}$ defines a separate scenario-wise evaluation condition.
For a fixed scenario, severity is sampled uniformly over $[0,1]$ and auxiliary randomness is drawn according to that scenario.
This yields one severity-averaged quantity per scenario $p$, which the worst-scenario quantities and the mean-case quantities in Appendix~\ref{app:metric_mapping} then aggregate over $\mathcal{P}$.
The scalar severity value has two roles.
It determines the scenario-specific perturbation parameter through $\theta_p(s)$.
For channel-scoped scenarios, the same severity value also determines the affected-channel count through the conditional distribution of the mask.
We first define the severity map for the perturbation parameter.
\begin{definition}[Severity map]\label{def:severity_map}
  For each benchmark scenario $p\in\mathcal{P}$, let $\Theta_p\subset\R^{d_p}$ be the set of parameters that define scenario $p$.
  Fix a benign endpoint and a strongest tested endpoint $\theta_p^{\min},\theta_p^{\max}\in\Theta_p$.
  A severity map is a measurable map $\theta_p:[0,1]\to\Theta_p$ with
  \begin{equation*}
    \theta_p(s) = \theta_p^{\min} + s (\theta_p^{\max}-\theta_p^{\min}).
  \end{equation*}
\end{definition}

At $s=0$ the perturbation parameter has no effect ($\theta_p^{\min}$).
At $s=1$ it reaches the strongest tested endpoint ($\theta_p^{\max}$).
The endpoint choices are scenario-specific modeling choices on the standardized data scale.
For channel-scoped scenarios, severity also changes the affected-channel count.
We encode this dependence in $P_{M,Z\mid p,s}$: the mask distribution varies with severity, and Section~\ref{subsec:perturbation_general_setup} defines the channel-count rule $k_p(s)$ used by the concrete scenarios.

Each concrete scenario definition fixes the eligible channels, the rule for the time window, the input transformation, and the random variables of the scenario.
Auxiliary randomness separates the affected-entry mask from any remaining random draws.

\begin{definition}[Auxiliary randomness]\label{def:aux_randomness}
  For a fixed benchmark scenario $p\in\mathcal{P}$ and severity value $s\in[0,1]$, draw auxiliary randomness $(M,Z)$ from the distribution $P_{M,Z\mid p,s}$ for that scenario and severity, independently of a sample $(X,Y)$.
  The mask $M\in \{0,1\}^{n\times m}$ indicates which entries of the input window are affected.
  The variable $Z$ collects any remaining scenario-specific random draws, such as Gaussian noise values.
  For deterministic scenarios, $Z$ is empty or degenerate.
\end{definition}

The Noise scenario illustrates the construction.
For $p=\mathrm{Noise}$, the endpoints are $\theta_p^{\min}=0$ and $\theta_p^{\max}=1$.
Under the channel-count rule in Section~\ref{subsec:perturbation_general_setup}, a mask realization $\bar{m}$ selects $k_{\mathrm{noise}}(s)$ affected continuous channels over the full input window.
Severity therefore controls two quantities: the noise standard deviation through $\theta_p(s)$ and the affected-channel count through the mask distribution.
Given $Z\sim\mathcal{N}(0,I)$, the perturbed input is
\begin{equation*}
    X' = X + \theta_p(s)\,(\bar{m}\odot Z).
\end{equation*}
At $s=0$, the added term vanishes.
At $s=1$, affected entries receive unit-scale Gaussian noise on the standardized data scale.

For any fixed benchmark scenario $p\in\mathcal{P}$ and fixed severity level $s\in[0,1]$, the corresponding scenario transformation changes only the observed input $X$.
The forecasting target remains the recorded future value $Y$ used by the original task.
The scenario definitions below therefore specify only the perturbed input $X'$.
The corresponding perturbed sample is $(X',Y)$.
The robustness score in Section~\ref{sec:robustness_score} and Appendix~\ref{ch:robustness_score} measures how much this input-side shift increases error.
Changing $Y$ would instead evaluate target corruption or require counterfactual targets for process-level or cascading shifts (Section~\ref{sec:selection_criteria}), rather than robustness to input-side sensor faults.
The sampled window $(X,Y)$ and the perturbation randomness indexed by $(p,s)$ are independent, so the same clean example can be paired with fresh severity and auxiliary draws.

This perturbation model meets the implementation criteria of replayable input-side evaluation, independence from models and simulators, and a scalar severity profile.
Whether evaluation cost scales depends on the concrete scenario definitions below.

\subsection{Concrete scenario definitions}
\label{sec:implementation_details_scenarios}

This subsection instantiates the benchmark scenarios from Section~\ref{sec:test_scenarios} using the forecasting setting from Section~\ref{sec:formal_def_cps} and the perturbation model from Section~\ref{sec:perturbation_model}.
Table~\ref{tab:perturbation_parameters_extended} gives the construction details behind the compact protocol summary in Table~\ref{tab:perturbation_parameters}.

\subsubsection{General setup}
\label{subsec:perturbation_general_setup}
Scenario-wise robustness profiles fix one benchmark scenario $p\in\mathcal{P}$ and sample severity uniformly over $[0,1]$.
The evaluated benchmark scenario set is fixed, with randomization only within each scenario.
$\mathcal{P}_{\mathrm{trans}}$ enters only when local training protocols with explicit fault perturbations are discussed.
Sequential or composed perturbation policies, for example AugMix-style compositions \citep{hendrycksAugMixSimpleData2020}, are outside the scored benchmark.

All datasets use the same scenario definitions, parameter intervals, severity sampling distribution, and channel-count rule as shown in Table~\ref{tab:perturbation_parameters_extended}, with normalization based on each dataset's training statistics (Section~\ref{sec:formal_def_cps}).
These common rules compare recurring effects in recorded values without assuming identical physical failure mechanisms or deployment distributions across sensor technologies.
Equal normalized severity therefore need not represent equal physical fault magnitude, nor need the shared scenarios cover every fault specific to a sensor technology.
Because the endpoints, the cap on affected channels, and the severity sampling distribution are explicit parameters of the formal definitions, they can be calibrated to the recorded faults of a specific system.
Such a calibrated variant is specific to that system and therefore lies outside the benchmark's default configuration.
Each benchmark scenario uses the severity map to interpolate its severity-controlled parameter between a benign configuration $\theta^{\min}$ and a strongest tested configuration $\theta^{\max}$ (see Definition~\ref{def:severity_map}).
The upper endpoint $\theta^{\max}$ is therefore a benchmark-defined stress-test setting.
It places scenario $p$ at the strongest tested configuration on the normalized input scale, not at an estimated field frequency or deployment-specific fault magnitude.
This deliberately strong input shift follows broader robustness-evaluation guidance for stress tests under unusual, extreme, or long-tail shifts \citep{Hendrycks.etal_2022}.
Where prior work provides usable ranges or controlled intensity grids, we treat them as scale anchors rather than calibration targets.
\citet{Balaban2009_IEEESensors_SensorFaultsAerospace} report such ranges for bias, scaling, drift, noise, and intermittent dropout, whereas \citet{Sharma2010_TOSN_SensorFaultPrevalence} study controlled intensities of short faults and noise faults together with the fractions of missing or faulty samples in the field.
NASA's Propulsion Diagnostic Method Evaluation Strategy benchmark defines magnitudes of engine sensor faults through continuous ranges scaled by $\sigma$ rather than estimates of deployment rates \citep{Simon2010_NASA_ProDiMES}.
Timing resampling endpoints remain benchmark-defined stress settings, not calibrated estimates of reported stretch or compression rates.

Channel-scoped perturbations affect a subset $S$ of the $m_{\mathrm{cont}}$ continuous channels (Table~\ref{tab:perturbation_parameters_extended}).
The benchmark fixes a maximum affected-channel fraction $\gamma_{\max}=0.5$ at $s=1$ and samples $S$ uniformly without replacement with size
\begin{equation*}
  k_p(s)=
  \begin{cases}
    0, & \text{if}\;\; s=0,\\
    1+\left\lfloor s\left(\left\lceil \gamma_{\max} m_{\mathrm{cont}}\right\rceil-1\right)\right\rfloor, & \text{if}\;\; s\in(0,1],
  \end{cases}
\end{equation*}
for a channel-scoped scenario $p$.
Thus every positive severity affects at least one continuous channel, while $s=1$ affects $\lceil \gamma_{\max} m_{\mathrm{cont}}\rceil$ channels.
The value $\gamma_{\max}=0.5$ is likewise a benchmark stress setting that keeps channel-scoped scenarios below all-channel outages, not a measured prevalence estimate for multi-sensor faults.
Sampling $s$ across $[0,1]$ therefore evaluates the full affected-channel profile, so $\gamma_{\max}$ sets the upper edge of the channel-count rule rather than the only affected-channel condition.
Section~\ref{subsec:channel_count_sensitivity} defines a sensitivity variant with a fixed fraction of selected channels that isolates this design choice.
For all-channel scenarios such as MissingData, no channel subset is sampled and all $m$ channels are affected.

Most benchmark scenarios affect a contiguous window $u=\{a,\dots,b\}\subseteq \{1,\dots,n\}$ of duration $l$ (Table~\ref{tab:perturbation_parameters_extended}).
The start index $a$ is sampled uniformly from $\{2,\dots,n-l+1\}$.
Excluding $i=1$ ensures that scenarios such as StuckSensor can reference the preceding value $X_{a-1,:}$.
The concrete scenario definitions below work directly with channel subsets and time windows rather than with the full binary mask whenever that yields simpler formulas.

The benchmark-facing perturbations shown in Figure~\ref{fig:scenarios} are formalized below.
Unless noted otherwise, all scored quantities below use only the evaluated scenario set $\mathcal{P}$ with severity sampled uniformly over $[0,1]$ within each scenario.
$\mathcal{P}_{\mathrm{trans}}$ is summarized in Table~\ref{tab:holdout_suite}.
\begin{table*}[!tb]
  \centering
  \small
  \setlength{\tabcolsep}{2.0pt}
  \renewcommand{\arraystretch}{1.12}
  \caption{Detailed construction parameters for the evaluated benchmark scenario set $\mathcal{P}$ summarized in Table~\ref{tab:perturbation_parameters}, including definition references and auxiliary random variables.
  For every channel-scoped scenario, the number of affected channels is $k_p(s)$ from the general setup above with $\gamma_{\max}=0.5$.
  MissingData always affects all $m$ channels.
  The duration is the number of time steps affected, while $n$ denotes the input length.
  $\theta$ is the parameter affected by the severity $s$.
  $Z$ lists auxiliary random variables beyond the mask, if present.}
  \vspace{7pt}
  \label{tab:perturbation_parameters_extended}
  \begin{tabular}{@{}c l l c c c c l@{}}
    \toprule
    Def. & Scenario & Scope & Duration & $\theta^{\min}$ & $\theta^{\max}$ & $Z$ & Effect at $s=1$ \\
    \midrule
    \ref{def:drift} & Drift & continuous & $n$ & 0 & 0.75 & -- & adds 0.75 std. deviation \\
    \ref{def:atten} & Attenuation & continuous & $n$ & 1 & 0.25 & -- & scales signal to 25\% \\
    \ref{def:noise} & Noise & continuous & $n$ & 0 & 1 & $\mathcal{N}(0,1)$ & adds noise with std. deviation 1 \\
    \ref{def:spike} & Spike & continuous & 1 & 0 & 7.5 & -- & adds 7.5 std. deviations once \\
    \addlinespace[2pt]
    \ref{def:stretch_compress} & TimeStretch & continuous & $\lceil n/2\rceil$ & 1 & 5 & -- & source step $0.2$, stretched \\
    \ref{def:stretch_compress} & TimeCompress & continuous & $\lceil n/2\rceil$ & 1 & 0.1 & -- & source step $10$, compressed \\
    \addlinespace[2pt]
    \ref{def:stuck} & StuckSensor & continuous & $\lceil\theta(n-1)\rceil$ & 0 & 1 & -- & frozen, forward-filled $n-1$ steps \\
    \ref{def:miss} & MissingData & all & $\lceil\theta(n-1)\rceil$ & 0 & 0.5 & -- & $\approx 50\%$ input gap, forward-filled \\
    \bottomrule
  \end{tabular}
\end{table*}

\subsubsection{Value perturbations}
For continuous sensor channels, studies of sensor faults and process monitoring document noise, spikes, calibration or gain errors, loss of accuracy, and drift \citep{Balaban2009_IEEESensors_SensorFaultsAerospace,Ni2009_TOSN_SensorNetworkDataFaultTypes,Sharma2010_TOSN_SensorFaultPrevalence,Hansen2022_Water_DriftDetectionProcessTanks,Luca2023_AppliedSciences_DOSensorFaultDiagnosis}.
The \textbf{Value} class collects these recorded-value perturbations.
Severity scales the additive or multiplicative parameter $\theta$ applied to the signal.

Drift adds a constant offset to the affected channels.

\begin{definition}[Drift]\label{def:drift}
Fix severity $s\in[0,1]$.
Let the drift offset be $\theta_{\mathrm{drift}}(s)$ with endpoints $\theta^{\min}=0$ and $\theta^{\max}=0.75$.
Sample a subset $S$ of size $k_{\mathrm{drift}}(s)$ from the $m_{\mathrm{cont}}$ continuous channels, with each affected channel spanning the whole input window.
For a sample $(X,Y)$, define the perturbed input $X'$ by
\begin{equation*}
X'_{i,j}=
  \begin{cases}
    X_{i,j} + \theta_{\mathrm{drift}}(s), & \text{if}\;\; j\in S,\\
    X_{i,j},  & \text{if}\;\; j \notin S,
  \end{cases}
  \quad \text{for}\;\; i=1,\dots,n.
\end{equation*}
\end{definition}

Deteriorated sensors can also exhibit amplitude-dependent distortion: high values and fast swings are measured less accurately than low-amplitude signals.
Attenuation simulates this effect by multiplying the affected channels by a factor below~1.

\begin{definition}[Attenuation]\label{def:atten}
Fix severity $s\in[0,1]$.
Let the attenuation factor be $\theta_{\mathrm{atten}}(s)$ with endpoints $\theta^{\min}=1$ and $\theta^{\max}=0.25$.
Sample a subset $S$ of size $k_{\mathrm{atten}}(s)$ from the $m_{\mathrm{cont}}$ continuous channels, with each affected channel spanning the whole input window.
For a sample $(X,Y)$, define the perturbed input $X'$ by
\begin{equation*}
  X'_{i,j}=
  \begin{cases}
    \theta_{\mathrm{atten}}(s)\,X_{i,j}, &  \text{if}\;\; j\in S,\\
    X_{i,j}, &  \text{if}\;\; j\notin S,
  \end{cases}
  \quad \text{for}\;\;  i=1,\dots,n.
\end{equation*}
\end{definition}

Every sensor measurement carries some noise floor.
A technical malfunction can raise this floor substantially.
Additive Gaussian noise models the effect.

\begin{definition}[Noise]\label{def:noise}
Fix severity $s\in[0,1]$.
Let the noise standard deviation be $\theta_{\mathrm{noise}}(s)$ with endpoints $\theta^{\min}=0$ and $\theta^{\max}=1$.
Sample a subset $S$ of size $k_{\mathrm{noise}}(s)$ from the $m_{\mathrm{cont}}$ continuous channels, with each affected channel spanning the whole input window.
Let $Z_{i,j}\sim\mathcal N(0,1)$ be independent across $i$ and $j$, and independent of $(X,Y)$ and of the sampled channel subset $S$.
For a sample $(X,Y)$, define the perturbed input $X'$ by
\begin{equation*}
  X'_{i,j}=
  \begin{cases}
    X_{i,j} + \theta_{\mathrm{noise}}(s) Z_{i,j}, &  \text{if}\;\; j\in S,\\
    X_{i,j}, &  \text{if}\;\; j\notin S,
  \end{cases}
  \quad \text{for}\;\;  i=1,\dots,n.
\end{equation*}
\end{definition}

Sensors occasionally produce outlier spikes, for example when an out-of-range reading is mapped to a binary key and then decoded as an extreme value.
Preprocessing can remove such outliers offline, but live systems often lack this step.
Spike adds a large additive impulse at a single random time step per affected channel.

\begin{definition}[Spike]\label{def:spike}
Fix severity $s\in[0,1]$.
Let the spike magnitude be $\theta_{\mathrm{spike}}(s)$ with endpoints $\theta^{\min}=0$ and $\theta^{\max}=7.5$.
Sample a subset $S$ of size $k_{\mathrm{spike}}(s)$ from the $m_{\mathrm{cont}}$ continuous channels.
For each $j\in S$, sample one time step $u_j\in \{2,\dots,n\}$, where $n$ denotes the length of the input time window.
For a sample $(X,Y)$, define the perturbed input $X'$ by
\begin{equation*}
    X'_{i,j}=
  \begin{cases}
    X_{i,j} + \theta_{\mathrm{spike}}(s), &  \text{if}\;\; i = u_j \;\;\text{and}\;\; j\in S\\
    X_{i,j}, &  \text{else},
  \end{cases}
  \quad \text{for}\;\;  i=1,\dots,n.
\end{equation*}
\end{definition}

\subsubsection{Timing perturbations}
\textbf{Timing} perturbations simulate errors in clock rate or synchronization of sampled channels that preprocessing has not corrected, so affected measurement channels are evaluated on a misaligned time base \citep{Funck2014_Measurement_TimeSynchronousSampling,Kalappa2006_IEEE1588_TimeSynchronizationDataQuality,Schmetz2022_ProcediaCIRP_TimeSynchronizationProblem}.
Related work on SCADA event logs documents synchronization delays, delayed records, and timestamp pathologies in operational records \citep{Andrade2022_IEEEAccess_SCADAAlarmAnomalyDetection}.
Each timing scenario first linearly interpolates the signal onto a continuous timeline, then traverses that timeline at a modified rate $\rho$ and maps the result back to the original discrete grid.
For example, consider five consecutive readings $(10,20,30,40,50)$.
Under the TimeStretch convention, setting $\rho=2$ advances through the source timeline in half steps, producing $(10,15,20,25,30)$ over the first five remapped positions.
A rate $\rho>1$ stretches the signal, while $\rho<1$ compresses it.
Unlike other perturbation classes, the affected time window has fixed length $|u|=\lceil n/2 \rceil$.
Severity instead scales the rate parameter $\rho$.

\newcommand{\Interp}{\operatorname{Interp}}

\begin{definition}[Linear interpolation with clipped endpoints]\label{def:interp}
Let $n\in\Nbb$, $x\in\R^{n}$, and $\tau\in\R$.
Set
\[
\bar{\tau}=\min\{n,\max\{1,\tau\}\},\qquad
a=\lfloor \bar{\tau}\rfloor,\qquad
b=\lceil \bar{\tau}\rceil,\qquad
\lambda=\bar{\tau}-a.
\]
Define
\[
\Interp(x;\tau)=(1-\lambda)\,x_a+\lambda\,x_b.
\]
This interpolation handles warped time indices that do not fall exactly on the integer grid.
\end{definition}

\begin{definition}[TimeStretch and TimeCompress]\label{def:stretch_compress}
Fix $p\in\{\mathrm{TimeStretch},\mathrm{TimeCompress}\}$ and severity $s\in[0,1]$.
Let $\rho=\theta_p(s)$, with $\rho\in[1,5]$ for $p=\mathrm{TimeStretch}$ and $\rho\in[0.1,1]$ for $p=\mathrm{TimeCompress}$.
Sample a subset $S$ of $k_p(s)$ continuous channels and a contiguous time window $u=\{a,\dots,b\}$ of fixed length $\lceil n/2\rceil$ with $a\ge 2$.
The perturbation replaces the selected segment by values from the same channel evaluated on a warped time grid.
For each $j\in S$ and $i=1,\dots,|u|$, define
\begin{equation*}
  X'_{a+i-1,j}
  =
  \Interp\!\left((X_{1,j},\dots,X_{n,j});\,a-1+\frac{i}{\rho}\right),
\end{equation*}
using linear interpolation with clipped endpoints (see Definition~\ref{def:interp}).
All entries outside the selected channels and time window remain unchanged.
Values $\rho>1$ traverse the source signal more slowly over the output grid and stretch the signal.
Values $\rho<1$ traverse it more quickly and compress the signal.
If a warped source index exceeds $n$, interpolation repeats the final observed value, so TimeCompress combines compression with endpoint holding.
\end{definition}

\subsubsection{Availability perturbations}
Studies of sensor faults and controlled fault injection document segments stuck at a constant value, loss of signal, and missing intervals in sensor data \citep{Balaban2009_IEEESensors_SensorFaultsAerospace,Ni2009_TOSN_SensorNetworkDataFaultTypes,Sharma2010_TOSN_SensorFaultPrevalence,deBruijn2016_SENSORNETS_BenchmarkFaultDatasets,Noshad2019_Sensors_RandomForestFaultDetection}.
Forward-filling with the last valid value is a common and simple strategy for handling gaps in live systems, so we adopt it as the default for the benchmark.
Other imputation methods or masking strategies are possible but out of scope here.
For \textbf{Availability} perturbations, severity controls the length of the affected window, $|u| = \lceil\theta_p(s)\,(n-1)\rceil$: higher severity means a larger fraction of the input carries no new information.

StuckSensor simulates a channel that freezes at its last valid reading.

\begin{definition}[StuckSensor]\label{def:stuck}
Fix severity $s\in[0,1]$.
Let the stuck duration fraction be $\theta_{\mathrm{stuck}}(s)$ with endpoints $\theta^{\min}=0$ and $\theta^{\max}=1$.
Sample a subset $S$ of size $k_{\mathrm{stuck}}(s)$ from the $m_{\mathrm{cont}}$ continuous channels.
For each affected channel $j\in S$, sample a time window $u_j=\{a_j,\dots,b_j\}$ with $a_j \ge 2$ of length $|u_j|=\lceil \theta_{\mathrm{stuck}}(s)\,(n-1)\rceil$, where $n$ denotes the length of the input window.
For a sample $(X,Y)$, define the perturbed input $X'$ by
\begin{equation}\label{eq:stuck}
  X'_{i,j}=
  \begin{cases}
    X_{a_j-1,j}, & \text{if}\;\;  j\in S,\ i\in u_j,\\
    X_{i,j}, & \text{else}.
  \end{cases}
\end{equation}
\end{definition}

When the entire system goes offline, for example during maintenance or power outages, all channels lose signal simultaneously.
MissingData simulates this by forward-filling a synchronized gap across all $m$ channels.
The gap is shorter than in StuckSensor because a system-wide outage removes more information per time step.

\begin{definition}[MissingData]\label{def:miss}
Fix severity $s\in[0,1]$.
Let the gap fraction be $\theta_{\mathrm{miss}}(s)$ with endpoints $\theta^{\min}=0$ and $\theta^{\max}=0.5$.
Sample a time window $u=\{a,\dots,b\}$ with $a \ge 2$ of length $|u|=\lceil \theta_{\mathrm{miss}}(s)\,(n-1)\rceil$, where $n$ denotes the length of the input window.
For a sample $(X,Y)$, define the perturbed input $X'$ by
\begin{equation}\label{eq:miss}
  X'_{i,j}=
  \begin{cases}
    X_{a-1,j}, & \text{if}\;\;  i\in u,\\
    X_{i,j}, & \text{else},
  \end{cases}
  \quad \text{for}\;\; j=1,\dots,m.
\end{equation}
\end{definition}

Both scenarios therefore present constant segments to the model, so their effects on the observed values overlap.
StuckSensor affects a subset of continuous channels in separately sampled windows, whereas MissingData affects all channels synchronously for a shorter duration.
Forward-filling gives the model no missingness indicator, so forecasters that consume an explicit mask of missing values remain untested.
Evaluating them would require the deployed data pipeline to detect gaps and pass that mask to the forecaster at inference time.
Such a mask would mark MissingData gaps, whereas a stuck sensor keeps reporting plausible values and would be marked only if a separate fault detector flagged it.

\section{Sample-level sensor-fault examples}
\label{app:sample_level_sensor_fault_examples}

\begin{figure}[!htb]
  \centering
  \includegraphics[width=1.0\textwidth]{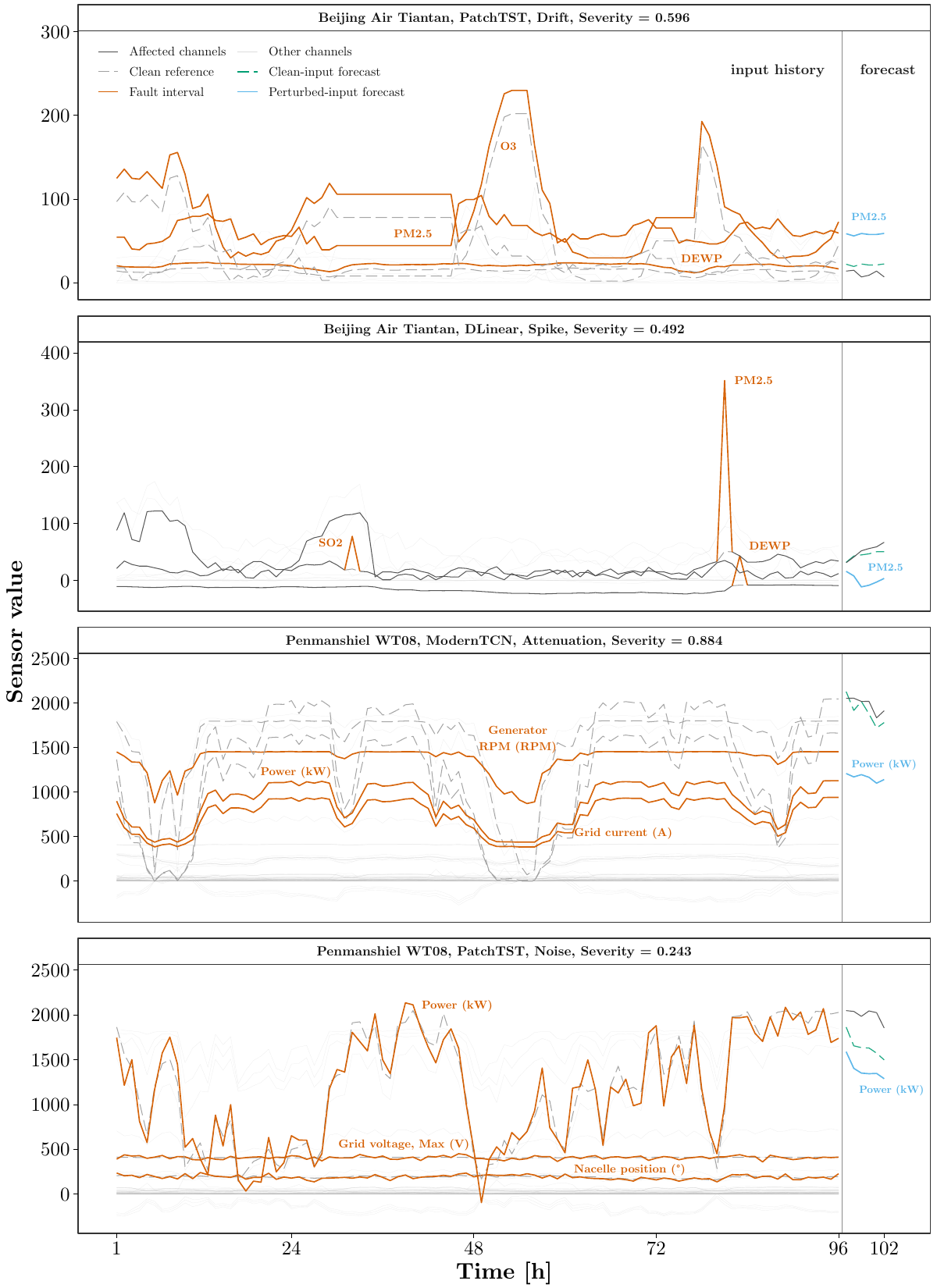}
  \caption{Sample-level sensor-fault examples for \BeijingTiantan{} and \PenmanshielWT{} under the value-fault scenarios Drift, Attenuation, Noise, and Spike.
  Each row shows one representative sample at one severity level for a forecasting architecture in the baseline setting.
  The vertical boundary separates the observed input history from the forecast horizon.}
  \label{fig:lines_small_data}
\end{figure}

\begin{figure}[!htb]
  \centering
  \includegraphics[width=1\textwidth]{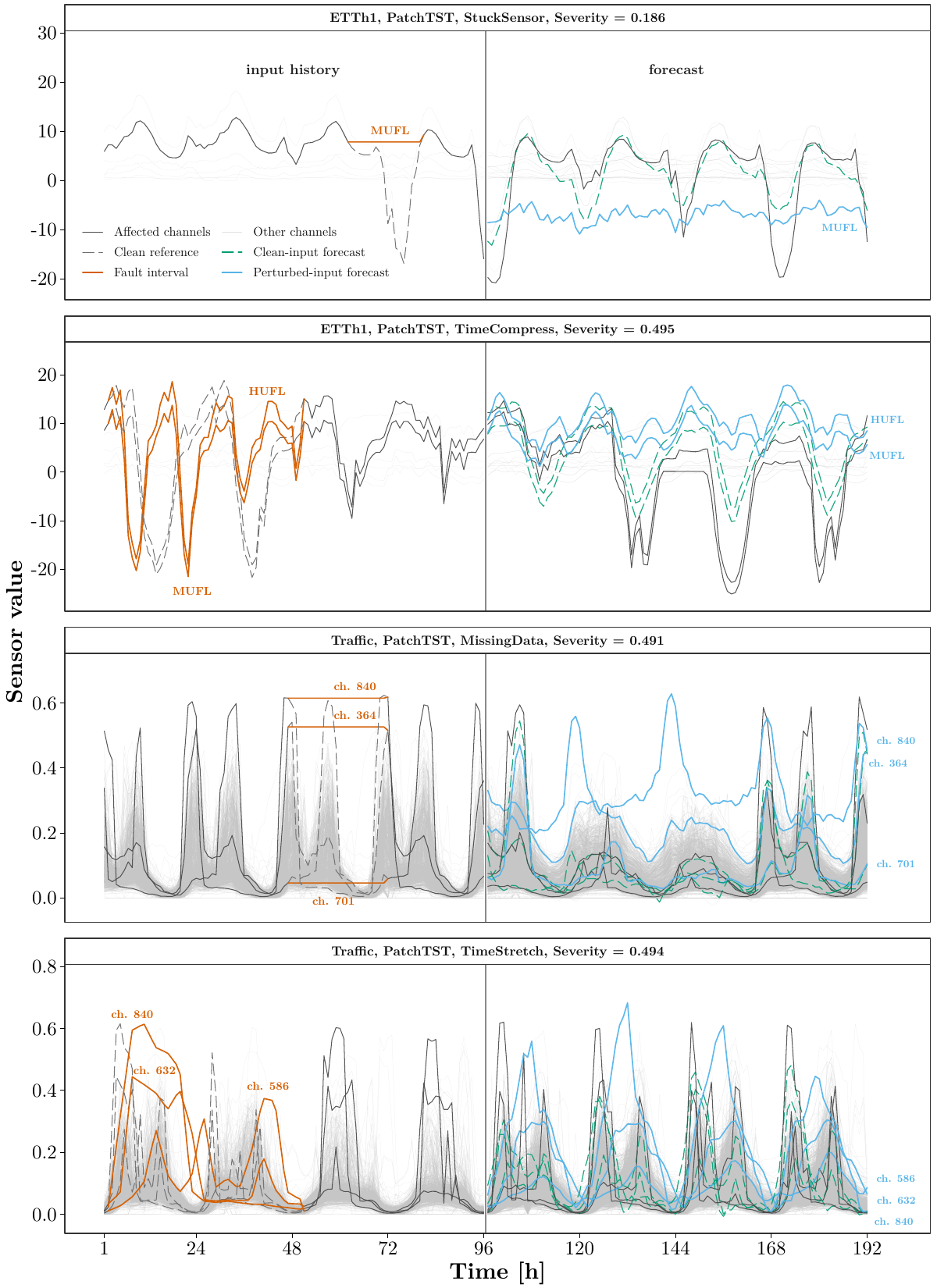}
  \caption{Sample-level sensor-fault examples for ETTh1 and Traffic under the timing and availability scenarios TimeStretch, TimeCompress, MissingData, and StuckSensor.
  Each row shows one representative PatchTST sample at one severity level.
  The vertical boundary separates the observed input history from the forecast horizon.}
  \label{fig:lines_large_data}
\end{figure}

Figures~\ref{fig:lines_small_data} and~\ref{fig:lines_large_data} visualize representative input and forecast windows from the four datasets under all eight scored benchmark scenarios.
Each panel overlays the clean input history, selected affected channels, and the forecasts from clean and perturbed inputs while keeping the forecast target fixed.
The examples are qualitative: each panel emphasizes up to three affected channels with the largest per-channel maximum absolute input perturbation over the observed input window, not channels chosen by forecast-error magnitude or manual inspection.
For \BeijingTiantan{}, CO and PRES are omitted from this display ranking because their absolute scale would otherwise dominate the display.

\newpage

\section{Robustness scores}
\label{ch:robustness_score}

The uniform-severity benchmark from Section~\ref{sec:robustness_score} evaluates every scenario explicitly and samples severity uniformly from $[0,1]$ within each scenario.
The corresponding Monte Carlo estimator, robustness quantities, and pairwise quantities are defined below, while score-family comparators are collected in Appendix~\ref{app:metric_mapping}.

\subsection{Scenario-level estimator under uniform severity sampling}
\label{sec:loss_shaping}

The benchmark scenarios formalized in Sections~\ref{sec:perturbation_model} and~\ref{sec:implementation_details_scenarios} are defined for all severity values $s\in[0,1]$.
At the population level, the scored benchmark evaluates each scenario explicitly and, within each scenario, averages perturbed MSE over severities sampled uniformly from $[0,1]$.
This yields one severity-averaged quantity per scenario.
Worst-scenario quantities take the maximum of these quantities over $\mathcal{P}$, and mean-case quantities take their uniform average, so neither aggregation acts on individual severity values.
The Monte Carlo estimator uses the same protocol, with one shared sample of test windows across scenarios.
Random sampling approximates the severity average, which could also be estimated using a deterministic severity grid.

\begin{definition}[Monte Carlo estimator under uniform severity]
  \label{def:mc_mse_and_degradation_estimators}
  Fix a forecasting model $f$ and a Monte Carlo budget of $K\in\mathbb{N}$ sampled test windows.
  Let $\hat P_{\mathrm{test}}$ denote the empirical distribution over the eligible held-out samples in $\mathcal{D}_{\mathrm{test}}$.
  Draw sampled test windows $(X^{(k)},Y^{(k)})$ from $\hat P_{\mathrm{test}}$ for $k=1,\dots,K$, that is, independently and uniformly with replacement from $\mathcal{D}_{\mathrm{test}}$.
  For each scenario $p\in\mathcal{P}$ and each sampled window $k=1,\dots,K$, draw $S_{p}^{(k)}\sim\mathrm{Unif}([0,1])$.
  Then draw auxiliary randomness $(M_{p}^{(k)},Z_{p}^{(k)})\sim P_{M,Z\mid p,S_p^{(k)}}$ (see Definition~\ref{def:aux_randomness}).
  Let $(X_{p}^{\prime(k)},Y^{(k)})$ denote the perturbed pair produced by scenario $p$ at severity $S_{p}^{(k)}$ under that auxiliary randomness.
  Define the clean and perturbed per-window MSE values
  \begin{equation*}
    \mathrm{MSE}_{c}^{(k)}(f)=\mathrm{MSE}\bigl(f(X^{(k)}),Y^{(k)}\bigr),
    \qquad
    \mathrm{MSE}_{p}^{(k)}(f)=\mathrm{MSE}\bigl(f(X_{p}^{\prime(k)}),Y^{(k)}\bigr).
  \end{equation*}
  The clean-MSE estimator is
  \begin{equation*}
    \widehat{\mathrm{MSE}}_c(f)=\frac{1}{K}\sum_{k=1}^{K}\mathrm{MSE}_{c}^{(k)}(f),
  \end{equation*}
  and, for each scenario $p\in\mathcal{P}$, the per-scenario perturbed MSE estimator is
  \begin{equation*}
    \widehat{\mathrm{MSE}}_{p}(f)=\frac{1}{K}\sum_{k=1}^{K}\mathrm{MSE}_{p}^{(k)}(f).
  \end{equation*}
  The corresponding scenario degradation estimator, defined when $\widehat{\mathrm{MSE}}_c(f)>0$, is
  \begin{equation*}
    \widehat{\mathcal{D}}_{p}(f)=\frac{\widehat{\mathrm{MSE}}_{p}(f)}{\widehat{\mathrm{MSE}}_c(f)}.
  \end{equation*}
\end{definition}

The benign limit $s\!=\!0$ remains part of each scenario definition, but continuous uniform severity sampling gives it probability zero.
The estimator therefore integrates over the full standardized severity range rather than over a single endpoint.
The scenario degradation score $\mathcal{D}_{p}(f)$ (Eq.~\eqref{eq:dp}) divides expected perturbed MSE by expected clean MSE, so the two expectations could in principle be estimated on different test windows.
The estimator instead computes clean and perturbed errors for every scenario on one shared sample of test windows.
Each perturbed input then differs from its clean counterpart only by the perturbation, and no scenario is scored on easier or harder windows than another.
The estimator $\widehat{\mathrm{MSE}}_{p}(f)$ makes the severity integration explicit: within each scenario $p$, the benchmark first averages perturbed MSE over severities sampled uniformly from $[0,1]$, then compares the resulting scenario-wise quantities, and only afterward takes a maximum or uniform average over scenarios.
Dividing by $\widehat{\mathrm{MSE}}_c(f)$ yields a dimensionless degradation factor intrinsic to the evaluated model.

\subsection{Worst-scenario score and pairwise quantities}
\label{sec:robustness_scores}

Common-corruption benchmarks often pair a corrupted-performance measure with a relative-degradation measure \citep{hendrycksBenchmarkingNeuralNetwork2019, Michaelis.etal_2019}.
This benchmark keeps that split, but identifies the relevant operating condition through the worst benchmark scenario rather than through a global mean.

\begin{definition}[Worst-scenario degradation]\label{def:worst_scenario_degradation}
  Let $\widehat{\mathcal{D}}_{p}(f)$ denote the scenario degradation estimators (see Definition~\ref{def:mc_mse_and_degradation_estimators}), and assume $\widehat{\mathrm{MSE}}_c(f)>0$.
  Define the worst-scenario degradation score of $f$ by
  \begin{equation*}
    \widehat{\mathcal{D}}_w(f)
    =
    \max_{p\in\mathcal{P}}\widehat{\mathcal{D}}_{p}(f).
  \end{equation*}
  If the identity of a worst scenario must be reported, exact ties are broken by the fixed benchmark scenario order.
\end{definition}

A single operational fault mode can dominate degradation.
Worst-scenario scoring therefore emphasizes the benchmark scenario with the largest degradation.
The maximum is an evaluation aggregation rather than a training objective.
It prevents low mean degradation across scenarios from masking a severe failure in one scenario.
Related evaluation work likewise estimates performance under worst-case subpopulations or shifted distributions \citep{liEvaluatingModelPerformance2021, Subbaswamy.etal_2021}, and broader uniform-performance and distributionally robust viewpoints motivate performance under distributional perturbations \citep{Duchi.Namkoong_2021, Dziugaite.etal_2020}.

Averaging over the full standardized severity range within each scenario is equally deliberate.
Scoring only at maximal severity $s=1$ would make the result depend too strongly on the endpoint choice $\theta_p^{\max}$.
Uniform severity sampling over $[0,1]$ instead captures how rapidly performance deteriorates across the full severity profile.
Two models can look similar at $s=1$ and still differ materially in where that deterioration begins.

\begin{definition}[Worst-scenario fault-time MSE]\label{def:worst_scenario_fault_time_mse}
  Define the reported worst-scenario fault-time MSE of $f$ by
  \begin{equation*}
    \widehat{\mathrm{MSE}}_w(f)
    =
    \max_{p\in\mathcal{P}}\widehat{\mathrm{MSE}}_{p}(f).
  \end{equation*}
\end{definition}
For a fixed model, any scenario maximizing degradation also maximizes perturbed MSE because $\widehat{\mathrm{MSE}}_c(f)$ is constant across scenarios.
Worst-scenario fault-time MSE therefore reports absolute perturbed error at the same model-specific condition that defines worst-scenario degradation, not under one common scenario for all models.

\begin{definition}[Mean-case degradation]\label{def:mean_case_degradation}
  For any forecasting model $f$ with $\widehat{\mathrm{MSE}}_c(f)>0$, define the mean corrupted MSE estimator and the corresponding mean degradation estimator by
  \begin{equation*}
    \widehat{\overline{\mathrm{MSE}}}(f)=\frac{1}{|\mathcal{P}|}\sum_{p\in\mathcal{P}}\widehat{\mathrm{MSE}}_{p}(f)
    \qquad \text{and} \qquad
    \widehat{\overline{\mathcal{D}}}(f)=\frac{\widehat{\overline{\mathrm{MSE}}}(f)}{\widehat{\mathrm{MSE}}_c(f)}=\frac{1}{|\mathcal{P}|}\sum_{p\in\mathcal{P}}\widehat{\mathcal{D}}_{p}(f).
  \end{equation*}
\end{definition}

In a method-baseline comparison, $\Delta\widehat{\mathcal{D}}_w$ and $\Delta\widehat{\overline{\mathcal{D}}}$ differ only in the scenario aggregation used inside each model's degradation estimate.
The worst-scenario delta subtracts $\widehat{\mathrm{MSE}}_w(f_b)/\widehat{\mathrm{MSE}}_c(f_b)$ from $\widehat{\mathrm{MSE}}_w(f')/\widehat{\mathrm{MSE}}_c(f')$, whereas the mean-case delta replaces $\widehat{\mathrm{MSE}}_w$ with $\widehat{\overline{\mathrm{MSE}}}$ in both terms.
Each degradation delta therefore asks whether the robustness-improved variant lowers the perturbed-to-clean error ratio relative to the paired baseline forecasting model.

\begin{definition}[Pairwise reporting deltas]\label{def:pairwise_reporting_deltas}
  Fix one dataset and one forecasting architecture.
  Let $f_b$ denote the evaluated no-intervention forecasting model and let $f'$ denote the evaluated forecasting variant produced by applying one robustness-improvement method to that same architecture.
  For this method-baseline pair, define
  \begin{equation*}
    \Delta\widehat{\mathcal{D}}_w(f',f_b)
    =
    \widehat{\mathcal{D}}_w(f')-\widehat{\mathcal{D}}_w(f_b),
  \end{equation*}
  \begin{equation*}
    \Delta\widehat{\overline{\mathcal{D}}}(f',f_b)
    =
    \widehat{\overline{\mathcal{D}}}(f')-\widehat{\overline{\mathcal{D}}}(f_b),
  \end{equation*}
  \begin{equation*}
    \Delta\widehat{\mathrm{MSE}}_c(f',f_b)
    =
    \widehat{\mathrm{MSE}}_c(f')-\widehat{\mathrm{MSE}}_c(f_b),
  \end{equation*}
  and the worst-scenario fault-time delta
  \begin{equation*}
    \Delta\widehat{\mathrm{MSE}}_w(f',f_b)
    =
    \widehat{\mathrm{MSE}}_w(f')-\widehat{\mathrm{MSE}}_w(f_b).
  \end{equation*}
  For mean-case sensitivity, also define
  \begin{equation*}
    \Delta\widehat{\overline{\mathrm{MSE}}}(f',f_b)
    =
    \widehat{\overline{\mathrm{MSE}}}(f')-\widehat{\overline{\mathrm{MSE}}}(f_b).
  \end{equation*}
\end{definition}

The reported method comparisons use $\Delta\widehat{\mathcal{D}}_w$ together with $\Delta\widehat{\mathrm{MSE}}_c$ and $\Delta\widehat{\mathrm{MSE}}_w$.
The corresponding mean-case method sensitivity reports $\Delta\widehat{\overline{\mathcal{D}}}$ together with $\Delta\widehat{\overline{\mathrm{MSE}}}$.
Table~\ref{tab:reported_quantities} summarizes the primary result quantities and mean-case sensitivity quantities used by the result tables.

\begin{table*}[!tb]
  \centering
  \small
  \setlength{\tabcolsep}{2.5pt}
  \renewcommand{\arraystretch}{1.15}
  \caption{Primary result quantities and mean-case sensitivity quantities.
  For quantities that are not deltas, lower values are better.
  For method-baseline deltas, negative values favor the robustness-improved variant.}
  \vspace{7pt}
  \label{tab:reported_quantities}
  \begin{tabularx}{0.98\textwidth}{@{}>{\RaggedRight\arraybackslash}p{0.18\textwidth}>{\RaggedRight\arraybackslash}p{0.22\textwidth}>{\RaggedRight\arraybackslash}X@{}}
    \toprule
    Quantity & Role & Reading \\
    \midrule
    $\mathcal{D}_w(f)$ & Relative robustness & Largest ratio of perturbed to clean MSE across scenarios. \\
    $\mathrm{MSE}_c(f)$ & Clean error & Clean forecasting error under intact sensing. \\
    $\mathrm{MSE}_w(f)$ & Fault-time error & Absolute fault-time error at the model's own worst scenario. \\
    $\Delta\mathcal{D}_w(f',f_b)$ & Method-baseline delta & Change in worst-scenario degradation. \\
    $\Delta\mathrm{MSE}_c(f',f_b)$ & Method-baseline delta & Change in clean forecasting error. \\
    $\Delta\mathrm{MSE}_w(f',f_b)$ & Method-baseline delta & Change in worst-scenario fault-time error. \\
    $\overline{\mathcal{D}}(f)$, $\overline{\mathrm{MSE}}(f)$ & Sensitivity & Counterparts to $\mathcal{D}_w$ and $\mathrm{MSE}_w$ that average over scenarios. \\
    $\Delta\overline{\mathcal{D}}(f',f_b)$ & Sensitivity & Change in mean degradation. \\
    $\Delta\overline{\mathrm{MSE}}(f',f_b)$ & Sensitivity & Change in mean corrupted MSE. \\
    \bottomrule
  \end{tabularx}
\end{table*}

\subsubsection{How to read the fault-time error}
\label{subsec:mse_pstar_reading}

For a fixed model, $\mathrm{MSE}_w$ is the maximum perturbed MSE over the evaluated benchmark scenario set (see Definition~\ref{def:worst_scenario_fault_time_mse}).
It reports the absolute error at that model's own worst benchmark scenario, with severity averaged inside each scenario.
Because the maximizing scenario can differ across architectures or methods, $\mathrm{MSE}_w$ is model-specific rather than a common-scenario error.
Read alongside $\mathcal{D}_w$, it separates worst-scenario degradation from worst-scenario fault-time error at the same model-specific scenario.
\citet{taoriMeasuringRobustnessNatural2020} make the corresponding clean-control point for distribution shifts: better shifted performance can reflect better standard performance rather than lower shift sensitivity.
Here, $\mathcal{D}_w$ is the clean-controlled degradation score, so the three method deltas separate changes in worst-scenario degradation from changes in clean MSE and worst-scenario fault-time MSE.
Traffic in Table~\ref{tab:main_results} makes the distinction concrete: PatchTST has the lowest clean MSE ($\mathrm{MSE}_c=0.453$) and high worst-scenario degradation ($\mathcal{D}_w=1.676$), but its worst-scenario fault-time MSE ($\mathrm{MSE}_w=0.759$) remains far below SeasonalNaive's ($\mathrm{MSE}_w=1.537$).
SeasonalNaive has lower worst-scenario degradation ($\mathcal{D}_w=1.210$) because its clean MSE is much larger ($\mathrm{MSE}_c=1.270$), so low relative fault sensitivity does not imply low worst-scenario fault-time error.
The practical reading is to use $\mathcal{D}_w$ for relative fault sensitivity, $\mathrm{MSE}_w$ for model-specific absolute error at the worst benchmark scenario, and per-scenario MSE for a fixed-scenario decision.
A fixed-scenario comparison should therefore fix the dataset and benchmark scenario before comparing the corresponding per-scenario errors directly.

For deployment comparisons, we prioritize $\mathrm{MSE}_w$ when assessing error magnitude and read it together with degradation and clean MSE.
Worst-scenario degradation remains the benchmark's score of relative robustness.
Deployment decisions additionally require errors resolved by channel and forecast horizon, with limits that reflect consequences and tail risk.
As aggregates of squared error, the three quantities also do not show how residuals are distributed over time.

\subsubsection{Sampling and Monte Carlo evaluation}
Algorithm~\ref{alg:benchmark_loop} summarizes the estimator together with validation-only model selection and reporting (see Definition~\ref{def:mc_mse_and_degradation_estimators}).
For each selected model, it evaluates clean MSE once on the shared sampled test-window set, then traverses the benchmark scenarios in fixed order with fresh severity and auxiliary randomness for each scenario-sample pair.

\begin{algorithm}[!htb]
  \caption{Monte Carlo evaluation under uniform severity sampling.}
  \label{alg:benchmark_loop}
  \small
  \begin{algorithmic}[1]
    \State For each dataset, identify the models selected on validation data for the architecture and method comparisons.
    \State Fix one selected forecasting model $f$.
    \State Draw test windows $\{(X^{(k)},Y^{(k)})\}_{k=1}^{K}$ independently with replacement from $\mathcal{D}_{\mathrm{test}}$.
    \For{$k=1,\dots,K$}
      \State Compute $\mathrm{MSE}_{c}^{(k)}(f)$ on $(X^{(k)},Y^{(k)})$ once.
      \For{each benchmark scenario $p\in\mathcal{P}$ in fixed order}
        \State Draw $S_{p}^{(k)}\sim\mathrm{Unif}([0,1])$.
        \State Draw the auxiliary randomness $(M,Z)$ of the perturbation conditional on $p$ and $S_{p}^{(k)}$.
        \State Apply $p$ at severity $S_{p}^{(k)}$ to $(X^{(k)},Y^{(k)})$ to obtain $(X_p'^{(k)},Y^{(k)})$.
        \State Compute $\mathrm{MSE}_{p}^{(k)}(f)$ on $(X_p'^{(k)},Y^{(k)})$.
      \EndFor
    \EndFor
    \State Estimate $\widehat{\mathrm{MSE}}_c(f)$ and $\widehat{\mathrm{MSE}}_{p}(f)$ for all $p\in\mathcal{P}$ by averaging the sampled values.
    \State Compute the scenario degradation estimates $\widehat{\mathcal{D}}_{p}(f)$ and worst-scenario degradation $\widehat{\mathcal{D}}_w(f)$, using the fixed scenario order only for exact ties.
    \State Report $\widehat{\mathcal{D}}_w(f)$ together with $\widehat{\mathrm{MSE}}_c(f)$ and $\widehat{\mathrm{MSE}}_w(f)$.
    \State For each method-baseline pair $(f',f_b)$, report $\Delta\widehat{\mathcal{D}}_w(f',f_b)$, $\Delta\widehat{\mathrm{MSE}}_c(f',f_b)$, and $\Delta\widehat{\mathrm{MSE}}_w(f',f_b)$.
    \State For the mean-case sensitivity of methods, report $\Delta\widehat{\overline{\mathcal{D}}}(f',f_b)$ together with $\Delta\widehat{\overline{\mathrm{MSE}}}(f',f_b)$.
  \end{algorithmic}
\end{algorithm}

Tables that compare architectures report $\widehat{\mathcal{D}}_w(f)$, $\widehat{\mathrm{MSE}}_c(f)$, and $\widehat{\mathrm{MSE}}_w(f)$.
For evaluations of robustness-improvement methods, $\Delta\widehat{\mathcal{D}}_w(f',f_b)$ gives the clean-controlled change in worst-scenario degradation, while $\Delta\widehat{\mathrm{MSE}}_c(f',f_b)$ and $\Delta\widehat{\mathrm{MSE}}_w(f',f_b)$ keep the nominal and worst-scenario fault-time error movements explicit.
The corresponding mean-case sensitivity reports $\overline{\mathcal{D}}$ and $\overline{\mathrm{MSE}}$ for architecture comparisons, and $\Delta\widehat{\overline{\mathcal{D}}}$ together with $\Delta\widehat{\overline{\mathrm{MSE}}}$ for method-baseline comparisons.
We report these quantities as $\mathrm{MSE}_c$, $\mathrm{MSE}_w$, $\overline{\mathrm{MSE}}$, and their corresponding deltas where applicable.
Models are selected using validation data only, and the test split is reserved for final evaluation.

\section{Benchmark design sensitivity checks}
\label{sec:benchmark_design_validation}
We examine how benchmark design choices affect model rankings and the interpretation of fault sensitivity.
Varying evaluation sampling or fault construction keeps the selected models of Table~\ref{tab:main_results} fixed, whereas changing the validation selector can choose different models from the same trained candidates.
The fault variants alter the number of affected channels, severity weighting, perturbation direction, or timing transformation, while the endpoints $\theta_p^{\max}$ and the rules that set the length and onset of fault windows keep their default settings.
These comparisons test which degradation leaders and worst-scenario classes persist, and where absolute error rankings or small method effects change.
Appendix~\ref{subsec:training_seed_sensitivity} separately examines variation from repeated training and search.

\subsection{Sensitivity to the evaluation data seed}
\label{subsec:eval_data_seed_sensitivity}

The evaluation data seed is a testing-only seed.
Changing it leaves the trained winner pool, temporal split, validation selector, evaluated scenario set, severity sampling distribution, and the budget of $K=10{,}000$ sampled test windows fixed.
It redraws the sampled test-window multiset and the random variables used when each benchmark scenario is applied, including severity and scenario-specific perturbation randomness.
By comparison, the architecture bootstrap resamples the realized window pool with its existing fault draws, so its intervals describe conditional variability within that pool.
The comparison uses five evaluation realizations: the canonical realization that uses each dataset's canonical data seed, plus explicit evaluation data seeds 0, 1, 2, and 3.
The comparison is a Monte Carlo stress test for the three reported quantities under the fixed benchmark setup.
The~check is not an estimate of retraining variability.
Table~\ref{tab:eval_data_seed_sensitivity} uses three stability diagnostics.
Absolute-shift columns report mean and largest absolute movement from the canonical realization over the seven baseline architectures and four explicit seeds, separately for $\mathcal{D}_w$ and $\mathrm{MSE}_w$.
Rank-correlation columns report the minimum pairwise Spearman rank correlation across the five realizations after ranking the seven baseline architectures separately by each fault-facing quantity.
This rank diagnostic differs from effective robustness $\overline{\rho}$ in the style of \citet{taoriMeasuringRobustnessNatural2020} in Appendix~\ref{app:metric_mapping}.
Worst-scenario agreement reports the minimum agreement of exact scenarios and of scenario classes across explicit seeds.
Smaller absolute shifts and larger agreement values indicate stronger stability across evaluation realizations.

Table~\ref{tab:eval_data_seed_sensitivity} shows that the two fault-facing reported quantities are qualitatively stable but not numerically invariant across evaluation data seeds.
Average absolute shifts are small, with the largest mean movements in both quantities on \BeijingTiantan{}: $0.037$ for $\mathcal{D}_w$ and $0.016$ for $\mathrm{MSE}_w$, compared with $0.009$ for Traffic's $\mathrm{MSE}_w$.
The largest single shifts are likewise concentrated on \BeijingTiantan{} for $\mathcal{D}_w$ and Traffic for $\mathrm{MSE}_w$.
Clean-MSE movements have the same small absolute scale: their mean absolute movement is at most $0.019$ and their largest single movement is $0.032$, with only the near tie between ModernTCN and DLinear on \PenmanshielWT{} changing the clean-MSE winner.
Rank correlations remain high across the five realizations, with the lowest value on the \BeijingTiantan{} $\mathrm{MSE}_w$ ranking.
Worst-scenario identities, which determine both $\mathcal{D}_w$ and $\mathrm{MSE}_w$, are less stable than scenario-class labels: three architectures change exact worst scenario on \BeijingTiantan{}, one changes on \PenmanshielWT{} and ETTh1, and only one \BeijingTiantan{} change crosses the scenario-class boundary.
Exact scenario labels should therefore be interpreted more cautiously than scenario-class labels under this budget of sampled test windows.
The corresponding method-delta recomputations also preserve the sign patterns for the three tracked methods except for two entries near zero whose sign changes: RevIN's $\Delta\mathrm{MSE}_c$ on \BeijingTiantan{} and PGD adversarial training's $\Delta\mathcal{D}_w$ on ETTh1.
The evaluation data seed check therefore leaves the qualitative architecture and tracked-method patterns intact across the three reported quantities under the fixed evaluation setup, while leaving exact score values, retraining, data-split, and model-seed variability outside scope.
Appendix~\ref{subsec:training_seed_sensitivity} evaluates the variability from retraining with different training and search seeds separately.

\begin{table*}[!tb]
  \centering
  \small
  \setlength{\tabcolsep}{2.5pt}
  \renewcommand{\arraystretch}{1.10}
  \caption{Evaluation data seed sensitivity for the baseline architecture comparison.
  Entries summarize score movement, rank stability, and worst-scenario identity stability across the canonical realization and explicit evaluation data seeds 0, 1, 2, and 3.
  Rank correlations are minimum pairwise Spearman values for the two fault-facing architecture rankings.}
  \vspace{7pt}
  \label{tab:eval_data_seed_sensitivity}
  \begin{tabularx}{\textwidth}{@{}l
    *{2}{>{\hsize=1.02\hsize\centering\arraybackslash}X}
    *{2}{>{\hsize=1.12\hsize\centering\arraybackslash}X}
    *{2}{>{\hsize=0.86\hsize\centering\arraybackslash}X}
    p{13pt}c@{\hspace{1.8em}}cp{13pt}@{}}
    \toprule
    & \multicolumn{2}{c}{Mean absolute shift}
    & \multicolumn{2}{c}{Largest absolute shift}
    & \multicolumn{2}{c}{Rank correlation}
    & \multicolumn{4}{c}{Worst-scenario agreement} \\
    \cmidrule(lr){2-3}\cmidrule(lr){4-5}\cmidrule(lr){6-7}\cmidrule(l){8-11}
    Dataset
    & $\mathcal{D}_w$ & $\mathrm{MSE}_w$
    & $\mathcal{D}_w$ & $\mathrm{MSE}_w$
    & $\mathcal{D}_w$ & $\mathrm{MSE}_w$
    & & Exact & Class & \\
    \midrule
    \BeijingTiantan{} & $0.037$ & $0.016$ & $0.106$ & $0.038$ & $0.964$ & $0.857$ & & $4/7$ & $6/7$ & \\
    \PenmanshielWT{} & $0.011$ & $0.004$ & $0.041$ & $0.012$ & $0.929$ & $0.964$ & & $6/7$ & $7/7$ & \\
    ETTh1 & $0.004$ & $0.003$ & $0.015$ & $0.013$ & $1.000$ & $0.964$ & & $6/7$ & $7/7$ & \\
    Traffic & $0.008$ & $0.009$ & $0.062$ & $0.049$ & $1.000$ & $1.000$ & & $7/7$ & $7/7$ & \\
    \bottomrule
  \end{tabularx}
\end{table*}

\subsection{Selector pressure}
\label{sec:selector_pressure}

The benchmark selection rule keeps the clean-validation winner unchanged.
The selector-pressure check replaces that rule for robustness-improvement methods with a worst-scenario objective on perturbed validation data across all four datasets.
Within each tried method family, the alternative selector minimizes worst-scenario perturbed validation MSE, and the resulting selected variant is then evaluated under the unchanged test-time benchmark protocol.
The baseline architecture comparison is unchanged across all 28 common baseline rows, so the audit isolates selector pressure within the robustness-improvement search space.
The perturbed validation selector chooses a different variant in 18 of the 28 searches over datasets and method families, affecting 79 of the 136 method-baseline rows.
This is a demanding selector stress test of whether the clean-selected conclusions depend on the validation objective.
Despite these selection changes, the qualitative method pattern remains aligned with the clean-selected analysis.
Under the alternative selector, PGD adversarial training remains the clearest balanced train-time positive, ensemble remains the broadest low-assumption comparator, and RevIN can still improve nominal accuracy without improving sensor-fault robustness.
Across PGD adversarial training, ensemble, and RevIN over four datasets, 10 of 12 $\Delta\mathcal{D}_w$ sign categories are preserved.
The only sign changes are small near-zero crossings: ensemble on \BeijingTiantan{} moves from 0.007 to $-0.011$, and PGD adversarial training on Traffic moves from 0.023 to $-0.001$.
The largest absolute selector-induced shifts in mean $\Delta\mathcal{D}_w$ appear for adaptive robust loss on \PenmanshielWT{} ($0.144$ to $0.069$), fault augmentation on \BeijingTiantan{} ($-0.039$ to $-0.090$), and randomized training on \PenmanshielWT{} ($-0.036$ to $-0.081$).
Fault augmentation also becomes more favorable on Traffic, moving from $-0.025$ to $-0.065$.
ETTh1 is much less selector-sensitive: all dataset-mean $\Delta\mathcal{D}_w$ shifts are 0.007 or smaller in magnitude.
By contrast, the dataset-mean clean-MSE shifts remain small: about $-0.001$ on \BeijingTiantan{}, about $-0.002$ on \PenmanshielWT{} and ETTh1, and $-0.006$ on Traffic.
Worst-scenario perturbed validation therefore changes many selections while preserving the qualitative method interpretation under the official clean selector.
Table~\ref{tab:selector_pressure_method_deltas} reports the corresponding per-dataset method deltas under this alternative selector.

\begin{table*}[!tb]
\centering
\scriptsize
\setlength{\tabcolsep}{0pt}
\renewcommand{\arraystretch}{1.18}

\newcommand{\shiftbettertight}[1]{#1}
\newcommand{\shiftworsetight}[1]{#1}
\newcommand{\shiftsametight}[1]{#1}

\newlength{\selparenwd}
\settowidth{\selparenwd}{(\textminus{}0.000)}

\newcommand{\selcell}[3]{%
  \makebox[5.75em][r]{#1\,#2{\makebox[\selparenwd][r]{(#3)}}}%
}

\newcommand{\selcellbf}[3]{%
  \makebox[5.75em][r]{\textbf{#1}\,#2{\makebox[\selparenwd][r]{(#3)}}}%
}

\caption{Robustness-improvement method-baseline deltas when variants are selected on perturbed validation data.
The table uses the same four datasets and method-comparison layout as Table~\ref{tab:main_results}.
Method variants are selected by the lowest worst-scenario MSE on perturbed validation data rather than by the clean validation loss.
Parenthesized entries show selector-induced changes relative to the clean-selected values in Table~\ref{tab:main_results}, computed from the displayed three-decimal entries.
Negative parenthesized changes are more favorable, positive parenthesized changes are less favorable, and zeros indicate no rounded change.
Lower main values are better, and bold marks the lowest unrounded main value within each dataset--measure row.}
\vspace{7pt}
\label{tab:selector_pressure_method_deltas}

\begin{tabular*}{\textwidth}{@{\extracolsep{\fill}}cl*{7}{c}@{}}
\toprule
Data & Measure
& PGD & Ensemble
& \makecell{Randomized\\Training}
& \makecell{Randomized\\Smoothing}
& \makecell{Adaptive\\Robust Loss}
& RevIN
& \makecell{Fault\\Augmentation} \\
\midrule

\multirow{3}{*}{\rotatebox[origin=c]{90}{\BeijingTiantanTinyCentered{}}}
& $\Delta\!\mathcal{D}_w$
& \selcell{\textminus{}0.147}{\shiftbettertight}{\textminus{}0.011}
& \selcell{\textminus{}0.011}{\shiftbettertight}{\textminus{}0.018}
& \selcellbf{\textminus{}0.159}{\shiftbettertight}{\textminus{}0.013}
& \selcell{\textminus{}0.001}{\shiftbettertight}{\textminus{}0.002}
& \selcell{\textminus{}0.023}{\shiftworsetight}{0.001}
& \selcell{0.053}{\shiftsametight}{0.000}
& \selcell{\textminus{}0.090}{\shiftbettertight}{\textminus{}0.051} \\
& $\Delta\!\mathrm{MSE}_c$
& \selcell{0.001}{\shiftbettertight}{\textminus{}0.003}
& \selcellbf{\textminus{}0.010}{\shiftbettertight}{\textminus{}0.001}
& \selcell{0.036}{\shiftworsetight}{0.010}
& \selcell{0.002}{\shiftsametight}{0.000}
& \selcell{\textminus{}0.008}{\shiftbettertight}{\textminus{}0.009}
& \selcell{0.005}{\shiftsametight}{0.000}
& \selcell{0.003}{\shiftbettertight}{\textminus{}0.002} \\
& $\Delta\!\mathrm{MSE}_w$
& \selcellbf{\textminus{}0.045}{\shiftbettertight}{\textminus{}0.007}
& \selcell{\textminus{}0.016}{\shiftbettertight}{\textminus{}0.007}
& \selcell{\textminus{}0.009}{\shiftworsetight}{0.009}
& \selcell{0.003}{\shiftsametight}{0.000}
& \selcell{\textminus{}0.017}{\shiftbettertight}{\textminus{}0.012}
& \selcell{0.025}{\shiftsametight}{0.000}
& \selcell{\textminus{}0.026}{\shiftbettertight}{\textminus{}0.019} \\

\addlinespace[4pt]
\multirow{3}{*}{\rotatebox[origin=c]{90}{\PenmanshielWTTinyCentered{}}}
& $\Delta\!\mathcal{D}_w$
& \selcell{\textminus{}0.067}{\shiftbettertight}{\textminus{}0.008}
& \selcell{\textminus{}0.040}{\shiftbettertight}{\textminus{}0.017}
& \selcellbf{\textminus{}0.081}{\shiftbettertight}{\textminus{}0.045}
& \selcell{\textminus{}0.002}{\shiftsametight}{0.000}
& \selcell{0.069}{\shiftbettertight}{\textminus{}0.075}
& \selcell{0.107}{\shiftbettertight}{\textminus{}0.003}
& \selcell{\textminus{}0.044}{\shiftbettertight}{\textminus{}0.025} \\
& $\Delta\!\mathrm{MSE}_c$
& \selcell{\textminus{}0.005}{\shiftsametight}{0.000}
& \selcellbf{\textminus{}0.007}{\shiftsametight}{0.000}
& \selcell{0.001}{\shiftworsetight}{0.004}
& \selcell{0.003}{\shiftsametight}{0.000}
& \selcell{0.022}{\shiftbettertight}{\textminus{}0.018}
& \selcell{0.000}{\shiftworsetight}{0.001}
& \selcell{\textminus{}0.001}{\shiftsametight}{0.000} \\
& $\Delta\!\mathrm{MSE}_w$
& \selcellbf{\textminus{}0.029}{\shiftbettertight}{\textminus{}0.003}
& \selcell{\textminus{}0.022}{\shiftbettertight}{\textminus{}0.006}
& \selcell{\textminus{}0.027}{\shiftbettertight}{\textminus{}0.011}
& \selcell{0.003}{\shiftsametight}{0.000}
& \selcell{0.053}{\shiftbettertight}{\textminus{}0.049}
& \selcell{0.037}{\shiftsametight}{0.000}
& \selcell{\textminus{}0.016}{\shiftbettertight}{\textminus{}0.008} \\

\addlinespace[4pt]
\multirow{3}{*}{\rotatebox[origin=c]{90}{\tiny ETTh1}}
& $\Delta\!\mathcal{D}_w$
& \selcell{0.000}{\shiftsametight}{0.000}
& \selcell{\textminus{}0.005}{\shiftsametight}{0.000}
& \selcell{\textminus{}0.005}{\shiftbettertight}{\textminus{}0.001}
& \selcell{0.000}{\shiftsametight}{0.000}
& \selcellbf{\textminus{}0.006}{\shiftbettertight}{\textminus{}0.007}
& \selcell{0.122}{\shiftbettertight}{\textminus{}0.002}
& \selcell{\textminus{}0.006}{\shiftbettertight}{\textminus{}0.005} \\
& $\Delta\!\mathrm{MSE}_c$
& \selcell{\textminus{}0.005}{\shiftsametight}{0.000}
& \selcell{\textminus{}0.005}{\shiftsametight}{0.000}
& \selcell{0.009}{\shiftsametight}{0.000}
& \selcell{\textminus{}0.001}{\shiftsametight}{0.000}
& \selcell{0.015}{\shiftbettertight}{\textminus{}0.018}
& \selcellbf{\textminus{}0.045}{\shiftsametight}{0.000}
& \selcell{0.005}{\shiftworsetight}{0.002} \\
& $\Delta\!\mathrm{MSE}_w$
& \selcell{\textminus{}0.005}{\shiftsametight}{0.000}
& \selcellbf{\textminus{}0.007}{\shiftsametight}{0.000}
& \selcell{0.007}{\shiftbettertight}{\textminus{}0.001}
& \selcell{\textminus{}0.001}{\shiftsametight}{0.000}
& \selcell{0.012}{\shiftbettertight}{\textminus{}0.025}
& \selcell{0.010}{\shiftsametight}{0.000}
& \selcell{0.003}{\shiftworsetight}{0.001} \\

\addlinespace[4pt]
\multirow{3}{*}{\rotatebox[origin=c]{90}{\tiny Traffic}}
& $\Delta\!\mathcal{D}_w$
& \selcell{\textminus{}0.001}{\shiftbettertight}{\textminus{}0.024}
& \selcell{\textminus{}0.023}{\shiftbettertight}{\textminus{}0.004}
& \selcell{\textminus{}0.008}{\shiftbettertight}{\textminus{}0.019}
& \selcell{\textminus{}0.025}{\shiftbettertight}{\textminus{}0.004}
& \selcell{\textminus{}0.023}{\shiftbettertight}{\textminus{}0.016}
& \selcell{0.250}{\shiftbettertight}{\textminus{}0.017}
& \selcellbf{\textminus{}0.065}{\shiftbettertight}{\textminus{}0.040} \\
& $\Delta\!\mathrm{MSE}_c$
& \selcell{\textminus{}0.015}{\shiftworsetight}{0.005}
& \selcell{\textminus{}0.017}{\shiftsametight}{0.000}
& \selcell{0.001}{\shiftworsetight}{0.008}
& \selcell{0.012}{\shiftworsetight}{0.001}
& \selcell{0.005}{\shiftbettertight}{\textminus{}0.070}
& \selcellbf{\textminus{}0.061}{\shiftworsetight}{0.002}
& \selcell{0.000}{\shiftworsetight}{0.013} \\
& $\Delta\!\mathrm{MSE}_w$
& \selcell{\textminus{}0.023}{\shiftbettertight}{\textminus{}0.007}
& \selcellbf{\textminus{}0.040}{\shiftbettertight}{\textminus{}0.003}
& \selcell{\textminus{}0.007}{\shiftbettertight}{\textminus{}0.004}
& \selcell{0.006}{\shiftbettertight}{\textminus{}0.002}
& \selcell{\textminus{}0.005}{\shiftbettertight}{\textminus{}0.105}
& \selcell{0.066}{\shiftbettertight}{\textminus{}0.005}
& \selcell{\textminus{}0.034}{\shiftbettertight}{\textminus{}0.008} \\

\bottomrule
\end{tabular*}
\end{table*}

\subsection{Sensitivity to the number of affected channels}
\label{subsec:channel_count_sensitivity}

The benchmark couples severity to the number of affected channels for every channel-scoped scenario.
This keeps the protocol one-dimensional, but it also means severity combines within-channel corruption intensity with affected-channel count.
To isolate the channel-count component, we define a sensitivity variant with a fixed fraction of selected channels that replaces the severity-dependent channel-count rule by
\begin{equation*}
  k_p^{\mathrm{fix}}(s;q)=
  \begin{cases}
    0, & \text{if}\;\; s=0,\\
    \left\lceil q\,m_{\mathrm{cont}}\right\rceil, & \text{if}\;\; s\in(0,1],
  \end{cases}
\end{equation*}
for every channel-scoped scenario $p$, with $0<q\leq\gamma_{\max}$.
We use the midpoint setting $q=0.25$ and the maximum-fraction setting $q=\gamma_{\max}=0.5$.
The latter fixes the positive-severity channel count at the benchmark upper edge.
All other ingredients remain unchanged: the scenario-specific severity map $\theta_p(s)$, the temporal-window rules, the uniform severity averaging within each $p$, and the final maximum over scenario degradations.
MissingData is unchanged because it already affects all channels.

This comparison isolates the effect of increasing within-channel corruption intensity while holding the affected-channel fraction fixed.
The baseline protocol instead treats severity as a joint axis over corruption intensity and affected-channel count.

The sensitivity check keeps the same tested winner pools, the same uniform severity sampling distribution on $[0,1]$, the same evaluation budget of $K=10{,}000$ sampled test windows, and the same definitions of $\mathcal{D}_w$, $\mathrm{MSE}_c$, and $\mathrm{MSE}_w$ as the baseline protocol.
For baselines, the table tracks architecture ranks, signed and absolute shifts, and worst-scenario agreement for the two fault-facing reported quantities.
For methods, the accompanying interpretation checks whether the sign-level conclusions for PGD adversarial training, ensemble, and RevIN remain unchanged for the corresponding deltas.
At $q=\gamma_{\max}$, the fixed-fraction variant activates the maximal channel subset for every positive severity and therefore targets the low-severity region where the default channel-count rule is most conservative.
The midpoint setting checks whether qualitative ranks, fault-family labels, and method signs remain stable under a smaller multi-channel subset.
The diagnostic is especially relevant for additive value faults on the low-channel datasets, where the benchmark default would otherwise perturb only one or two channels at mild severity.
Agreement of qualitative conclusions is therefore more informative than equality of absolute score values.

\begin{table}[!tb]
  \centering
  \small
  \setlength{\tabcolsep}{1.0pt}
  \renewcommand{\arraystretch}{1.08}
  \caption{Channel-count sensitivity under reruns with fixed fractions of selected channels $q\in\{0.25,0.5\}$.
  The reruns keep severity maps, temporal rules, reporting definitions, and the budget of sampled test windows fixed, changing only the channel-count rule for channel-scoped scenarios.
  The $k_p^{\mathrm{fix}}$ column reports $\lceil q\,m_{\mathrm{cont}}\rceil$ for $s>0$ over the eligible continuous-channel pool.
  Mean signed shift is the fixed-fraction minus baseline-protocol mean over the seven baseline architectures, and largest absolute shift is the largest absolute per-architecture movement.
  Clean MSE is unchanged and omitted.
  Higher rank correlation and worst-scenario agreement values indicate stronger stability.}
  \label{tab:channel_count_sensitivity}
  \begin{tabularx}{\textwidth}{@{}
    >{\RaggedRight\arraybackslash}p{0.130\textwidth}
    >{\centering\arraybackslash}p{0.048\textwidth}
    >{\raggedleft\arraybackslash}p{0.060\textwidth}
    @{\hspace{1.7em}}
    *{2}{S[table-format=-1.3,table-number-alignment=center,mode=text]}
    @{\hspace{0.45em}}c@{\hspace{0.45em}}
    *{2}{S[table-format=-1.3,table-number-alignment=center,mode=text]}
    @{\hspace{0.45em}}c@{\hspace{0.45em}}
    *{2}{S[table-format=-1.3,table-number-alignment=center,mode=text]}
    @{\hspace{0.45em}}c@{\hspace{0.45em}}
    X
    c@{\hspace{1.8em}}c
    X
    @{}}
    \toprule
      & &
      & \multicolumn{2}{c}{Mean signed shift}
      &
      & \multicolumn{2}{c}{Largest abs. shift}
      &
      & \multicolumn{2}{c}{Rank corr.}
      &
      & \multicolumn{4}{c}{Worst-scenario match} \\
    \cmidrule{4-5}
    \cmidrule{7-8}
    \cmidrule(lr){10-11}
    \cmidrule(lr){13-16}
      Data
      & {$q$}
      & {$k_p^{\mathrm{fix}}$}
      & {$\mathcal{D}_w$} & {$\mathrm{MSE}_w$}
      &
      & {$\mathcal{D}_w$} & {$\mathrm{MSE}_w$}
      &
      & {$\mathcal{D}_w$} & {$\mathrm{MSE}_w$}
      &
      & & Exact & Class & \\
    \midrule
    \multirow{2}{*}{\BeijingTiantanShort{}}
      & {$0.25$} & 3   & -0.102 & -0.034 & & 0.163 & 0.054 & & 1.000 & 1.000 & & & 5/7 & 7/7 & \\
      & {$0.50$} & 6   &  0.148 &  0.048 & & 0.273 & 0.090 & & 1.000 & 0.964 & & & 7/7 & 7/7 & \\
    \addlinespace[2pt]
    \multirow{2}{*}{\PenmanshielWTShort{}}
      & {$0.25$} & 17  & -0.071 & -0.026 & & 0.108 & 0.042 & & 0.964 & 0.964 & & & 7/7 & 7/7 & \\
      & {$0.50$} & 33  &  0.068 &  0.025 & & 0.100 & 0.039 & & 1.000 & 1.000 & & & 6/7 & 7/7 & \\
    \addlinespace[2pt]
    \multirow{2}{*}{ETTh1}
      & {$0.25$} & 2   & -0.038 & -0.017 & & 0.097 & 0.048 & & 0.964 & 0.821 & & & 5/7 & 7/7 & \\
      & {$0.50$} & 4   &  0.107 &  0.051 & & 0.209 & 0.104 & & 0.964 & 0.964 & & & 5/7 & 7/7 & \\
    \addlinespace[2pt]
    \multirow{2}{*}{Traffic}
      & {$0.25$} & 216 & -0.037 & -0.023 & & 0.262 & 0.158 & & 0.964 & 1.000 & & & 7/7 & 7/7 & \\
      & {$0.50$} & 431 &  0.088 &  0.050 & & 0.284 & 0.171 & & 0.857 & 0.857 & & & 2/7 & 7/7 & \\
    \bottomrule
  \end{tabularx}
\end{table}

The two fixed-fraction settings in Table~\ref{tab:channel_count_sensitivity} show that the channel-count rule changes stress calibration without changing clean forecasting error.
Each fixed-fraction row is compared with the baseline protocol separately.
The clean-MSE quantity is unchanged because the trained models, clean sampled windows, and clean evaluation are identical.
At $q=0.25$, $\mathcal{D}_w$ and $\mathrm{MSE}_w$ decrease on average for every dataset.
At $q=0.5$, they increase on average for every dataset.
For $\mathcal{D}_w$, the largest mean score shifts appear on \BeijingTiantan{} at both fractions, while Traffic has the largest single-architecture score shift at both fractions.
Architecture ranks remain identical or nearly identical in most affected rows, with the largest rank movement appearing for ETTh1 $\mathrm{MSE}_w$ at $q=0.25$ and for both Traffic affected quantities at $q=0.5$.
Exact worst-scenario labels can change, but the scenario class matches for all seven baseline architectures on every dataset and both fixed fractions.
For the three tracked robustness-improvement methods, signs stay unchanged for $\Delta\mathcal{D}_w$ and $\Delta\mathrm{MSE}_c$ at both fractions and for $\Delta\mathrm{MSE}_w$ at $q=0.5$.
The only sign change is RevIN's near-zero ETTh1 $\Delta\mathrm{MSE}_w$ at $q=0.25$.
The fixed-fraction check therefore strengthens the architecture and fault-family interpretation and leaves the method reading unchanged except for that near-zero RevIN sign change.
It also establishes, within this benchmark, that the channel-count rule is a stress-calibration choice rather than a claim about a unique prevalence of multi-sensor faults.

\FloatBarrier
\begingroup
\makeatletter
\setlength{\@fptop}{0pt}
\setlength{\@fpsep}{\floatsep}
\makeatother

\subsection{Sensitivity to severity weighting}
\label{subsec:severity_weighting_sensitivity}

Weighting severity toward mild faults preserves all four degradation leaders, 26 of 28 scenario-class assignments, and 19 of 24 signs of method effects.
We reevaluate the models of Table~\ref{tab:main_results}, selected on clean validation data at one training and search seed, and draw severity from $\mathrm{Beta}(1,2)$ instead of the uniform $\mathrm{Beta}(1,1)$ of the default configuration.
The support and endpoints remain unchanged, while mean severity falls from $1/2$ to $1/3$.
Neither weighting estimates how often severities occur in deployment, but a distribution calibrated to the recorded faults of a specific system can replace either one.
Clean errors remain fixed, so changes in fault quantities reflect the alternative severity weighting.
Table~\ref{tab:beta_ranks} summarizes the architecture comparisons, and Table~\ref{tab:beta_architecture} gives the complete results.

Two leaders in worst-scenario fault-time MSE change: GRU to ModernTCN on \PenmanshielWT{}, with a margin of approximately $0.001$, and ModernTCN to PatchTST on Traffic, with a margin of approximately $0.042$.
The worst scenario changes class for Chronos-2 on \BeijingTiantan{} (StuckSensor to Drift) and GRU on \PenmanshielWT{} (Drift to MissingData).
Other exact scenario labels can also change within a class, as shown in Table~\ref{tab:beta_architecture}.

\begin{table}[!tb]
  \centering
  \small
  \setlength{\tabcolsep}{2.4pt}
  \renewcommand{\arraystretch}{1.10}
  \caption{Architecture rankings under two severity sampling distributions for the models in Table~\ref{tab:main_results}.
  $\mathrm{Beta}(1,1)$ is the uniform distribution of the default configuration, whereas $\mathrm{Beta}(1,2)$ weights mild faults more heavily.
  Spearman correlations compare the two distributions, and class counts give unchanged assignments among seven architectures.}

  \label{tab:beta_ranks}
  \begin{tabular*}{\textwidth}{@{\extracolsep{\fill}}l*{2}{S[table-format=1.3]}cll@{}}
    \toprule
 & \multicolumn{2}{c}{Rank correlation} & & \multicolumn{2}{c}{Lowest $\mathrm{MSE}_w$} \\
\cmidrule(lr){2-3}\cmidrule(lr){5-6}
Dataset & {$\mathcal{D}_w$} & {$\mathrm{MSE}_w$} & Classes preserved & $\mathrm{Beta}(1,1)$ & $\mathrm{Beta}(1,2)$ \\
    \midrule
\BeijingTiantan{} & 1.000 & 0.857 & 6/7 & GRU & GRU \\
\PenmanshielWT{} & 0.964 & 0.929 & 6/7 & GRU & ModernTCN \\
ETTh1 & 0.929 & 0.821 & 7/7 & DLinear & DLinear \\
Traffic & 0.964 & 0.893 & 7/7 & ModernTCN & PatchTST \\
    \bottomrule
  \end{tabular*}
\end{table}

\begin{table}[!tb]
  \centering
  \small
  \setlength{\tabcolsep}{2.4pt}
  \renewcommand{\arraystretch}{1.10}
  \caption{Architecture results under the severity sampling distributions $\mathrm{Beta}(1,1)$ and $\mathrm{Beta}(1,2)$ for the models in Table~\ref{tab:main_results}, with selected models and clean MSE held fixed.
  Lower values are better for all three quantities.}

  \label{tab:beta_architecture}
  \begin{tabular*}{\textwidth}{@{\extracolsep{\fill}}clS[table-format=1.3]*{2}{S[table-format=1.3]S[table-format=1.3]l}@{}}
    \toprule
 & & & \multicolumn{3}{c}{$\mathrm{Beta}(1,1)$} & \multicolumn{3}{c}{$\mathrm{Beta}(1,2)$} \\
\cmidrule(lr){4-6}\cmidrule(lr){7-9}
Data & Architecture & {$\mathrm{MSE}_c$} & {$\mathcal{D}_w$} & {$\mathrm{MSE}_w$} & Worst scenario & {$\mathcal{D}_w$} & {$\mathrm{MSE}_w$} & Worst scenario \\
    \midrule
    \multirow{7}{*}{\rotatebox[origin=c]{90}{\footnotesize \begin{tabular}{@{}c@{}}Beijing Air\\Tiantan\end{tabular}}} & SeasonalNaive & 0.353 & 1.353 & 0.477 & Noise & 1.158 & 0.408 & Noise \\
     & DLinear & 0.329 & 1.628 & 0.536 & Noise & 1.283 & 0.422 & Noise \\
     & GRU & 0.281 & 1.251 & 0.352 & Attenuation & 1.119 & 0.315 & Noise \\
     & ModernTCN & 0.301 & 1.502 & 0.452 & Noise & 1.232 & 0.371 & Noise \\
     & TSMixer & 0.359 & 1.153 & 0.414 & Attenuation & 1.060 & 0.380 & Attenuation \\
     & PatchTST & 0.338 & 1.376 & 0.465 & Noise & 1.167 & 0.394 & Noise \\
     & Chronos-2 & 0.349 & 1.230 & 0.429 & StuckSensor & 1.080 & 0.377 & Drift \\
    \addlinespace[4pt]
    \multirow{7}{*}{\rotatebox[origin=c]{90}{\footnotesize \begin{tabular}{@{}c@{}}Penmanshiel\\WT08\end{tabular}}} & SeasonalNaive & 0.389 & 1.320 & 0.514 & Noise & 1.117 & 0.435 & Noise \\
     & DLinear & 0.340 & 1.259 & 0.428 & Noise & 1.088 & 0.370 & Noise \\
     & GRU & 0.346 & 1.064 & 0.368 & Drift & 1.032 & 0.357 & MissingData \\
     & ModernTCN & 0.336 & 1.183 & 0.397 & Noise & 1.061 & 0.356 & Noise \\
     & TSMixer & 0.353 & 1.256 & 0.443 & Noise & 1.090 & 0.385 & Noise \\
     & PatchTST & 0.354 & 1.202 & 0.426 & Drift & 1.079 & 0.382 & Drift \\
     & Chronos-2 & 0.401 & 1.160 & 0.466 & Drift & 1.060 & 0.426 & Drift \\
    \addlinespace[4pt]
    \multirow{7}{*}{\rotatebox[origin=c]{90}{\footnotesize ETTh1}} & SeasonalNaive & 0.634 & 1.288 & 0.817 & MissingData & 1.200 & 0.761 & MissingData \\
     & DLinear & 0.438 & 1.251 & 0.548 & StuckSensor & 1.108 & 0.485 & MissingData \\
     & GRU & 0.664 & 1.064 & 0.707 & MissingData & 1.043 & 0.693 & MissingData \\
     & ModernTCN & 0.462 & 1.216 & 0.562 & StuckSensor & 1.115 & 0.515 & MissingData \\
     & TSMixer & 0.475 & 1.169 & 0.555 & StuckSensor & 1.103 & 0.524 & MissingData \\
     & PatchTST & 0.431 & 1.312 & 0.566 & StuckSensor & 1.142 & 0.493 & MissingData \\
     & Chronos-2 & 0.496 & 1.448 & 0.718 & StuckSensor & 1.209 & 0.599 & StuckSensor \\
    \addlinespace[4pt]
    \multirow{7}{*}{\rotatebox[origin=c]{90}{\footnotesize Traffic}} & SeasonalNaive & 1.270 & 1.210 & 1.537 & MissingData & 1.154 & 1.466 & MissingData \\
     & DLinear & 0.684 & 1.177 & 0.806 & MissingData & 1.124 & 0.769 & MissingData \\
     & GRU & 0.726 & 1.284 & 0.932 & MissingData & 1.203 & 0.873 & MissingData \\
     & ModernTCN & 0.575 & 1.265 & 0.728 & MissingData & 1.158 & 0.666 & MissingData \\
     & TSMixer & 0.631 & 1.290 & 0.814 & MissingData & 1.183 & 0.746 & MissingData \\
     & PatchTST & 0.453 & 1.676 & 0.759 & MissingData & 1.378 & 0.624 & MissingData \\
     & Chronos-2 & 0.602 & 1.911 & 1.150 & StuckSensor & 1.382 & 0.831 & StuckSensor \\
    \bottomrule
  \end{tabular*}
\end{table}

For PGD adversarial training, fault augmentation, and RevIN, five of the 24 effects on dataset means reverse sign across the two fault quantities (Table~\ref{tab:beta_methods}).
The degradation reversals are RevIN on \BeijingTiantan{} ($0.053$ to $-0.016$) and fault augmentation on ETTh1 ($-0.001$ to $0.004$).
The reversals in worst-scenario fault-time MSE are fault augmentation on \BeijingTiantan{} ($-0.007$ to $0.001$), RevIN on ETTh1 ($0.010$ to $-0.029$), and RevIN on Traffic ($0.071$ to $-0.002$).
All five final effects have absolute magnitude below $0.030$.
PGD retains all four degradation signs, fault augmentation still reduces degradation on Traffic, and RevIN still increases degradation on three datasets.
Tables~\ref{tab:beta_pairs_single_target} and~\ref{tab:beta_pairs_all_channel} give all 56 comparisons of a method with its baseline.

\begin{table}[!tb]
  \centering
  \small
  \setlength{\tabcolsep}{2.4pt}
  \renewcommand{\arraystretch}{1.10}
  \caption{Mean method-baseline deltas under the severity sampling distributions $\mathrm{Beta}(1,1)$ and $\mathrm{Beta}(1,2)$ for the models in Table~\ref{tab:main_results}.
  Means cover five architectures for PGD adversarial training and fault augmentation, and four for RevIN, excluding PatchTST.
  Negative values indicate improvement, with $\Delta\mathrm{MSE}_w$ comparing each pair's respective worst scenarios.
  Clean-error deltas match the corresponding entries in Table~\ref{tab:main_results}.}

  \label{tab:beta_methods}
  \begin{tabular}{@{}l@{\hspace{1.5em}}l@{\hspace{2.5em}}S[table-format=-1.3]@{\hspace{2em}}S[table-format=-1.3]@{}p{3em}@{}S[table-format=-1.3]@{\hspace{2em}}S[table-format=-1.3]@{}}
    \toprule
 & & \multicolumn{2}{c}{$\mathrm{Beta}(1,1)$} & & \multicolumn{2}{c}{$\mathrm{Beta}(1,2)$} \\
\cmidrule(lr){3-4}\cmidrule(lr){6-7}
Dataset & Method & {$\Delta\mathcal{D}_w$} & {$\Delta\mathrm{MSE}_w$} & & {$\Delta\mathcal{D}_w$} & {$\Delta\mathrm{MSE}_w$} \\
    \midrule
\multirow{3}{*}{\BeijingTiantan{}} & PGD & -0.136 & -0.038 & & -0.057 & -0.013 \\
 & Fault augmentation & -0.039 & -0.007 & & -0.014 & 0.001 \\
 & RevIN & 0.053 & 0.025 & & -0.016 & 0.002 \\
\addlinespace[3pt]
\multirow{3}{*}{\PenmanshielWT{}} & PGD & -0.059 & -0.026 & & -0.018 & -0.012 \\
 & Fault augmentation & -0.019 & -0.008 & & -0.008 & -0.004 \\
 & RevIN & 0.110 & 0.037 & & 0.028 & 0.009 \\
\addlinespace[3pt]
\multirow{3}{*}{ETTh1} & PGD & 0.000 & -0.005 & & 0.006 & -0.003 \\
 & Fault augmentation & -0.001 & 0.002 & & 0.004 & 0.005 \\
 & RevIN & 0.124 & 0.010 & & 0.039 & -0.029 \\
\addlinespace[3pt]
\multirow{3}{*}{Traffic} & PGD & 0.023 & -0.016 & & 0.017 & -0.017 \\
 & Fault augmentation & -0.025 & -0.026 & & -0.017 & -0.021 \\
 & RevIN & 0.267 & 0.071 & & 0.128 & -0.002 \\
\bottomrule
  \end{tabular}
\end{table}

\begin{table}[!tb]
  \centering
  \small
  \setlength{\tabcolsep}{2.4pt}
  \renewcommand{\arraystretch}{1.10}
  \caption{Method-baseline deltas under the severity sampling distributions $\mathrm{Beta}(1,1)$ and $\mathrm{Beta}(1,2)$ on \BeijingTiantan{} and \PenmanshielWT{} for the models in Table~\ref{tab:main_results}.
  PGD denotes PGD adversarial training and Fault Aug. denotes fault augmentation, with RevIN excluding PatchTST.
  Negative values indicate improvement, with $\Delta\mathrm{MSE}_w$ comparing each pair's respective worst scenarios.
  Clean-error deltas are unchanged.}

  \label{tab:beta_pairs_single_target}
  \begin{tabular*}{\textwidth}{@{\extracolsep{\fill}}ll*{8}{S[table-format=-1.3]}@{}}
    \toprule
     & & \multicolumn{4}{c}{\BeijingTiantan{}} & \multicolumn{4}{c}{\PenmanshielWT{}} \\
    \cmidrule(lr){3-6}\cmidrule(lr){7-10}
     & & \multicolumn{2}{c}{$\mathrm{Beta}(1,1)$} & \multicolumn{2}{c}{$\mathrm{Beta}(1,2)$} & \multicolumn{2}{c}{$\mathrm{Beta}(1,1)$} & \multicolumn{2}{c}{$\mathrm{Beta}(1,2)$} \\
    \cmidrule(lr){3-4}\cmidrule(lr){5-6}\cmidrule(lr){7-8}\cmidrule(lr){9-10}
    Architecture & Method & {$\Delta\mathcal{D}_w$} & {$\Delta\mathrm{MSE}_w$} & {$\Delta\mathcal{D}_w$} & {$\Delta\mathrm{MSE}_w$} & {$\Delta\mathcal{D}_w$} & {$\Delta\mathrm{MSE}_w$} & {$\Delta\mathcal{D}_w$} & {$\Delta\mathrm{MSE}_w$} \\
    \midrule
    \multirow{3}{*}{DLinear} & PGD & -0.335 & -0.111 & -0.151 & -0.050 & -0.037 & -0.015 & -0.009 & -0.005 \\
     & Fault Aug. & -0.070 & -0.015 & -0.032 & -0.004 & -0.005 & -0.002 & -0.002 & -0.001 \\
     & RevIN & -0.194 & -0.056 & -0.087 & -0.022 & 0.000 & 0.008 & 0.000 & 0.007 \\
    \addlinespace[3pt]
    \multirow{3}{*}{GRU} & PGD & -0.018 & 0.010 & -0.012 & 0.010 & -0.027 & -0.015 & -0.006 & -0.007 \\
     & Fault Aug. & 0.009 & 0.025 & 0.007 & 0.022 & 0.002 & 0.000 & 0.000 & -0.001 \\
     & RevIN & 0.234 & 0.097 & 0.006 & 0.025 & 0.381 & 0.133 & 0.091 & 0.033 \\
    \addlinespace[3pt]
    \multirow{3}{*}{ModernTCN} & PGD & -0.207 & -0.038 & -0.084 & -0.004 & -0.082 & -0.024 & -0.022 & -0.004 \\
     & Fault Aug. & -0.019 & 0.002 & -0.006 & 0.005 & -0.018 & -0.004 & -0.004 & 0.001 \\
     & RevIN & -0.060 & 0.010 & -0.030 & 0.014 & 0.032 & 0.005 & 0.018 & 0.001 \\
    \addlinespace[3pt]
    \multirow{3}{*}{TSMixer} & PGD & -0.005 & -0.017 & 0.005 & -0.012 & -0.146 & -0.070 & -0.053 & -0.036 \\
     & Fault Aug. & -0.049 & -0.018 & -0.014 & -0.005 & -0.092 & -0.029 & -0.042 & -0.012 \\
     & RevIN & 0.233 & 0.048 & 0.048 & -0.011 & 0.027 & 0.002 & 0.002 & -0.006 \\
    \addlinespace[3pt]
    \multirow{2}{*}{PatchTST} & PGD & -0.117 & -0.035 & -0.044 & -0.011 & 0.000 & -0.008 & 0.000 & -0.007 \\
     & Fault Aug. & -0.064 & -0.027 & -0.023 & -0.013 & 0.019 & -0.004 & 0.007 & -0.007 \\
    \bottomrule
  \end{tabular*}
\end{table}

\begin{table}[!htb]
  \centering
  \small
  \setlength{\tabcolsep}{2.4pt}
  \renewcommand{\arraystretch}{1.10}
  \caption{Method-baseline deltas under the severity sampling distributions $\mathrm{Beta}(1,1)$ and $\mathrm{Beta}(1,2)$ on ETTh1 and Traffic for the models in Table~\ref{tab:main_results}.
  PGD denotes PGD adversarial training and Fault Aug. denotes fault augmentation, with RevIN excluding PatchTST.
  Negative values indicate improvement, with $\Delta\mathrm{MSE}_w$ comparing each pair's respective worst scenarios.
  Clean-error deltas are unchanged.}

  \label{tab:beta_pairs_all_channel}
  \begin{tabular*}{\textwidth}{@{\extracolsep{\fill}}ll*{8}{S[table-format=-1.3]}@{}}
    \toprule
     & & \multicolumn{4}{c}{ETTh1} & \multicolumn{4}{c}{Traffic} \\
    \cmidrule(lr){3-6}\cmidrule(lr){7-10}
     & & \multicolumn{2}{c}{$\mathrm{Beta}(1,1)$} & \multicolumn{2}{c}{$\mathrm{Beta}(1,2)$} & \multicolumn{2}{c}{$\mathrm{Beta}(1,1)$} & \multicolumn{2}{c}{$\mathrm{Beta}(1,2)$} \\
    \cmidrule(lr){3-4}\cmidrule(lr){5-6}\cmidrule(lr){7-8}\cmidrule(lr){9-10}
    Architecture & Method & {$\Delta\mathcal{D}_w$} & {$\Delta\mathrm{MSE}_w$} & {$\Delta\mathcal{D}_w$} & {$\Delta\mathrm{MSE}_w$} & {$\Delta\mathcal{D}_w$} & {$\Delta\mathrm{MSE}_w$} & {$\Delta\mathcal{D}_w$} & {$\Delta\mathrm{MSE}_w$} \\
    \midrule
    \multirow{3}{*}{DLinear} & PGD & 0.002 & 0.005 & 0.005 & 0.006 & -0.002 & -0.002 & -0.003 & -0.003 \\
     & Fault Aug. & -0.006 & -0.003 & 0.000 & 0.000 & -0.001 & 0.000 & -0.001 & 0.000 \\
     & RevIN & 0.039 & 0.030 & 0.012 & 0.016 & 0.135 & 0.082 & 0.046 & 0.023 \\
    \addlinespace[3pt]
    \multirow{3}{*}{GRU} & PGD & 0.004 & -0.031 & 0.001 & -0.032 & -0.003 & -0.041 & -0.005 & -0.039 \\
     & Fault Aug. & -0.002 & 0.015 & -0.002 & 0.015 & 0.040 & -0.013 & 0.038 & -0.012 \\
     & RevIN & 0.209 & -0.040 & 0.063 & -0.114 & 0.247 & 0.045 & 0.147 & -0.012 \\
    \addlinespace[3pt]
    \multirow{3}{*}{ModernTCN} & PGD & 0.000 & 0.003 & 0.001 & 0.003 & -0.029 & -0.009 & -0.019 & -0.003 \\
     & Fault Aug. & -0.004 & 0.001 & -0.002 & 0.002 & -0.021 & 0.012 & -0.013 & 0.015 \\
     & RevIN & 0.092 & 0.020 & 0.041 & -0.001 & 0.338 & 0.102 & 0.181 & 0.027 \\
    \addlinespace[3pt]
    \multirow{3}{*}{TSMixer} & PGD & 0.002 & -0.007 & -0.003 & -0.009 & 0.083 & -0.031 & 0.060 & -0.038 \\
     & Fault Aug. & -0.003 & 0.000 & -0.001 & 0.001 & 0.030 & -0.050 & 0.020 & -0.050 \\
     & RevIN & 0.157 & 0.032 & 0.039 & -0.019 & 0.346 & 0.054 & 0.136 & -0.047 \\
    \addlinespace[3pt]
    \multirow{2}{*}{PatchTST} & PGD & -0.006 & 0.004 & 0.026 & 0.017 & 0.063 & 0.001 & 0.049 & 0.000 \\
     & Fault Aug. & 0.009 & 0.000 & 0.023 & 0.006 & -0.174 & -0.078 & -0.131 & -0.058 \\
    \bottomrule
  \end{tabular*}
\end{table}

\subsection{Sensitivity to fault direction}
\label{subsec:fault_direction_sensitivity}

Replacing Drift and Spike with negative variants preserves every architecture leader, all rankings by worst-scenario fault-time MSE, and all 28 scenario-class assignments for the models in Table~\ref{tab:main_results}.
The original scenarios add positive offsets, whereas negative Drift and negative Spike subtract the same magnitudes at the same sampled channels and times, with selected models, clean controls, and severity draws held fixed.
The other six scenarios are unchanged.
The only change in degradation ranks swaps ranks 2 and 3 on \PenmanshielWT{} below GRU, giving a Spearman correlation of $0.964$.
Table~\ref{tab:direction_architecture} gives the complete architecture comparison.
This diagnostic covers the opposite direction at one training and search seed, while the default configuration retains positive Drift and Spike.

\begin{table}[!htb]
  \centering
  \small
  \setlength{\tabcolsep}{2.4pt}
  \renewcommand{\arraystretch}{1.10}
  \caption{Architecture results with original and negative Drift and Spike scenarios for the models in Table~\ref{tab:main_results}.
  Original scenarios add positive offsets, while negative variants subtract the same magnitudes at the same sampled channels and times.
  Each group reports worst-scenario scores across eight scenarios, with only Drift and Spike replaced in the negative variant and selected models, clean MSE, and the other six scenarios held fixed.
  Lower values are better, and bold marks the best unrounded score for each quantity, dataset, and direction.}

  \label{tab:direction_architecture}
  \begin{tabular*}{\textwidth}{@{\extracolsep{\fill}}clS[table-format=1.3,detect-weight=true,mode=text]*{2}{S[table-format=1.3,detect-weight=true,mode=text]S[table-format=1.3,detect-weight=true,mode=text]l}@{}}
    \toprule
 & & & \multicolumn{3}{c}{Original Drift and Spike} & \multicolumn{3}{c}{Negative Drift and Spike} \\
\cmidrule(lr){4-6}\cmidrule(lr){7-9}
Data & Architecture & {$\mathrm{MSE}_c$} & {$\mathcal{D}_w$} & {$\mathrm{MSE}_w$} & Worst scenario & {$\mathcal{D}_w$} & {$\mathrm{MSE}_w$} & Worst scenario \\
    \midrule
    \multirow{7}{*}{\rotatebox[origin=c]{90}{\footnotesize \begin{tabular}{@{}c@{}}Beijing Air\\Tiantan\end{tabular}}} & SeasonalNaive & 0.353 & 1.353 & 0.477 & Noise & 1.353 & 0.477 & Noise \\
     & DLinear & 0.329 & 1.628 & 0.536 & Noise & 1.628 & 0.536 & Noise \\
     & GRU & \bfseries 0.281 & 1.251 & \bfseries 0.352 & Attenuation & 1.251 & \bfseries 0.352 & Attenuation \\
     & ModernTCN & 0.301 & 1.502 & 0.452 & Noise & 1.502 & 0.452 & Noise \\
     & TSMixer & 0.359 & \bfseries 1.153 & 0.414 & Attenuation & \bfseries 1.153 & 0.414 & Attenuation \\
     & PatchTST & 0.338 & 1.376 & 0.465 & Noise & 1.376 & 0.465 & Noise \\
     & Chronos-2 & 0.349 & 1.230 & 0.429 & StuckSensor & 1.230 & 0.429 & StuckSensor \\
    \addlinespace[4pt]
    \multirow{7}{*}{\rotatebox[origin=c]{90}{\footnotesize \begin{tabular}{@{}c@{}}Penmanshiel\\WT08\end{tabular}}} & SeasonalNaive & 0.389 & 1.320 & 0.514 & Noise & 1.320 & 0.514 & Noise \\
     & DLinear & 0.340 & 1.259 & 0.428 & Noise & 1.259 & 0.428 & Noise \\
     & GRU & 0.346 & \bfseries 1.064 & \bfseries 0.368 & Drift & \bfseries 1.065 & \bfseries 0.368 & Negative Drift \\
     & ModernTCN & \bfseries 0.336 & 1.183 & 0.397 & Noise & 1.183 & 0.397 & Noise \\
     & TSMixer & 0.353 & 1.256 & 0.443 & Noise & 1.256 & 0.443 & Noise \\
     & PatchTST & 0.354 & 1.202 & 0.426 & Drift & 1.197 & 0.424 & Negative Drift \\
     & Chronos-2 & 0.401 & 1.160 & 0.466 & Drift & 1.192 & 0.479 & Negative Drift \\
    \addlinespace[4pt]
    \multirow{7}{*}{\rotatebox[origin=c]{90}{\footnotesize ETTh1}} & SeasonalNaive & 0.634 & 1.288 & 0.817 & MissingData & 1.288 & 0.817 & MissingData \\
     & DLinear & 0.438 & 1.251 & \bfseries 0.548 & StuckSensor & 1.251 & \bfseries 0.548 & StuckSensor \\
     & GRU & 0.664 & \bfseries 1.064 & 0.707 & MissingData & \bfseries 1.064 & 0.707 & MissingData \\
     & ModernTCN & 0.462 & 1.216 & 0.562 & StuckSensor & 1.216 & 0.562 & StuckSensor \\
     & TSMixer & 0.475 & 1.169 & 0.555 & StuckSensor & 1.169 & 0.555 & StuckSensor \\
     & PatchTST & \bfseries 0.431 & 1.312 & 0.566 & StuckSensor & 1.312 & 0.566 & StuckSensor \\
     & Chronos-2 & 0.496 & 1.448 & 0.718 & StuckSensor & 1.448 & 0.718 & StuckSensor \\
    \addlinespace[4pt]
    \multirow{7}{*}{\rotatebox[origin=c]{90}{\footnotesize Traffic}} & SeasonalNaive & 1.270 & 1.210 & 1.537 & MissingData & 1.210 & 1.537 & MissingData \\
     & DLinear & 0.684 & \bfseries 1.177 & 0.806 & MissingData & \bfseries 1.177 & 0.806 & MissingData \\
     & GRU & 0.726 & 1.284 & 0.932 & MissingData & 1.284 & 0.932 & MissingData \\
     & ModernTCN & 0.575 & 1.265 & \bfseries 0.728 & MissingData & 1.265 & \bfseries 0.728 & MissingData \\
     & TSMixer & 0.631 & 1.290 & 0.814 & MissingData & 1.290 & 0.814 & MissingData \\
     & PatchTST & \bfseries 0.453 & 1.676 & 0.759 & MissingData & 1.676 & 0.759 & MissingData \\
     & Chronos-2 & 0.602 & 1.911 & 1.150 & StuckSensor & 1.911 & 1.150 & StuckSensor \\
    \bottomrule
  \end{tabular*}
\end{table}

\subsection{Dependence on TimeCompress}
\label{subsec:timecompress_dependence}

TimeCompress traverses the source signal faster than the output grid and repeats the final observed value once the warped source index exceeds the available history (see Definition~\ref{def:stretch_compress}).
Its degradation thus mixes compression with endpoint holding, so we test whether the benchmark conclusions depend on this construction.
TimeCompress is not the worst scenario of any model in the 84 architecture results and 168 method-baseline comparisons across the three training and search seeds (Appendix~\ref{subsec:training_seed_sensitivity}).
These comparisons cover PGD adversarial training, fault augmentation, and RevIN, each paired with the baseline forecasting model from the same seed.
Omitting TimeCompress therefore leaves $\mathcal{D}_w$, $\mathrm{MSE}_w$, and their method-baseline deltas unchanged.

To quantify how much of the warped window holds the final observed value, we define its clipped share as the fraction of the 48 window positions whose source index exceeds the history.
Averaged over window starts, this share is $3.8\%$, $19.7\%$, $52.7\%$, and $86.2\%$ at severities $0.25$, $0.5$, $0.75$, and $1$, respectively, and $29.2\%$ in expectation under uniform severity.

A separate timing diagnostic on the models in Table~\ref{tab:main_results} replaces TimeCompress with a local warp that preserves both endpoints of the same selected 48-step window.
For relative position $v\in[0,1]$ within that window and severity $s\in[0,1]$, the map to source coordinates is
\begin{equation*}
  g_s(v)=v+\frac{0.9s}{\pi}\sin(\pi v).
\end{equation*}
The map locally compresses and stretches time without sampling beyond the window.
With this replacement, all 28 architecture results remain unchanged, including the worst-scenario quantities, their rankings, and the worst-scenario classes.
The two timing scenarios rank the seven architectures by degradation in close agreement on three datasets, with Spearman correlations of $0.929$ on \PenmanshielWT{}, $0.857$ on ETTh1, and $0.964$ on Traffic.
On \BeijingTiantan{}, the correlation is $-0.571$, but it orders near-zero effects, because the local warp changes error by only $-1.00\%$ to $0.04\%$ relative to clean error.
Because the two warps also map the window to different source coordinates, this comparison does not isolate the effect of clipping.
Compression at a constant rate without endpoint holding remains untested.

\FloatBarrier
\endgroup

\section{Full experimental details}
\label{app:full_experimental_details}

Table~\ref{tab:experimental_protocol} collects the choices that define the default configuration.
It separates data and evaluation choices, training and selection rules, and the recorded compute of all experiments.

\begin{table}[!htb]
  \centering
  \small
  \setlength{\tabcolsep}{4.0pt}
  \renewcommand{\arraystretch}{1.12}
  \caption{Full experimental protocol for the uniform-severity benchmark and recorded compute of all experiments.}
  \label{tab:experimental_protocol}
  \begin{tabularx}{\textwidth}{@{}
    >{\RaggedRight\arraybackslash}p{0.27\textwidth}
    >{\arraybackslash}X
    @{}}
    \toprule
    Component & Benchmark choice \\
    \midrule
    \multicolumn{2}{@{}l}{\textit{Data and evaluation}} \\
    \addlinespace[1pt]
    Dataset windows & \BeijingTiantan{} and \PenmanshielWT{} use 96 input steps and forecast one target for 6 steps, whereas ETTh1 and Traffic use 96 input steps and forecast all channels for 96 steps. \\
    Data splits & Temporal split with 0.6 train, 0.2 validation, and 0.2 test proportions. \\
    Evaluation budget & $K=10{,}000$ sampled test windows per dataset. \\
    Benchmark perturbations & Eight scenarios, severity sampled uniformly over $[0,1]$, and a cap of $\gamma_{\max}=0.5$ on the fraction of affected channels. \\
    \addlinespace[3pt]
    \multicolumn{2}{@{}l}{\textit{Training and selection}} \\
    \addlinespace[1pt]
    Training sample budget & $n_{\mathrm{train}}=10{,}000$ and $n_{\mathrm{val}}=3{,}000$ sampled windows per configuration. \\
    Tuning regime & At most 40 candidate settings from a seeded random subgrid, with at most 200 epochs and early stopping with patience 10. \\
    Batch size & Batch size 64 for \BeijingTiantan{} and \PenmanshielWT{}, and 16 for ETTh1 and Traffic. Smaller batches appear only in reruns limited by memory. \\
    Selection of trained models & Clean validation objective for baseline models and methods trained end-to-end, with deterministic tie-breaking. \\
    Selection of post-hoc wrappers & Clean validation MSE of the wrapper for ensemble aggregation and randomized smoothing, with deterministic tie-breaking. \\
    Seed policy & Default master seed 42, with derived data, model, and evaluation seeds. \\
    \addlinespace[3pt]
    \multicolumn{2}{@{}l}{\textit{Compute}} \\
    \addlinespace[1pt]
    Compute environment & One graphics processing unit (GPU) per job on NVIDIA H100 NVL workers partitioned with Multi-Instance GPU (MIG), run through Kubeflow virtual machines (VMs) with MLflow and MinIO. The worker pool has two 47 GB slices and two 24 GB slices. \\
    Training compute & For the default configuration, $2{,}236$ completed training setup spans, totaling approximately $624$ GPU-hours on single MIG devices (ETTh1 $43.3$, \BeijingTiantan{} $17.6$, Traffic $512.5$, \PenmanshielWT{} $50.3$). \\
    Evaluation compute & For the default configuration, $164$ evaluation spans of selected winners, totaling approximately $17$ GPU-hours on single MIG devices as a lower bound over the canonical metric spans of perturbation evaluation (ETTh1 $0.3$, \BeijingTiantan{} $0.2$, Traffic $15.3$, \PenmanshielWT{} $1.0$). \\
    Training seed sensitivity & The reruns with seeds 1001 and 1002 repeat the training and evaluation of the default configuration (Appendix~\ref{subsec:training_seed_sensitivity}), each with $2{,}236$ training setup spans and $164$ evaluation spans of selected winners. Training takes approximately $696$ and $612$ GPU-hours and evaluation approximately $18.7$ and $16.6$ GPU-hours on single MIG devices. Both reruns together take approximately $1{,}343$ GPU-hours. \\
    ETTh1 task ablation & The reruns for the three ETTh1 task settings beyond the benchmark task (Appendix~\ref{subsec:etth1_task_ablation}) add $633$ training setup spans and $21$ evaluation spans of selected winners, which take approximately $31$ and $0.3$ GPU-hours on single MIG devices, respectively. \\
    Compute ledger per run & The official compute of these experiments is itemized as $7{,}341$ training setup spans and $513$ evaluation spans of selected winners, totaling approximately $2{,}015$ GPU-hours on single MIG devices. These records are keyed by $7{,}461$ unique MLflow parent runs because selected trained winners contribute timing records for both setup and evaluation. \\
    Sensitivity checks & The sensitivity checks in Appendix~\ref{sec:benchmark_design_validation} comprise four further evaluation data seeds for the selected models, the models of the selector pressure check, and the fault variants of Appendices~\ref{subsec:severity_weighting_sensitivity}, \ref{subsec:fault_direction_sensitivity}, and~\ref{subsec:timecompress_dependence}. They add $1{,}132$ evaluation spans and approximately $113$ GPU-hours on single MIG devices, a lower bound like the evaluation compute above. \\
    Recorded project compute audit & The overhead audit adds $5{,}973$ parent-run records from prior, stale, deleted, failed or killed, and meta-analysis runs. The official spans, the sensitivity evaluations, and the recorded overhead total $14{,}959$ records and approximately $3{,}208$ GPU-hours on single MIG devices. \\
    \bottomrule
  \end{tabularx}
\end{table}
\FloatBarrier

The compute ledgers report the duration of each run for training and for the evaluation of selected winners, both in the default configuration and in the reruns with other training seeds and ETTh1 task settings.
They also report each evaluation for the sensitivity checks and separately disclose recorded project compute outside these experiments.
The additional compute is dominated by the preceding experiment rounds, stale or deleted runs from scope revisions, and failed or killed jobs.
The robustness-evaluation compute is a lower bound over the recorded perturbation-evaluation span.
It captures perturbation forward passes, bootstrap confidence-interval computation, and degradation metric logging, while excluding model loading, clean validation passes before the recorded span, artifact uploads, optional diagnostics, and child-run timing for nested stages.
The perturbed validation passes that score the candidate variants for the selector in Appendix~\ref{sec:selector_pressure}, the evaluations with fixed channel fractions in Appendix~\ref{subsec:channel_count_sensitivity}, and the near-tie comparison of PatchTST configurations in Appendix~\ref{subsec:training_seed_sensitivity} recorded no timing spans and are not itemized.

The method-comparison scope is restricted to architectures with paired no-intervention baselines: DLinear, GRU, ModernTCN, PatchTST, and TSMixer.
SeasonalNaive and Chronos-2 enter only the architecture comparison.
Within the five method-comparison architectures, all robustness-improvement methods are evaluated on all architectures except RevIN, which is not evaluated on PatchTST because PatchTST already applies per-window instance normalization and de-normalization.
Thus each dataset contributes 34 method-baseline pairs to the method tables: five pairs over DLinear, GRU, ModernTCN, PatchTST, and TSMixer for every robustness-improvement method except RevIN, which contributes four.
This scope keeps every method-baseline delta tied to a trained paired baseline forecasting model rather than averaging over architectures for which no comparable improved variant is evaluated.
Table~\ref{tab:randomness_selection_accounting} summarizes how the fixed seeds, selectors, resampling choices, and reported quantities align across this protocol.

\begin{table}[!tb]
  \centering
  \small
  \setlength{\tabcolsep}{4.5pt}
  \renewcommand{\arraystretch}{1.14}
  \caption{Selection, randomness, and resampling accounting for the benchmark protocol.}
  \label{tab:randomness_selection_accounting}
  \begin{tabularx}{0.98\textwidth}{@{}
    p{0.27\textwidth}
    X
    X
    @{}}
    \toprule
    Component & Fixed or random object & Role in reported quantities \\
    \midrule
    \multicolumn{3}{@{}l}{\textit{Training and selection}} \\
    \addlinespace[1pt]
    Data split and sampled training windows & Data seed derived from the master seed and the signature of each dataset's data configuration. & Fixes chronological splits and sampled training and validation windows independently of architecture or method. \\
    Model fitting and tuning & Model seed derived from the master seed and seeded random subgrid over the versioned hyperparameter grids. & Determines trained candidates within the common tuning budget. \\
    Selection of trained models & Clean validation objective with deterministic tie-breaking. & Selects winners among baselines and methods applied during training before robustness evaluation. \\
    Selection of wrappers & Clean validation MSE of the wrapper with the same deterministic tie-breaking. & Selects ensemble and randomized smoothing wrappers without using test metrics. \\
    \addlinespace[3pt]
    \multicolumn{3}{@{}l}{\textit{Evaluation and intervals}} \\
    \addlinespace[1pt]
    Realization of robustness evaluation & Evaluation data seed controlling sampled test windows, severity draws, and the auxiliary randomness of each scenario. & Produces the clean and perturbed losses used for $\mathrm{MSE}_c$, $\mathcal{D}_w$, and $\mathrm{MSE}_w$. \\
    Architecture bootstrap intervals & Resampled shared pool of test windows with clean and perturbed summaries recomputed jointly. & Summarizes variation over sampled test windows for fixed selected models, scenario set, and severity sampling distribution. \\
    Method bootstrap intervals & Resampled method-baseline architecture pairs within each dataset and method. & Summarizes variation across the evaluated architecture pairs for the mean deltas of each dataset. \\
    \bottomrule
  \end{tabularx}
\end{table}

Baseline models and methods trained end-to-end are selected by their clean validation objective only.
Post-hoc methods such as ensemble aggregation and the randomized smoothing wrapper are selected by the wrapper's own clean validation MSE under the same deterministic tiebreaking rule.
The test split is reserved for final robustness evaluation and never feeds back into winner selection.
Reported intervals are 95\% percentile bootstrap intervals, following the percentile-interval construction in \citet{efronIntroductionBootstrap1994}, and are conditional on the selected models.
For per-run degradation quantities in the architecture comparison, each draw resamples one shared pool of sampled test windows and reuses that pool across the clean summary and all benchmark scenarios before recomputing the reported degradation quantities, including the worst-scenario summary.
Those intervals summarize variability over the sampled test-window pool under the fixed benchmark scenario set and severity sampling distribution, rather than a second layer of scenario uncertainty.
For dataset-level method means, each draw resamples the method-baseline architecture pairs within a dataset and method, then recomputes the mean delta.
Score-family comparisons are reported separately and do not alter the main result tables.

\section{Benchmark workflow and repository guide}
\label{app:repo_guide}

The public code and benchmark repository\footnote{\SensorFaultBenchRepo{}} defines the interface for reproducing and extending the benchmark.
A clean clone, together with the documented dataset files under \texttt{DATA\_ROOT}, can run a smoke benchmark, execute the configured benchmark scope, and regenerate analysis tables and figures from the documented entry points.
The repository documentation covers installation, data acquisition, object-store configuration, artifact locations, and troubleshooting.
The implementation uses Python with PyTorch and PyTorch Lightning for model training and MLflow for experiment tracking and artifact logging~\citep{paszkePyTorchImperativeStyle2019,falconPyTorchLightning2026,zahariaAcceleratingMachineLearning2018}.
Exact dependency resolution for the Python benchmark environment, including the scientific Python stack and Plotly diagnostics, is recorded in \path{pyproject.toml} and \path{uv.lock}.
The polished paper figures are rendered from an optional R/TikZ subartifact under \path{paper_artifact/R/}, its R packages and TeX tools are documented separately in \path{paper_artifact/R/README.md} and are intentionally outside the \texttt{uv sync} environment.
Datasets, forecasting architectures, and robustness-improvement methods are selected by stable keys, so the same training, testing, and analysis commands produce comparable tables and figures.

The workflow is YAML-first: \texttt{configs/defaults.yaml} defines the default datasets, architecture scope, method execution order, and perturbation scenarios, while \texttt{configs/dataset\_windows.yaml} resolves dataset-specific horizon overrides.
The benchmark-scope manifest \texttt{configs/benchmark\_scope.yaml} records display labels, architecture roles, and method-architecture applicability for the repository surface.
The benchmark defaults span all four datasets, all benchmark architectures, and all evaluated robustness-improvement methods.
Local training and testing runs can narrow that scope with \texttt{--data-files}, \texttt{--model}, and \texttt{--method}.
Registered datasets are resolved under a configurable \texttt{DATA\_ROOT}.
Local filesystem paths are the default, and object-store paths can be supplied explicitly for a remote execution profile.

From the clone root, the README smoke path narrows the dataset, model, method, and scenario scope explicitly:
\begin{lstlisting}[style=repoShell]
uv sync
uv run python run_training.py --data-files ETTh1 --model DLinear --method baseline --max-epochs 1 --max-hp-trials-per-model 1 --logdir runs
uv run python run_testing.py --data-files ETTh1 --model DLinear --method baseline --perturbation-scenarios missing_data noise --max-hp-trials-per-model 1 --logdir runs
uv run python run_analysis.py --data-files ETTh1 --model DLinear --method baseline --perturbation-scenarios missing_data noise --max-hp-trials-per-model 1 --logdir runs
\end{lstlisting}

The training command fits one baseline forecaster under the smoke scope.
The testing command evaluates the selected winner over the requested scenario subset under the same perturbation semantics as the full benchmark.
The analysis command writes result tables and figures for the tested run set.
The full benchmark workflow uses the configured defaults and strict coverage checks:
\begin{lstlisting}[style=repoShell]
uv run python run_training.py
uv run python run_testing.py --full-coverage
uv run python run_analysis.py --full-coverage
\end{lstlisting}

The no-argument training command expands across all benchmark datasets, architectures, and applicable methods under the configured method-architecture matrix.
For exact reproduction of the training profile of the default configuration, use the batch-size policy and tuning regime in Table~\ref{tab:experimental_protocol}.
Extensions may use larger or architecture-specific tuning budgets when needed, but those runs should report the changed tuning regime relative to the default configuration in Table~\ref{tab:experimental_protocol}.
The extension interface is repository-level and key-based: each new component contributes a stable key plus the owner files that downstream training, testing, and analysis consume.
This structure makes the benchmark reusable as a controlled comparison harness rather than as a fixed list of models.

\paragraph{Evaluation under limited compute.}
We recommend starting with \BeijingTiantan{} and ETTh1, which cover forecasting one target and forecasting all channels.
Together, these datasets account for approximately 61 of the 624 recorded training GPU-hours of the default configuration in Table~\ref{tab:experimental_protocol}.
A useful first comparison evaluates multiple architectures with and without the new method, ideally all five compatible architectures, while retaining all eight scenarios and all three reported quantities.
Direct comparison with the default configuration requires all four datasets.

\paragraph{Dataset interface.}
A dataset key resolves to a file under \texttt{DATA\_ROOT} plus registry metadata for split mode, target alias, input and target channels, continuous and discrete channel typing, and dataset-specific window defaults.
Curated or format-normalized datasets additionally provide deterministic preprocessing and validation paths.
For Traffic, the public benchmark CSV is validated and converted to the Parquet runtime file used by the benchmark without changing the sensor series.
The conversion keeps the runtime dataset compact and fast to load while preserving the configured input and target channels.
Those fields, rather than the storage location, determine split semantics and configuration-signature comparability.

\paragraph{Architecture interface.}
Forecasting architectures follow the shared tensor convention for time series with axes for batch, input window, and input channels, so robustness testing can query every selected checkpoint through its prediction path.
A new architecture added through \texttt{docs/extending\_models.md} joins the same validation selector, checkpoint-loading path, applicability checks, and scenario evaluation loop as the existing comparison set.

\paragraph{Method interface.}
A robustness-improvement method enters through the recipe and method-key path documented in \texttt{docs/extending\_methods.md}.
The method metadata fixes whether the implementation is applied during training or as a post-hoc wrapper, which architectures it applies to, and how testing reconstructs the selected variant.
Paired comparisons then use the same baseline matching, selection metric, run tags, and delta analysis as the evaluated methods.
The transfer fault families reserved for training can also provide paired clean and corrupted histories with annotations for the fault and affected channels.
These pairs could supervise reconstruction of the clean history before forecasting or consistency of internal representations.
The evaluated use is fault augmentation, while those architectural uses remain open and the disjoint scored scenario set remains reserved for evaluation.
Unsupported method-architecture pairs, unknown keys, and missing metadata are rejected during scope validation, so extension errors do not change the evaluated benchmark scope.
Table~\ref{tab:artifact_surface} summarizes the repository surfaces that implement this reproduction and extension interface.

\begin{table}[!htb]
  \centering
  \footnotesize
  \setlength{\tabcolsep}{4.5pt}
  \renewcommand{\arraystretch}{1.15}
  \caption{Repository surface for reproducing and extending the benchmark.}
  \label{tab:artifact_surface}
  \begin{tabularx}{0.98\textwidth}{@{}
    >{\RaggedRight\arraybackslash}p{0.17\textwidth}
    >{\RaggedRight\arraybackslash}p{0.45\textwidth}
    X
    @{}}
    \toprule
    Component & Repository surface & Purpose \\
    \midrule
    Source repository & Repository, with setup and workflow described by \path{README.md} and \path{docs/reproducing_benchmark.md}. & Defines the source tree and reproduction workflow. \\
    Workflow entry points & \path{run_training.py}, \path{run_testing.py}, and \path{run_analysis.py}. & Train the configured scope, evaluate selected winners on the scored benchmark scenarios, and build result tables and figures. \\
    Data access & \path{data/README.md}, upstream source links, deterministic preprocessing scripts for \BeijingTiantan{}, \PenmanshielWT{}, and the Traffic CSV-to-Parquet conversion, and validation commands. & Documents acquisition, format normalization, or reconstruction of the four datasets under \texttt{DATA\_ROOT}. \\
    Benchmark scope and configs & \path{configs/defaults.yaml}, \path{configs/dataset_windows.yaml}, \path{configs/baseline_hparams.yaml}, \path{configs/pipelines/*.yaml}, and \path{configs/benchmark_scope.yaml}. & Records runtime defaults, dataset windows, tuning grids, recipes, display labels, and which methods apply to which architectures. \\
    Analysis artifacts & \path{run_analysis.py} outputs under \path{tables/}, \path{figures/}, \path{tables/figures_manifest.csv}, and \path{config/meta_analysis_args.yaml}. & Provides generated tables and figures for a tested run set. \\
    Supplementary tables & \path{paper_artifact/supplementary_tables/} and \path{paper_artifact/MANIFEST.json}. & Provides selected clean-validation hyperparameters, complete three-training-seed architecture and method results, the PatchTST near-tie diagnostic, and the tables of the severity-weighting, fault-direction, timing, and ETTh1 task checks. \\
    Compute accounting & \path{compute_ledger_official_runs.csv}, \path{compute_evaluation_episodes.csv}, \path{compute_aggregates.csv}, and \path{compute_overhead_audit.csv} under \path{paper_artifact/compute_accounting/}, written by \path{scripts/compute_accounting_ledger.py}. & Provides timing records for each run and evaluation episode, aggregate compute summaries for each experiment, and the overhead audit from the live MLflow server for preliminary, stale, deleted, failed, or killed parent runs. \\
    Extension documentation & \path{docs/extending_datasets.md}, \path{docs/extending_models.md}, and \path{docs/extending_methods.md}. & Specifies the registry, metadata, grid, recipe, and applicability fields needed to add datasets, forecasting architectures, or robustness-improvement methods. \\
    Fault operators & \path{data/perturbations.py} and \path{docs/using_fault_operators.md}. & Implement the scored scenarios and transfer fault families and document the parameters that can be calibrated to the recorded faults of a specific system. \\
    Licenses and notices & \path{LICENSE}, \path{THIRD_PARTY_NOTICES.md}, and \path{data/README.md}. & Separates the Apache-2.0 code license from upstream dataset terms and third-party source notices. \\
    \bottomrule
  \end{tabularx}
\end{table}

\clearpage
\section{Extended results}
\label{app:extended_results}

The extended results provide the per-dataset evidence cited from the Results section.

\subsection{Architectures and scenarios by dataset}
\label{app:per_dataset_arch}

\begin{figure}[!htb]
  \centering
  \includegraphics[width=0.97\textwidth]{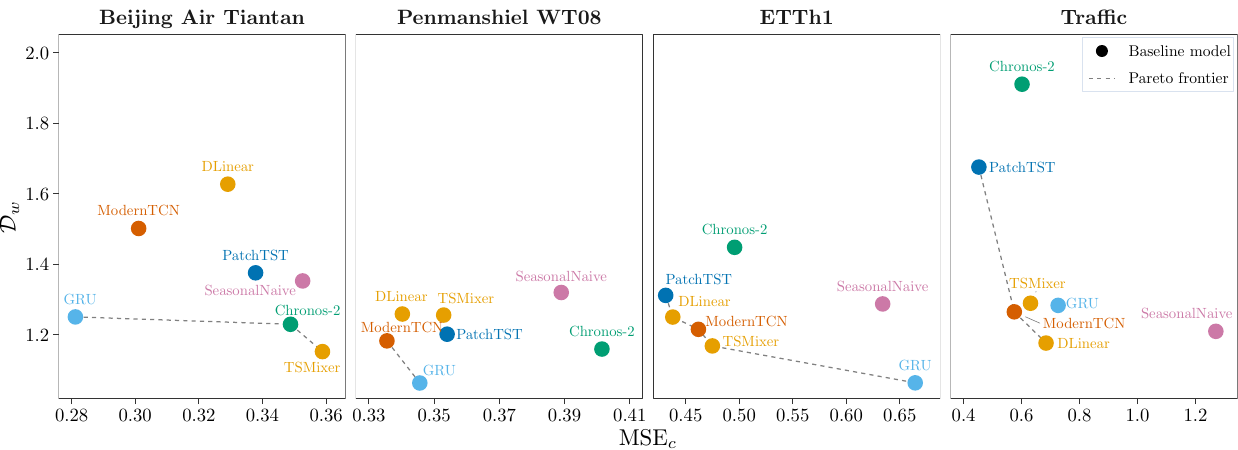}
  \includegraphics[width=0.97\textwidth]{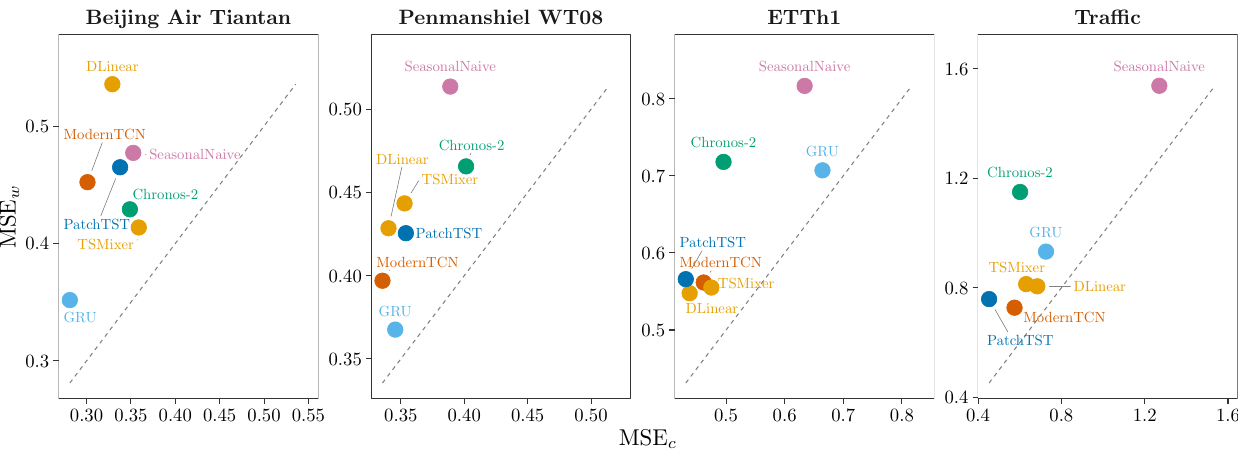}
  \caption{Additional architecture trade-off views for the baseline setting.
  The top row plots clean MSE $\mathrm{MSE}_c$ against worst-scenario degradation $\mathcal{D}_w$ (Eq.~\eqref{eq:dmax}) for the evaluated forecasting architectures within each dataset.
  The bottom row plots $\mathrm{MSE}_c$ against worst-scenario fault-time MSE $\mathrm{MSE}_w$ for the same architectures.
  Together these views show where normalized degradation and absolute fault-time error lead to different architecture frontiers.}
  \label{fig:appendix_architecture_tradeoffs}
\end{figure}

Figure~\ref{fig:appendix_architecture_tradeoffs} and Table~\ref{tab:appendix_architecture_score_ci} show the trade-offs between quantities and the uncertainty around the leading rankings for the result values of each dataset in Table~\ref{tab:main_results}.
Mean-case and score-family alternatives are collected separately in Table~\ref{tab:score_family_architecture_comparison}.

Availability scenarios produce the largest Chronos-2 degradation on ETTh1 and Traffic, with StuckSensor as the worst scenario on both datasets.
For ETTh1, the largest degradation of Chronos-2 in the Value, Timing, and Availability classes is 1.146, 1.181, and 1.448, while for Traffic the values are 1.117, 1.317, and 1.911, respectively.
The constant segment of StuckSensor carries no fault indicator and removes variation that conveys recent dynamics, offering a plausible explanation for this concentration.
Chronos-2 is evaluated without task-specific adaptation, so it must distinguish sensor failure from stable process behavior through context.
This dependence on contextual inference may hinder recovery relative to architectures trained on each task, although the aggregate results do not isolate the mechanism.
Representation analysis under clean and corrupted inputs could test this mechanism, as could an ablation that gives the model an explicit fault mask and measures whether fault information improves recovery.

\begin{table}[!htb]
  \centering
  \small
  \setlength{\tabcolsep}{3pt}
  \renewcommand{\arraystretch}{1.12}
  \caption{Bootstrap intervals for the leading architecture rankings in the baseline setting.
  Rows compare the best and runner-up forecasting architectures for each reported score in Table~\ref{tab:main_results}.
  Parentheses give 95\% percentile bootstrap intervals over the sampled test-window budget.}
  \label{tab:appendix_architecture_score_ci}
  \newcommand{\archdatasetlabel}[1]{\multirow{3}{*}{\centering #1}}
  \newcommand{\scoreci}[3]{\num{#1}\,(\num{#2},\,\num{#3})}
  \begin{tabular*}{0.98\textwidth}{@{\extracolsep{\fill}}>{\centering\arraybackslash}p{0.18\textwidth} l l c l c@{}}
    \toprule
    Data & Score & Winner & Estimate (95\% CI) & Runner-up & Estimate (95\% CI) \\
    \midrule
    \archdatasetlabel{\BeijingTiantan{}} & $\mathcal{D}_w$ & TSMixer & \scoreci{1.153}{1.119}{1.197} & Chronos-2 & \scoreci{1.230}{1.190}{1.296} \\
     & $\mathrm{MSE}_c$ & GRU & \scoreci{0.281}{0.263}{0.301} & ModernTCN & \scoreci{0.301}{0.282}{0.322} \\
     & $\mathrm{MSE}_w$ & GRU & \scoreci{0.352}{0.335}{0.377} & TSMixer & \scoreci{0.414}{0.390}{0.442} \\
    \addlinespace[2pt]
    \archdatasetlabel{\PenmanshielWT{}} & $\mathcal{D}_w$ & GRU & \scoreci{1.064}{1.056}{1.072} & Chronos-2 & \scoreci{1.160}{1.146}{1.178} \\
     & $\mathrm{MSE}_c$ & ModernTCN & \scoreci{0.336}{0.324}{0.346} & DLinear & \scoreci{0.340}{0.329}{0.351} \\
     & $\mathrm{MSE}_w$ & GRU & \scoreci{0.368}{0.357}{0.378} & ModernTCN & \scoreci{0.397}{0.384}{0.409} \\
    \addlinespace[2pt]
    \archdatasetlabel{ETTh1} & $\mathcal{D}_w$ & GRU & \scoreci{1.064}{1.059}{1.069} & TSMixer & \scoreci{1.169}{1.164}{1.177} \\
     & $\mathrm{MSE}_c$ & PatchTST & \scoreci{0.431}{0.428}{0.435} & DLinear & \scoreci{0.438}{0.434}{0.442} \\
     & $\mathrm{MSE}_w$ & DLinear & \scoreci{0.548}{0.542}{0.553} & TSMixer & \scoreci{0.555}{0.551}{0.560} \\
    \addlinespace[2pt]
    \archdatasetlabel{Traffic} & $\mathcal{D}_w$ & DLinear & \scoreci{1.177}{1.171}{1.183} & SeasonalNaive & \scoreci{1.210}{1.198}{1.221} \\
     & $\mathrm{MSE}_c$ & PatchTST & \scoreci{0.453}{0.450}{0.456} & ModernTCN & \scoreci{0.575}{0.572}{0.578} \\
     & $\mathrm{MSE}_w$ & ModernTCN & \scoreci{0.728}{0.723}{0.732} & PatchTST & \scoreci{0.759}{0.751}{0.767} \\
    \bottomrule
  \end{tabular*}
\end{table}

\subsubsection{Dataset grouping interpretation}
\label{subsec:regime_interpretation}

The scenario-level degradation pattern in Figure~\ref{fig:baseline_architecture_heatmap} supports a descriptive grouping between the two single-target datasets with 96-step inputs and 6-step horizons and the two all-channel datasets with 96-step inputs and 96-step horizons.
\BeijingTiantan{} and \PenmanshielWT{} use single-target outputs with low to medium redundancy across channels, whereas ETTh1 and Traffic use all-channel outputs with 7 and 862 targets respectively.
The grouped pairs therefore vary forecast horizon together with target multiplicity, channel count, and perturbation coupling rather than isolating horizon alone.
The perturbation protocol also treats MissingData differently from the other benchmark scenarios because it removes all channels at once, whereas the other benchmark scenarios are capped at a severity-dependent subset of the continuous channels.
Within this design, \BeijingTiantan{} and \PenmanshielWT{} expose stronger value-corruption sensitivity and larger method separation, whereas ETTh1 and Traffic compress many pairwise gaps and shift the worst scenarios toward availability and timing faults.
The ablation within ETTh1 in Appendix~\ref{subsec:etth1_task_ablation} varies target scope and horizon with a fixed number of input channels, but does not explain differences across datasets.

\subsection{Effects of robustness-improvement methods by dataset}
\label{sec:improv_effects}

\subsubsection{RevIN under persistent faults}

For the models in Table~\ref{tab:main_results}, RevIN increases mean scenario degradation under Drift and StuckSensor on all four datasets, and StuckSensor also increases mean absolute fault-time MSE.
RevIN computes a mean and standard deviation separately for each channel over the observed window.
A persistent fault can alter these statistics and hence the representation throughout that channel's history.
When the affected channel is also forecast, its statistics determine the restored output level and scale.
A smaller clean MSE in the denominator contributes to degradation but cannot explain the increase in absolute fault-time error.
Altered window statistics could therefore explain the observed pattern, although these comparisons do not isolate the mechanism.

\subsubsection{PGD trajectory details}
\label{sec:pgd_trajectory}

\begin{figure}[!htb]
  \centering
  \includegraphics[width=1\textwidth]{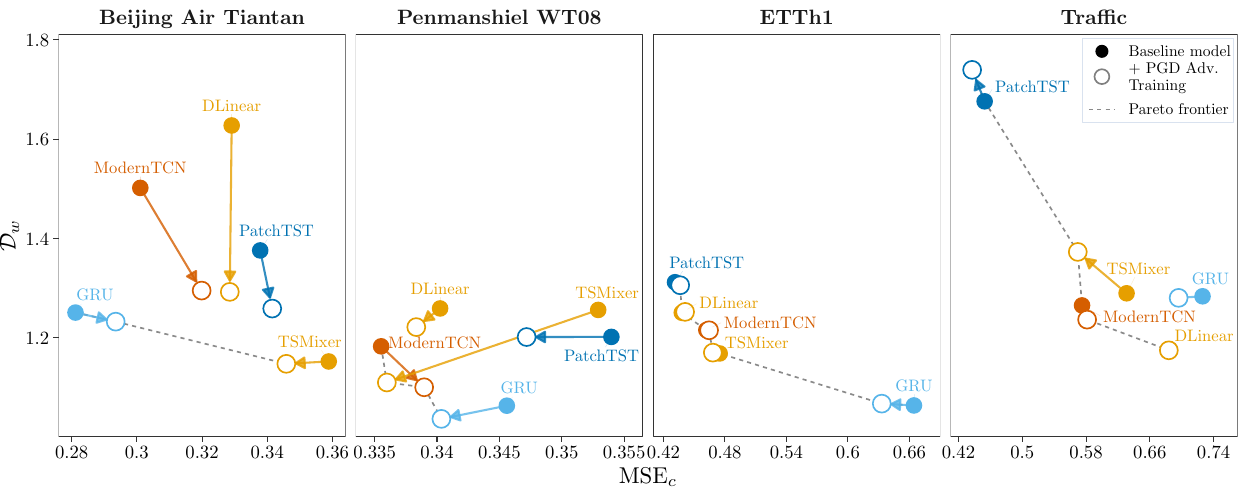}
  \caption{PGD adversarial training trajectories across the four datasets.
  Each panel plots clean MSE $\mathrm{MSE}_c$ against worst-scenario degradation $\mathcal{D}_w$ (Eq.~\eqref{eq:dmax}) for paired baseline forecasting models and selected PGD adversarial training variants.
  Filled markers denote paired baselines, hollow markers denote the corresponding PGD variants, and arrows show the baseline-to-method shift for each architecture.
  The dashed line marks the Pareto frontier within each dataset panel.}
  \label{fig:pgd_trajectory_dw}
\end{figure}

Table~\ref{tab:main_results} gives dataset-level means, while Figure~\ref{fig:pgd_trajectory_dw} shows the paired changes for PGD adversarial training.
On \PenmanshielWT{}, four of five pairs move down and left, so PGD lowers worst-scenario degradation together with clean MSE for most paired architectures.
PGD thereby moves the Pareto frontier of the panel in the same direction, with its variants forming three of the four frontier points.
\BeijingTiantan{} shows the same downward pattern in worst-scenario degradation, but usually with a modest rightward clean-error cost.
ETTh1 remains near-neutral, and Traffic is the clearest split case: PatchTST and TSMixer move left and up, whereas the other three pairs stay flat or move downward in worst-scenario degradation.
The trajectory view therefore supports a scoped interpretation: PGD adversarial training has its clearest reductions where value faults produce the largest observed degradation on the two single-target datasets, but does not produce one architecture-independent direction across all four datasets.

\subsubsection{Method-delta interval details}
\label{sec:method_delta_intervals}

Table~\ref{tab:appendix_method_delta_ci} gives the full four-dataset interval view for the method-baseline deltas summarized in Table~\ref{tab:main_results}.
These intervals are conservative stability summaries over the evaluated method-baseline architecture pairs, not population-level significance tests.
When an interval crosses zero, the corresponding method-baseline delta is not sign-stable for that dataset and reported quantity, so we do not treat it as evidence of a transferable improvement.

\begin{table}[!htb]
  \centering
  \footnotesize
    \newcommand{\BeijingTiantanRot}{%
    \begin{tabular}{@{}c@{}}
      Beijing Air\\
      Tiantan
    \end{tabular}%
  }

  \newcommand{\PenmanshielWTRot}{%
    \begin{tabular}{@{}c@{}}
      Penmanshiel\\
      WT08
    \end{tabular}%
  }
  \caption{Bootstrap intervals for method-baseline delta stability by dataset and robustness-improvement method.
  Each metric cell reports the mean delta followed by its 95\% percentile bootstrap interval in parentheses, computed by resampling method-baseline architecture pairs within the method and dataset.
  Pairs gives the number of method-baseline architecture pairs.
  Intervals whose displayed endpoints have the same nonzero sign do not include zero.
  Intervals summarize variation across the evaluated paired architecture set rather than population-level significance.
  PGD denotes PGD adversarial training, Rand. Train randomized training, Rand. Smooth randomized smoothing, ARL adaptive robust loss, and Fault Aug. fault augmentation.
  Lower values are better, and negative deltas favor the robustness-improved variant.}
  \label{tab:appendix_method_delta_ci}
  \setlength{\tabcolsep}{1.2pt}
  \renewcommand{\arraystretch}{1.06}
  \newcommand{\methodcitwo}[3]{%
    \makebox[3.10em][r]{\ensuremath{#1}}%
    \hspace{0.25em}%
    \makebox[0.80em][c]{\ensuremath{(}}%
    \makebox[3.10em][r]{\ensuremath{#2}}%
    \ensuremath{,}%
    \hspace{0.20em}%
    \makebox[3.10em][r]{\ensuremath{#3}}%
    \makebox[0.80em][c]{\ensuremath{)}}%
  }
  \newcommand{\methoddatasetlabel}[1]{\rotatebox[origin=c]{90}{#1}}
  \begin{tabularx}{\textwidth}{@{}
    >{\centering\arraybackslash}p{0.050\textwidth}
    @{\hspace{0.9em}}
    >{\RaggedRight\arraybackslash}p{0.130\textwidth}
    >{\centering\arraybackslash}p{0.040\textwidth}
    *{3}{>{\centering\arraybackslash}p{0.241\textwidth}}
    @{}}
    \toprule
    Data & Method & Pairs & $\Delta\!\mathcal{D}_w$ & $\Delta\!\mathrm{MSE}_c$ & $\Delta\!\mathrm{MSE}_w$ \\
    \midrule
    \multirow{7}{*}{\methoddatasetlabel{\BeijingTiantanRot{}}} & PGD & 5 & \methodcitwo{-0.136}{-0.246}{-0.033} & \methodcitwo{0.004}{-0.006}{0.014} & \methodcitwo{-0.038}{-0.081}{-0.009} \\
     & Ensemble & 5 & \methodcitwo{0.007}{-0.011}{0.026} & \methodcitwo{-0.009}{-0.026}{0.003} & \methodcitwo{-0.009}{-0.023}{0.005} \\
     & Rand. Train & 5 & \methodcitwo{-0.146}{-0.246}{-0.048} & \methodcitwo{0.026}{0.003}{0.066} & \methodcitwo{-0.018}{-0.074}{0.041} \\
     & Rand.~Smooth & 5 & \methodcitwo{0.001}{-0.006}{0.006} & \methodcitwo{0.002}{-0.002}{0.005} & \methodcitwo{0.003}{0.000}{0.006} \\
     & ARL & 5 & \methodcitwo{-0.024}{-0.075}{0.027} & \methodcitwo{0.001}{-0.025}{0.034} & \methodcitwo{-0.005}{-0.032}{0.040} \\
     & RevIN & 4 & \methodcitwo{0.053}{-0.127}{0.233} & \methodcitwo{0.005}{-0.014}{0.020} & \methodcitwo{0.025}{-0.030}{0.075} \\
     & Fault Aug. & 5 & \methodcitwo{-0.039}{-0.062}{-0.011} & \methodcitwo{0.005}{-0.001}{0.013} & \methodcitwo{-0.007}{-0.021}{0.011} \\
    \addlinespace
    \multirow{7}{*}{\methoddatasetlabel{\PenmanshielWTRot{}}} & PGD & 5 & \methodcitwo{-0.059}{-0.104}{-0.018} & \methodcitwo{-0.005}{-0.012}{0.000} & \methodcitwo{-0.026}{-0.049}{-0.012} \\
     & Ensemble & 5 & \methodcitwo{-0.023}{-0.072}{0.008} & \methodcitwo{-0.007}{-0.010}{-0.003} & \methodcitwo{-0.016}{-0.032}{-0.007} \\
     & Rand. Train & 5 & \methodcitwo{-0.036}{-0.062}{-0.011} & \methodcitwo{-0.003}{-0.006}{0.000} & \methodcitwo{-0.016}{-0.022}{-0.010} \\
     & Rand.~Smooth & 5 & \methodcitwo{-0.002}{-0.004}{0.001} & \methodcitwo{0.003}{0.001}{0.005} & \methodcitwo{0.003}{0.001}{0.005} \\
     & ARL & 5 & \methodcitwo{0.144}{-0.013}{0.370} & \methodcitwo{0.040}{0.019}{0.065} & \methodcitwo{0.102}{0.026}{0.183} \\
     & RevIN & 4 & \methodcitwo{0.110}{0.008}{0.292} & \methodcitwo{-0.001}{-0.005}{0.003} & \methodcitwo{0.037}{0.003}{0.101} \\
     & Fault Aug. & 5 & \methodcitwo{-0.019}{-0.058}{0.008} & \methodcitwo{-0.001}{-0.005}{0.002} & \methodcitwo{-0.008}{-0.019}{-0.001} \\
    \addlinespace
    \multirow{7}{*}{\methoddatasetlabel{ETTh1}} & PGD & 5 & \methodcitwo{0.000}{-0.003}{0.003} & \methodcitwo{-0.005}{-0.020}{0.004} & \methodcitwo{-0.005}{-0.019}{0.004} \\
     & Ensemble & 5 & \methodcitwo{-0.005}{-0.010}{0.001} & \methodcitwo{-0.005}{-0.008}{-0.001} & \methodcitwo{-0.007}{-0.010}{-0.003} \\
     & Rand. Train & 5 & \methodcitwo{-0.004}{-0.012}{0.002} & \methodcitwo{0.009}{0.001}{0.021} & \methodcitwo{0.008}{0.001}{0.020} \\
     & Rand.~Smooth & 5 & \methodcitwo{0.000}{-0.002}{0.001} & \methodcitwo{-0.001}{-0.003}{0.000} & \methodcitwo{-0.001}{-0.003}{0.000} \\
     & ARL & 5 & \methodcitwo{0.001}{-0.013}{0.012} & \methodcitwo{0.033}{0.006}{0.068} & \methodcitwo{0.037}{0.012}{0.067} \\
     & RevIN & 4 & \methodcitwo{0.124}{0.066}{0.183} & \methodcitwo{-0.045}{-0.110}{-0.004} & \methodcitwo{0.010}{-0.023}{0.031} \\
     & Fault Aug. & 5 & \methodcitwo{-0.001}{-0.005}{0.004} & \methodcitwo{0.003}{-0.002}{0.009} & \methodcitwo{0.002}{-0.002}{0.009} \\
    \addlinespace
    \multirow{7}{*}{\methoddatasetlabel{Traffic}} & PGD & 5 & \methodcitwo{0.023}{-0.008}{0.062} & \methodcitwo{-0.020}{-0.043}{-0.001} & \methodcitwo{-0.016}{-0.031}{-0.003} \\
     & Ensemble & 5 & \methodcitwo{-0.019}{-0.060}{0.014} & \methodcitwo{-0.017}{-0.035}{-0.004} & \methodcitwo{-0.037}{-0.075}{-0.003} \\
     & Rand. Train & 5 & \methodcitwo{0.011}{-0.019}{0.035} & \methodcitwo{-0.007}{-0.032}{0.019} & \methodcitwo{-0.003}{-0.026}{0.019} \\
     & Rand.~Smooth & 5 & \methodcitwo{-0.021}{-0.063}{0.000} & \methodcitwo{0.011}{-0.005}{0.036} & \methodcitwo{0.008}{-0.007}{0.028} \\
     & ARL & 5 & \methodcitwo{-0.007}{-0.060}{0.050} & \methodcitwo{0.075}{0.053}{0.096} & \methodcitwo{0.100}{0.054}{0.155} \\
     & RevIN & 4 & \methodcitwo{0.267}{0.186}{0.342} & \methodcitwo{-0.063}{-0.094}{-0.028} & \methodcitwo{0.071}{0.049}{0.092} \\
     & Fault Aug. & 5 & \methodcitwo{-0.025}{-0.102}{0.028} & \methodcitwo{-0.013}{-0.036}{0.008} & \methodcitwo{-0.026}{-0.054}{0.002} \\
    \bottomrule
  \end{tabularx}
\end{table}

The score-family method comparison in Table~\ref{tab:score_family_method_comparison} refines the worst-scenario interpretation in the main results.
Its mean-case point estimates keep the same fault-family reading as the main table: PGD adversarial training and randomized training reduce mean degradation most where value faults dominate the single-target profiles, while ensemble remains the broadest low-assumption comparator.
Randomized smoothing and fault augmentation are more visible on Traffic's availability-heavy profile, with randomized smoothing trading that relative-degradation gain for higher mean corrupted MSE.
RevIN does not improve mean degradation on any of the four datasets, although ETTh1 and Traffic still benefit in mean corrupted MSE through clean-error gains rather than improved relative robustness.
Fault augmentation remains mild, with mean-case degradation gains clearest on Traffic and \BeijingTiantan{} and mean corrupted MSE improvement most visible on Traffic.
Adaptive robust loss remains mixed or secondary in the mean case.

\FloatBarrier
\begingroup
\makeatletter
\setlength{\@fptop}{0pt}
\setlength{\@fpsep}{\floatsep}
\makeatother

\subsection{Training seed sensitivity}
\label{subsec:training_seed_sensitivity}

All 12 architecture leaders persist across training and search seeds 42, 1001, and 1002, where seed 42 supplies the models in Table~\ref{tab:main_results}.
We repeated training and clean validation selection for the five learned architectures within the same tuning budget, keeping the dataset split and evaluation sampling fixed within each dataset.
Chronos-2 and SeasonalNaive remain fixed references.
Each seed contributes 28 architecture results and 136 method-baseline pairs across four datasets, with RevIN excluding PatchTST.
The three seeds probe variation from training and search, separately from the conditional bootstrap intervals over windows and the analysis of evaluation data seeds in Appendix~\ref{subsec:eval_data_seed_sensitivity}.

\subsubsection{Architecture leaders and magnitude variation}

For 18 of the 20 pairs of retrained architecture and dataset, the range of $\mathcal{D}_w$ is at most $0.118$ (Table~\ref{tab:training_seed_ranges}).
The larger ranges are $2.113$ for PatchTST on \BeijingTiantan{} and $0.342$ for TSMixer on \PenmanshielWT{}.
The ranges of clean MSE have median $0.007$ and maximum $0.052$, with 18 of 20 no larger than $0.020$.
The leaders thus persist while exact rankings and degradation magnitudes vary.
Table~\ref{tab:training_seed_architecture} gives all three quantities for every architecture at each seed.

\begin{table}[!htb]
  \centering
  \small
  \setlength{\tabcolsep}{7pt}
  \renewcommand{\arraystretch}{1.10}
  \caption{Ranges of worst-scenario degradation across three training and search seeds for the five retrained architectures.
  Each range is the maximum minus the minimum over seeds 42, 1001, and 1002, without bootstrap resampling.} 
  \label{tab:training_seed_ranges}
  \begin{tabular}{@{}l*{3}{S[table-format=1.3]}S[table-format=<1.3]@{}}
    \toprule
Architecture & {\BeijingTiantan{}} & {\PenmanshielWT{}} & {ETTh1} & {Traffic} \\
    \midrule
    DLinear & 0.017 & 0.037 & 0.004 & <0.001 \\
    GRU & 0.059 & 0.018 & 0.028 & 0.097 \\
    ModernTCN & 0.076 & 0.033 & 0.004 & 0.041 \\
    TSMixer & 0.045 & 0.342 & 0.017 & 0.021 \\
    PatchTST & 2.113 & 0.027 & 0.025 & 0.118 \\
    \bottomrule
  \end{tabular}
\end{table}

\begin{table}[!htb]
  \centering
  \small
  \setlength{\tabcolsep}{2.4pt}
  \renewcommand{\arraystretch}{1.10}
  \caption{Architecture results across training and search seeds on all four datasets.
  Lower values are better for all three quantities, and bold marks the best score for each dataset, seed, and quantity.
  Chronos-2 and SeasonalNaive are fixed references across seeds.}
  \label{tab:training_seed_architecture}
  \begin{tabular*}{\textwidth}{@{\extracolsep{\fill}}cl*{9}{S[table-format=1.3,detect-weight=true,mode=text]}@{}}
    \toprule
 & & \multicolumn{3}{c}{Seed 42} & \multicolumn{3}{c}{Seed 1001} & \multicolumn{3}{c}{Seed 1002} \\
\cmidrule(lr){3-5}\cmidrule(lr){6-8}\cmidrule(lr){9-11}
Data & Architecture & {$\mathrm{MSE}_c$} & {$\mathcal{D}_w$} & {$\mathrm{MSE}_w$} & {$\mathrm{MSE}_c$} & {$\mathcal{D}_w$} & {$\mathrm{MSE}_w$} & {$\mathrm{MSE}_c$} & {$\mathcal{D}_w$} & {$\mathrm{MSE}_w$} \\
    \midrule
    \multirow{7}{*}{\rotatebox[origin=c]{90}{\footnotesize \begin{tabular}{@{}c@{}}Beijing Air\\Tiantan\end{tabular}}} & SeasonalNaive & 0.353 & 1.353 & 0.477 & 0.353 & 1.353 & 0.477 & 0.353 & 1.353 & 0.477 \\
     & DLinear & 0.329 & 1.628 & 0.536 & 0.330 & 1.610 & 0.531 & 0.329 & 1.624 & 0.534 \\
     & GRU & \bfseries 0.281 & 1.251 & \bfseries 0.352 & \bfseries 0.285 & 1.255 & \bfseries 0.357 & \bfseries 0.283 & 1.310 & \bfseries 0.371 \\
     & ModernTCN & 0.301 & 1.502 & 0.452 & 0.298 & 1.545 & 0.461 & 0.316 & 1.578 & 0.498 \\
     & TSMixer & 0.359 & \bfseries 1.153 & 0.414 & 0.342 & \bfseries 1.198 & 0.410 & 0.339 & \bfseries 1.171 & 0.397 \\
     & PatchTST & 0.338 & 1.376 & 0.465 & 0.333 & 3.460 & 1.154 & 0.330 & 1.347 & 0.445 \\
     & Chronos-2 & 0.349 & 1.230 & 0.429 & 0.349 & 1.230 & 0.429 & 0.349 & 1.230 & 0.429 \\
    \addlinespace[4pt]
    \multirow{7}{*}{\rotatebox[origin=c]{90}{\footnotesize \begin{tabular}{@{}c@{}}Penmanshiel\\WT08\end{tabular}}} & SeasonalNaive & 0.389 & 1.320 & 0.514 & 0.389 & 1.320 & 0.514 & 0.389 & 1.320 & 0.514 \\
     & DLinear & 0.340 & 1.259 & 0.428 & 0.341 & 1.223 & 0.417 & 0.341 & 1.223 & 0.417 \\
     & GRU & 0.346 & \bfseries 1.064 & \bfseries 0.368 & 0.340 & \bfseries 1.081 & \bfseries 0.367 & 0.342 & \bfseries 1.068 & \bfseries 0.365 \\
     & ModernTCN & \bfseries 0.336 & 1.183 & 0.397 & \bfseries 0.334 & 1.216 & 0.406 & \bfseries 0.335 & 1.214 & 0.406 \\
     & TSMixer & 0.353 & 1.256 & 0.443 & 0.359 & 1.127 & 0.405 & 0.351 & 1.469 & 0.516 \\
     & PatchTST & 0.354 & 1.202 & 0.426 & 0.349 & 1.229 & 0.429 & 0.341 & 1.225 & 0.418 \\
     & Chronos-2 & 0.401 & 1.160 & 0.466 & 0.401 & 1.160 & 0.466 & 0.401 & 1.160 & 0.466 \\
    \addlinespace[4pt]
    \multirow{7}{*}{\rotatebox[origin=c]{90}{\footnotesize ETTh1}} & SeasonalNaive & 0.634 & 1.288 & 0.817 & 0.634 & 1.288 & 0.817 & 0.634 & 1.288 & 0.817 \\
     & DLinear & 0.438 & 1.251 & \bfseries 0.548 & 0.446 & 1.249 & \bfseries 0.556 & 0.438 & 1.253 & \bfseries 0.549 \\
     & GRU & 0.664 & \bfseries 1.064 & 0.707 & 0.675 & \bfseries 1.079 & 0.728 & 0.657 & \bfseries 1.092 & 0.718 \\
     & ModernTCN & 0.462 & 1.216 & 0.562 & 0.461 & 1.218 & 0.561 & 0.460 & 1.220 & 0.561 \\
     & TSMixer & 0.475 & 1.169 & 0.555 & 0.478 & 1.186 & 0.567 & 0.471 & 1.175 & 0.554 \\
     & PatchTST & \bfseries 0.431 & 1.312 & 0.566 & \bfseries 0.435 & 1.287 & 0.559 & \bfseries 0.432 & 1.305 & 0.564 \\
     & Chronos-2 & 0.496 & 1.448 & 0.718 & 0.496 & 1.448 & 0.718 & 0.496 & 1.448 & 0.718 \\
    \addlinespace[4pt]
    \multirow{7}{*}{\rotatebox[origin=c]{90}{\footnotesize Traffic}} & SeasonalNaive & 1.270 & 1.210 & 1.537 & 1.270 & 1.210 & 1.537 & 1.270 & 1.210 & 1.537 \\
     & DLinear & 0.684 & \bfseries 1.177 & 0.806 & 0.684 & \bfseries 1.177 & 0.806 & 0.684 & \bfseries 1.177 & 0.806 \\
     & GRU & 0.726 & 1.284 & 0.932 & 0.740 & 1.187 & 0.878 & 0.752 & 1.228 & 0.923 \\
     & ModernTCN & 0.575 & 1.265 & \bfseries 0.728 & 0.581 & 1.246 & \bfseries 0.724 & 0.575 & 1.287 & \bfseries 0.740 \\
     & TSMixer & 0.631 & 1.290 & 0.814 & 0.579 & 1.310 & 0.758 & 0.618 & 1.311 & 0.810 \\
     & PatchTST & \bfseries 0.453 & 1.676 & 0.759 & \bfseries 0.453 & 1.794 & 0.813 & \bfseries 0.444 & 1.786 & 0.793 \\
     & Chronos-2 & 0.602 & 1.911 & 1.150 & 0.602 & 1.911 & 1.150 & 0.602 & 1.911 & 1.150 \\
    \bottomrule
  \end{tabular*}
\end{table}

\subsubsection{PatchTST configurations with nearly tied clean validation loss}

On \BeijingTiantan{}, PatchTST has $\mathcal{D}_w=1.376$, $3.460$, and $1.347$ at seeds 42, 1001, and 1002, respectively.
Its $\mathrm{MSE}_w$ range is $0.708$, while clean MSE varies only from $0.330$ to $0.338$.
An exploratory comparison at seed 1001 shown in Table~\ref{tab:patchtst_near_ties} evaluates the 14 configurations within 3\% of the best clean validation loss.
The four configurations with embedding width $d_{\mathrm{model}}=128$ and learning rate $10^{-3}$ have Spike as their worst scenario and mean degradation $2.388$.
The remaining ten have Noise as their worst scenario and average $1.332$, despite similar mean clean test MSE in the two groups ($0.340$ and $0.336$).
At seed 1001, clean validation selects one of the four configurations whose worst scenario is Spike, whereas the models selected at seeds 42 and 1002 use $d_{\mathrm{model}}=16$ and learning rate $5\times10^{-4}$ and have Noise as their worst scenario.

\begin{table}[!tb]
  \centering
  \small
  \setlength{\tabcolsep}{6pt}
  \renewcommand{\arraystretch}{1.10}
  \caption{PatchTST configurations within an exploratory 3\% band of the best clean validation loss on \BeijingTiantan{} at seed 1001.
  Rank and relative loss gap refer to clean validation.
  The remaining columns give embedding width, learning rate, test quantities, and worst scenario.
  Rank 1 is the model selected on clean validation loss.
  Other hyperparameters can also differ between configurations.}
  \label{tab:patchtst_near_ties}
  \begin{tabular}{@{}cS[table-format=1.2]cc*{3}{S[table-format=1.3,table-number-alignment=center,table-text-alignment=center]}c@{}}
    \toprule
Rank & {Gap (\%)} & $d_{\mathrm{model}}$ & Learning rate & {$\mathrm{MSE}_c$} & {$\mathcal{D}_w$} & {$\mathrm{MSE}_w$} & Worst scenario \\
    \midrule
    1 & 0.00 & 128 & \num[print-unity-mantissa=false]{1e-3} & 0.333 & 3.460 & 1.154 & Spike \\
    2 & 0.72 & 16 & \num{5e-4} & 0.342 & 1.332 & 0.456 & Noise \\
    3 & 0.80 & 128 & \num[print-unity-mantissa=false]{1e-3} & 0.336 & 1.742 & 0.586 & Spike \\
    4 & 0.94 & 16 & \num[print-unity-mantissa=false]{1e-4} & 0.334 & 1.315 & 0.440 & Noise \\
    5 & 0.95 & 16 & \num[print-unity-mantissa=false]{1e-4} & 0.328 & 1.371 & 0.450 & Noise \\
    6 & 1.79 & 16 & \num[print-unity-mantissa=false]{1e-3} & 0.335 & 1.395 & 0.468 & Noise \\
    7 & 2.11 & 16 & \num{5e-4} & 0.338 & 1.298 & 0.438 & Noise \\
    8 & 2.23 & 16 & \num{5e-4} & 0.340 & 1.284 & 0.436 & Noise \\
    9 & 2.44 & 128 & \num[print-unity-mantissa=false]{1e-4} & 0.332 & 1.358 & 0.451 & Noise \\
    10 & 2.73 & 16 & \num{5e-4} & 0.336 & 1.320 & 0.444 & Noise \\
    11 & 2.77 & 128 & \num[print-unity-mantissa=false]{1e-3} & 0.348 & 1.931 & 0.672 & Spike \\
    12 & 2.78 & 16 & \num[print-unity-mantissa=false]{1e-3} & 0.342 & 1.321 & 0.451 & Noise \\
    13 & 2.84 & 16 & \num{5e-4} & 0.337 & 1.321 & 0.445 & Noise \\
    14 & 2.85 & 128 & \num[print-unity-mantissa=false]{1e-3} & 0.344 & 2.421 & 0.832 & Spike \\
    \bottomrule
  \end{tabular}
\end{table}

\subsubsection{Method effects against same-seed baselines}

PGD adversarial training reduces degradation on both single-target datasets and increases it on Traffic at all three seeds, while its ETTh1 effect remains near zero.
RevIN's mean deltas are positive for both $\mathcal{D}_w$ and $\mathrm{MSE}_w$ on every dataset and seed.
Fault augmentation consistently reduces degradation on Traffic, stays near zero on ETTh1, and changes sign on \BeijingTiantan{} and \PenmanshielWT{}.
Randomized training also reduces degradation on both single-target datasets at every seed, and ensemble aggregation reduces clean MSE on all four datasets.
Table~\ref{tab:training_seed_method_means} reports dataset-level means for every method, computed by taking each method-baseline delta within a seed before averaging over compatible architectures.

The mean fault-augmentation effect on \BeijingTiantan{} depends strongly on the PatchTST pair.
At seed 1001, PatchTST's degradation falls from $3.460$ to $2.445$, a delta of $-1.015$, while the other four architectures average $-0.018$.
At seed 1002, its delta is $0.161$.
Excluding PatchTST gives dataset means $-0.033$, $-0.018$, and $-0.029$ at seeds 42, 1001, and 1002, respectively.
The means over all five architectures remain $-0.039$, $-0.218$, and $0.009$.
Tables~\ref{tab:training_seed_pairs_beijing}--\ref{tab:training_seed_pairs_traffic} give every method-baseline pair underlying these means.

\begin{table}[!htb]
  \centering
  \small
  \setlength{\tabcolsep}{2.4pt}
  \renewcommand{\arraystretch}{1.10}
\caption{Dataset-level mean method-baseline deltas at each training and search seed.
  Each delta uses the same-seed baseline before averaging over five architectures, or four for RevIN, which excludes PatchTST.
  Negative values indicate improvement, with $\Delta\mathrm{MSE}_w$ comparing each pair's respective worst scenarios.
  PGD denotes PGD adversarial training, Rand. Train randomized training, Rand. Smooth randomized smoothing, ARL adaptive robust loss, and Fault Aug. fault augmentation.}
  \label{tab:training_seed_method_means}
  \begin{tabular*}{\textwidth}{@{\extracolsep{\fill}}l*{9}{S[table-format=-1.3]}@{}}
    \toprule
 & \multicolumn{3}{c}{Seed 42} & \multicolumn{3}{c}{Seed 1001} & \multicolumn{3}{c}{Seed 1002} \\
\cmidrule(lr){2-4}\cmidrule(lr){5-7}\cmidrule(lr){8-10}
Method & $\Delta\mathrm{MSE}_c$ & $\Delta\mathcal{D}_w$ & $\Delta\mathrm{MSE}_w$ & $\Delta\mathrm{MSE}_c$ & $\Delta\mathcal{D}_w$ & $\Delta\mathrm{MSE}_w$ & $\Delta\mathrm{MSE}_c$ & $\Delta\mathcal{D}_w$ & $\Delta\mathrm{MSE}_w$ \\
 \midrule
    \multicolumn{10}{@{}l}{\textit{\BeijingTiantan{}}} \\
    Rand. Train & 0.026 & -0.146 & -0.018 & 0.025 & -0.366 & -0.086 & 0.020 & -0.169 & -0.030 \\
    PGD & 0.004 & -0.136 & -0.038 & 0.002 & -0.341 & -0.107 & 0.001 & -0.170 & -0.052 \\
    Fault Aug. & 0.005 & -0.039 & -0.007 & -0.001 & -0.218 & -0.072 & 0.007 & 0.009 & 0.013 \\
    RevIN & 0.005 & 0.053 & 0.025 & 0.010 & 0.099 & 0.045 & 0.006 & 0.043 & 0.023 \\
    ARL & 0.001 & -0.024 & -0.005 & 0.016 & 0.357 & 0.166 & 0.005 & -0.027 & -0.002 \\
    Ensemble & -0.009 & 0.007 & -0.009 & -0.010 & -0.422 & -0.153 & -0.012 & 0.092 & 0.014 \\
    Rand. Smooth & 0.002 & 0.001 & 0.003 & 0.003 & -0.043 & -0.008 & 0.002 & -0.009 & 0.000 \\
    \addlinespace[3pt]
    \multicolumn{10}{@{}l}{\textit{\PenmanshielWT{}}} \\
    Rand. Train & -0.003 & -0.036 & -0.016 & 0.002 & -0.031 & -0.008 & 0.003 & -0.058 & -0.016 \\
    PGD & -0.005 & -0.059 & -0.026 & -0.006 & -0.033 & -0.018 & -0.003 & -0.103 & -0.039 \\
    Fault Aug. & -0.001 & -0.019 & -0.008 & 0.000 & 0.043 & 0.016 & 0.001 & -0.052 & -0.018 \\
    RevIN & -0.001 & 0.110 & 0.037 & -0.002 & 0.144 & 0.047 & 0.001 & 0.041 & 0.015 \\
    ARL & 0.040 & 0.144 & 0.102 & 0.045 & 0.124 & 0.100 & 0.047 & 0.024 & 0.066 \\
    Ensemble & -0.007 & -0.023 & -0.016 & -0.002 & -0.022 & -0.010 & -0.002 & -0.057 & -0.023 \\
    Rand. Smooth & 0.003 & -0.002 & 0.003 & 0.001 & 0.001 & 0.001 & 0.005 & 0.006 & 0.009 \\
    \addlinespace[3pt]
    \multicolumn{10}{@{}l}{\textit{ETTh1}} \\
    Rand. Train & 0.009 & -0.004 & 0.008 & 0.010 & 0.002 & 0.011 & 0.006 & -0.003 & 0.005 \\
    PGD & -0.005 & 0.000 & -0.005 & 0.000 & 0.002 & 0.000 & -0.011 & -0.001 & -0.012 \\
    Fault Aug. & 0.003 & -0.001 & 0.002 & 0.000 & -0.003 & -0.002 & 0.004 & -0.003 & 0.003 \\
    RevIN & -0.045 & 0.124 & 0.010 & -0.048 & 0.131 & 0.010 & -0.042 & 0.121 & 0.011 \\
    ARL & 0.033 & 0.001 & 0.037 & 0.020 & 0.000 & 0.023 & 0.033 & -0.008 & 0.032 \\
    Ensemble & -0.005 & -0.005 & -0.007 & -0.006 & -0.002 & -0.008 & -0.001 & -0.006 & -0.005 \\
    Rand. Smooth & -0.001 & 0.000 & -0.001 & 0.000 & 0.001 & 0.000 & 0.000 & -0.001 & 0.000 \\
    \addlinespace[3pt]
    \multicolumn{10}{@{}l}{\textit{Traffic}} \\
    Rand. Train & -0.007 & 0.011 & -0.003 & -0.001 & 0.012 & 0.011 & -0.004 & -0.002 & -0.001 \\
    PGD & -0.020 & 0.023 & -0.016 & -0.005 & 0.023 & 0.012 & -0.020 & 0.009 & -0.016 \\
    Fault Aug. & -0.013 & -0.025 & -0.026 & 0.003 & -0.009 & 0.003 & -0.010 & -0.043 & -0.030 \\
    RevIN & -0.063 & 0.267 & 0.071 & -0.050 & 0.291 & 0.109 & -0.063 & 0.274 & 0.077 \\
    ARL & 0.075 & -0.007 & 0.100 & 0.065 & -0.005 & 0.089 & 0.059 & -0.011 & 0.079 \\
    Ensemble & -0.017 & -0.019 & -0.037 & -0.013 & -0.018 & -0.023 & -0.020 & -0.023 & -0.040 \\
    Rand. Smooth & 0.011 & -0.021 & 0.008 & 0.011 & -0.023 & 0.006 & 0.008 & -0.017 & 0.005 \\
    \bottomrule
  \end{tabular*}
\end{table}

\begin{table}[!htb]
  \centering
  \footnotesize
  \setlength{\tabcolsep}{2.4pt}
  \renewcommand{\arraystretch}{1.06}
 \caption{Complete method-baseline deltas on \BeijingTiantan{} for seeds 42, 1001, and 1002.
  Each pair uses its same-seed baseline.
  Negative values improve, and $\Delta\mathrm{MSE}_w$ compares the pair's respective worst scenarios.
  RevIN excludes PatchTST.
  PGD denotes PGD adversarial training, Rand. Train randomized training, Rand. Smooth randomized smoothing, ARL adaptive robust loss, and Fault Aug. fault augmentation.}
  \label{tab:training_seed_pairs_beijing}
  \begin{tabular}{@{}ll*{9}{S[table-format=-1.3]}@{}}
    \toprule
 & & \multicolumn{3}{c}{Seed 42} & \multicolumn{3}{c}{Seed 1001} & \multicolumn{3}{c}{Seed 1002} \\
\cmidrule(lr){3-5}\cmidrule(lr){6-8}\cmidrule(lr){9-11}
Architecture & Method & $\Delta\mathrm{MSE}_c$ & $\Delta\mathcal{D}_w$ & $\Delta\mathrm{MSE}_w$ & $\Delta\mathrm{MSE}_c$ & $\Delta\mathcal{D}_w$ & $\Delta\mathrm{MSE}_w$ & $\Delta\mathrm{MSE}_c$ & $\Delta\mathcal{D}_w$ & $\Delta\mathrm{MSE}_w$ \\
     \midrule
    DLinear & Rand. Train & 0.003 & -0.323 & -0.103 & -0.003 & -0.309 & -0.105 & 0.002 & -0.325 & -0.105 \\
     & PGD & -0.001 & -0.335 & -0.111 & 0.000 & -0.319 & -0.105 & 0.000 & -0.333 & -0.109 \\
     & Fault Aug. & 0.005 & -0.070 & -0.015 & 0.002 & -0.023 & -0.004 & 0.003 & -0.034 & -0.006 \\
     & RevIN & 0.005 & -0.194 & -0.056 & 0.002 & 0.029 & 0.013 & 0.002 & 0.030 & 0.014 \\
     & ARL & -0.001 & -0.084 & -0.029 & -0.001 & -0.061 & -0.022 & -0.002 & -0.083 & -0.030 \\
     & Ensemble & -0.002 & -0.019 & -0.009 & -0.002 & -0.006 & -0.006 & -0.002 & -0.015 & -0.008 \\
     & Rand. Smooth & 0.004 & 0.000 & 0.006 & 0.005 & -0.009 & 0.006 & 0.004 & -0.008 & 0.004 \\
    \addlinespace[3pt]
    GRU & Rand. Train & 0.015 & -0.017 & 0.014 & 0.010 & -0.032 & 0.002 & 0.023 & -0.138 & -0.012 \\
     & PGD & 0.012 & -0.018 & 0.010 & 0.008 & -0.019 & 0.004 & 0.016 & -0.075 & -0.001 \\
     & Fault Aug. & 0.018 & 0.009 & 0.025 & -0.002 & 0.001 & -0.003 & 0.006 & -0.023 & 0.002 \\
     & RevIN & 0.021 & 0.234 & 0.097 & 0.019 & 0.233 & 0.094 & 0.025 & 0.129 & 0.072 \\
     & ARL & 0.061 & 0.020 & 0.084 & 0.062 & 0.022 & 0.085 & 0.063 & -0.047 & 0.066 \\
     & Ensemble & 0.009 & 0.008 & 0.013 & -0.006 & 0.006 & -0.006 & -0.007 & -0.001 & -0.009 \\
     & Rand. Smooth & 0.002 & 0.000 & 0.002 & 0.003 & 0.002 & 0.004 & 0.002 & -0.008 & 0.000 \\
    \addlinespace[3pt]
    ModernTCN & Rand. Train & 0.004 & -0.223 & -0.062 & 0.007 & -0.285 & -0.076 & -0.003 & -0.304 & -0.100 \\
     & PGD & 0.019 & -0.207 & -0.038 & 0.004 & -0.268 & -0.075 & -0.005 & -0.353 & -0.117 \\
     & Fault Aug. & 0.005 & -0.019 & 0.002 & -0.002 & 0.000 & -0.003 & 0.018 & -0.050 & 0.011 \\
     & RevIN & 0.019 & -0.060 & 0.010 & 0.017 & -0.028 & 0.018 & 0.008 & -0.218 & -0.058 \\
     & ARL & -0.006 & -0.100 & -0.038 & -0.002 & -0.016 & -0.009 & -0.009 & -0.120 & -0.051 \\
     & Ensemble & -0.005 & 0.016 & -0.002 & -0.008 & -0.036 & -0.023 & -0.021 & -0.098 & -0.061 \\
     & Rand. Smooth & 0.007 & -0.009 & 0.008 & 0.005 & -0.008 & 0.005 & 0.005 & -0.029 & -0.002 \\
    \addlinespace[3pt]
    TSMixer & Rand. Train & 0.105 & -0.038 & 0.103 & 0.088 & -0.062 & 0.079 & 0.067 & 0.019 & 0.086 \\
     & PGD & -0.013 & -0.005 & -0.017 & -0.008 & -0.016 & -0.015 & -0.006 & -0.005 & -0.009 \\
     & Fault Aug. & 0.000 & -0.049 & -0.018 & -0.012 & -0.050 & -0.031 & 0.011 & -0.010 & 0.010 \\
     & RevIN & -0.025 & 0.233 & 0.048 & 0.000 & 0.162 & 0.056 & -0.011 & 0.229 & 0.063 \\
     & ARL & -0.039 & 0.053 & -0.029 & -0.012 & 0.059 & 0.005 & -0.020 & 0.064 & -0.003 \\
     & Ensemble & -0.041 & 0.040 & -0.034 & -0.030 & -0.001 & -0.036 & -0.018 & 0.009 & -0.019 \\
     & Rand. Smooth & -0.005 & 0.008 & -0.003 & -0.005 & 0.003 & -0.005 & -0.004 & -0.001 & -0.005 \\
    \addlinespace[3pt]
    PatchTST & Rand. Train & 0.002 & -0.128 & -0.041 & 0.021 & -1.143 & -0.333 & 0.011 & -0.100 & -0.019 \\
     & PGD & 0.004 & -0.117 & -0.035 & 0.008 & -1.083 & -0.343 & 0.002 & -0.082 & -0.025 \\
     & Fault Aug. & -0.004 & -0.064 & -0.027 & 0.008 & -1.015 & -0.318 & -0.004 & 0.161 & 0.047 \\
     & ARL & -0.009 & -0.010 & -0.016 & 0.033 & 1.782 & 0.769 & -0.007 & 0.048 & 0.006 \\
     & Ensemble & -0.007 & -0.008 & -0.012 & -0.003 & -2.075 & -0.695 & -0.010 & 0.563 & 0.166 \\
     & Rand. Smooth & 0.000 & 0.006 & 0.002 & 0.005 & -0.202 & -0.052 & 0.001 & 0.003 & 0.002 \\
    \bottomrule
  \end{tabular}
\end{table}

\begin{table}[!htb]
  \centering
  \footnotesize
  \setlength{\tabcolsep}{2.4pt}
  \renewcommand{\arraystretch}{1.06}
  \caption{Complete method-baseline deltas on \PenmanshielWT{} for seeds 42, 1001, and 1002.
  Each pair uses its same-seed baseline.
  Negative values improve, and $\Delta\mathrm{MSE}_w$ compares the pair's respective worst scenarios.
  RevIN excludes PatchTST.
  PGD denotes PGD adversarial training, Rand. Train randomized training, Rand. Smooth randomized smoothing, ARL adaptive robust loss, and Fault Aug. fault augmentation.}
  \label{tab:training_seed_pairs_penmanshiel}
  \begin{tabular}{@{}ll*{9}{S[table-format=-1.3]}@{}}
    \toprule
 & & \multicolumn{3}{c}{Seed 42} & \multicolumn{3}{c}{Seed 1001} & \multicolumn{3}{c}{Seed 1002} \\
\cmidrule(lr){3-5}\cmidrule(lr){6-8}\cmidrule(lr){9-11}
Architecture & Method & $\Delta\mathrm{MSE}_c$ & $\Delta\mathcal{D}_w$ & $\Delta\mathrm{MSE}_w$ & $\Delta\mathrm{MSE}_c$ & $\Delta\mathcal{D}_w$ & $\Delta\mathrm{MSE}_w$ & $\Delta\mathrm{MSE}_c$ & $\Delta\mathcal{D}_w$ & $\Delta\mathrm{MSE}_w$ \\
    \midrule
     DLinear & Rand. Train & 0.001 & -0.048 & -0.015 & 0.001 & -0.002 & 0.001 & 0.002 & -0.004 & 0.001 \\
     & PGD & -0.002 & -0.037 & -0.015 & -0.001 & -0.005 & -0.003 & -0.001 & -0.007 & -0.004 \\
     & Fault Aug. & 0.000 & -0.005 & -0.002 & 0.000 & -0.007 & -0.002 & 0.000 & 0.043 & 0.014 \\
     & RevIN & 0.006 & 0.000 & 0.008 & 0.007 & 0.027 & 0.018 & 0.007 & 0.027 & 0.018 \\
     & ARL & 0.018 & 0.102 & 0.058 & 0.015 & 0.132 & 0.066 & 0.016 & 0.133 & 0.067 \\
     & Ensemble & -0.001 & -0.014 & -0.006 & -0.001 & 0.007 & 0.001 & -0.002 & 0.008 & 0.001 \\
     & Rand. Smooth & 0.002 & -0.003 & 0.002 & 0.002 & 0.000 & 0.002 & 0.001 & 0.001 & 0.002 \\
    \addlinespace[3pt]
    GRU & Rand. Train & -0.009 & -0.006 & -0.011 & 0.000 & -0.020 & -0.007 & 0.003 & 0.003 & 0.004 \\
     & PGD & -0.005 & -0.027 & -0.015 & -0.001 & -0.027 & -0.010 & -0.002 & -0.042 & -0.016 \\
     & Fault Aug. & -0.001 & 0.002 & 0.000 & 0.002 & 0.007 & 0.004 & -0.002 & 0.018 & 0.004 \\
     & RevIN & 0.001 & 0.381 & 0.133 & 0.004 & 0.402 & 0.142 & 0.007 & 0.351 & 0.130 \\
     & ARL & 0.062 & 0.066 & 0.092 & 0.068 & 0.030 & 0.086 & 0.111 & 0.219 & 0.218 \\
     & Ensemble & -0.008 & 0.001 & -0.008 & 0.001 & -0.012 & -0.003 & -0.004 & 0.003 & -0.004 \\
     & Rand. Smooth & 0.000 & 0.000 & 0.000 & 0.000 & 0.000 & 0.000 & 0.000 & 0.002 & 0.001 \\
    \addlinespace[3pt]
    ModernTCN & Rand. Train & -0.001 & -0.054 & -0.019 & 0.004 & -0.104 & -0.031 & -0.001 & -0.074 & -0.026 \\
     & PGD & 0.003 & -0.082 & -0.024 & -0.003 & -0.097 & -0.036 & -0.002 & -0.101 & -0.036 \\
     & Fault Aug. & 0.002 & -0.018 & -0.004 & 0.002 & -0.017 & -0.003 & 0.000 & -0.011 & -0.004 \\
     & RevIN & -0.005 & 0.032 & 0.005 & -0.004 & 0.010 & -0.001 & -0.002 & 0.007 & 0.000 \\
     & ARL & 0.085 & 0.070 & 0.129 & 0.089 & -0.022 & 0.099 & 0.094 & -0.040 & 0.097 \\
     & Ensemble & -0.005 & -0.006 & -0.007 & -0.003 & -0.043 & -0.018 & -0.004 & -0.029 & -0.014 \\
     & Rand. Smooth & 0.001 & -0.002 & 0.001 & 0.001 & 0.000 & 0.002 & 0.001 & 0.001 & 0.002 \\
    \addlinespace[3pt]
    TSMixer & Rand. Train & -0.001 & -0.074 & -0.027 & 0.003 & -0.003 & 0.002 & 0.007 & -0.195 & -0.060 \\
     & PGD & -0.017 & -0.146 & -0.070 & -0.022 & -0.019 & -0.032 & -0.017 & -0.331 & -0.135 \\
     & Fault Aug. & 0.003 & -0.092 & -0.029 & -0.009 & 0.027 & 0.000 & 0.002 & -0.309 & -0.106 \\
     & RevIN & -0.006 & 0.027 & 0.002 & -0.016 & 0.136 & 0.029 & -0.009 & -0.221 & -0.089 \\
     & ARL & 0.008 & -0.089 & -0.022 & -0.001 & 0.150 & 0.053 & 0.005 & -0.256 & -0.084 \\
     & Ensemble & -0.007 & -0.117 & -0.049 & -0.003 & -0.054 & -0.023 & -0.002 & -0.265 & -0.095 \\
     & Rand. Smooth & 0.007 & -0.006 & 0.006 & -0.002 & 0.008 & 0.001 & 0.019 & 0.027 & 0.037 \\
    \addlinespace[3pt]
    PatchTST & Rand. Train & -0.006 & 0.002 & -0.006 & 0.003 & -0.028 & -0.006 & 0.005 & -0.021 & -0.002 \\
     & PGD & -0.007 & 0.000 & -0.008 & -0.004 & -0.018 & -0.011 & 0.007 & -0.032 & -0.003 \\
     & Fault Aug. & -0.009 & 0.019 & -0.004 & 0.006 & 0.205 & 0.080 & 0.004 & 0.000 & 0.004 \\
     & ARL & 0.030 & 0.570 & 0.254 & 0.053 & 0.328 & 0.197 & 0.008 & 0.066 & 0.032 \\
     & Ensemble & -0.013 & 0.018 & -0.010 & -0.005 & -0.006 & -0.008 & 0.000 & -0.001 & -0.001 \\
     & Rand. Smooth & 0.003 & 0.003 & 0.005 & 0.003 & -0.002 & 0.002 & 0.002 & -0.001 & 0.003 \\
    \bottomrule
  \end{tabular}
\end{table}

\begin{table}[!htb]
  \centering
  \footnotesize
  \setlength{\tabcolsep}{2.4pt}
  \renewcommand{\arraystretch}{1.06}
  \caption{Complete method-baseline deltas on ETTh1 for seeds 42, 1001, and 1002.
  Each pair uses its same-seed baseline.
  Negative values improve, and $\Delta\mathrm{MSE}_w$ compares the pair's respective worst scenarios.
  RevIN excludes PatchTST.
  PGD denotes PGD adversarial training, Rand. Train randomized training, Rand. Smooth randomized smoothing, ARL adaptive robust loss, and Fault Aug. fault augmentation.}
  \label{tab:training_seed_pairs_etth1}
  \begin{tabular}{@{}ll*{9}{S[table-format=-1.3]}@{}}
    \toprule
 & & \multicolumn{3}{c}{Seed 42} & \multicolumn{3}{c}{Seed 1001} & \multicolumn{3}{c}{Seed 1002} \\
\cmidrule(lr){3-5}\cmidrule(lr){6-8}\cmidrule(lr){9-11}
Architecture & Method & $\Delta\mathrm{MSE}_c$ & $\Delta\mathcal{D}_w$ & $\Delta\mathrm{MSE}_w$ & $\Delta\mathrm{MSE}_c$ & $\Delta\mathcal{D}_w$ & $\Delta\mathrm{MSE}_w$ & $\Delta\mathrm{MSE}_c$ & $\Delta\mathcal{D}_w$ & $\Delta\mathrm{MSE}_w$ \\
    \midrule
    DLinear & Rand. Train & 0.003 & 0.006 & 0.007 & -0.003 & 0.006 & -0.001 & -0.001 & 0.002 & 0.000 \\
     & PGD & 0.003 & 0.002 & 0.005 & -0.001 & 0.003 & 0.000 & 0.003 & -0.001 & 0.003 \\
     & Fault Aug. & -0.001 & -0.006 & -0.003 & -0.001 & -0.003 & -0.002 & -0.001 & -0.002 & -0.002 \\
     & RevIN & 0.010 & 0.039 & 0.030 & 0.007 & 0.037 & 0.025 & 0.010 & 0.038 & 0.029 \\
     & ARL & 0.012 & 0.013 & 0.021 & 0.010 & 0.020 & 0.022 & 0.013 & 0.011 & 0.021 \\
     & Ensemble & 0.002 & 0.000 & 0.002 & -0.004 & 0.002 & -0.004 & 0.001 & -0.001 & 0.000 \\
     & Rand. Smooth & 0.000 & 0.000 & 0.000 & 0.001 & 0.000 & 0.001 & 0.000 & 0.000 & 0.000 \\
    \addlinespace[3pt]
    GRU & Rand. Train & 0.031 & -0.006 & 0.029 & 0.034 & -0.012 & 0.028 & 0.015 & -0.010 & 0.010 \\
     & PGD & -0.031 & 0.004 & -0.031 & 0.005 & -0.012 & -0.003 & -0.058 & -0.012 & -0.071 \\
     & Fault Aug. & 0.016 & -0.002 & 0.015 & 0.008 & -0.007 & 0.004 & 0.008 & -0.005 & 0.006 \\
     & RevIN & -0.141 & 0.209 & -0.040 & -0.156 & 0.210 & -0.059 & -0.139 & 0.188 & -0.055 \\
     & ARL & 0.093 & -0.004 & 0.096 & 0.019 & -0.016 & 0.009 & 0.081 & -0.049 & 0.053 \\
     & Ensemble & -0.010 & 0.004 & -0.008 & 0.000 & -0.007 & -0.005 & 0.013 & -0.018 & 0.003 \\
     & Rand. Smooth & 0.001 & 0.000 & 0.000 & 0.000 & 0.000 & 0.000 & 0.000 & 0.000 & 0.000 \\
    \addlinespace[3pt]
    ModernTCN & Rand. Train & 0.002 & 0.001 & 0.003 & 0.002 & 0.009 & 0.007 & 0.004 & 0.004 & 0.007 \\
     & PGD & 0.003 & 0.000 & 0.003 & -0.002 & 0.007 & 0.002 & -0.002 & 0.003 & 0.000 \\
     & Fault Aug. & 0.002 & -0.004 & 0.001 & -0.001 & 0.003 & 0.000 & 0.001 & 0.000 & 0.001 \\
     & RevIN & -0.018 & 0.092 & 0.020 & -0.011 & 0.095 & 0.030 & -0.014 & 0.095 & 0.025 \\
     & ARL & 0.050 & -0.026 & 0.048 & 0.044 & -0.031 & 0.038 & 0.048 & 0.004 & 0.061 \\
     & Ensemble & -0.004 & -0.014 & -0.012 & -0.005 & -0.012 & -0.012 & -0.007 & -0.001 & -0.009 \\
     & Rand. Smooth & 0.000 & 0.000 & 0.000 & 0.000 & 0.000 & 0.000 & 0.000 & 0.000 & 0.000 \\
    \addlinespace[3pt]
    TSMixer & Rand. Train & -0.001 & -0.004 & -0.003 & 0.018 & -0.012 & 0.016 & 0.008 & -0.003 & 0.008 \\
     & PGD & -0.007 & 0.002 & -0.007 & -0.012 & -0.013 & -0.020 & 0.000 & 0.004 & 0.002 \\
     & Fault Aug. & 0.001 & -0.003 & 0.000 & -0.007 & 0.009 & -0.004 & 0.007 & -0.001 & 0.007 \\
     & RevIN & -0.032 & 0.157 & 0.032 & -0.032 & 0.183 & 0.044 & -0.024 & 0.163 & 0.045 \\
     & ARL & -0.001 & 0.003 & 0.000 & 0.014 & -0.002 & 0.015 & 0.008 & -0.010 & 0.005 \\
     & Ensemble & -0.006 & -0.004 & -0.009 & -0.011 & -0.013 & -0.019 & -0.009 & -0.004 & -0.012 \\
     & Rand. Smooth & -0.004 & 0.002 & -0.004 & -0.003 & 0.001 & -0.003 & -0.002 & 0.000 & -0.002 \\
    \addlinespace[3pt]
    PatchTST & Rand. Train & 0.011 & -0.018 & 0.006 & -0.002 & 0.018 & 0.005 & 0.001 & -0.009 & -0.002 \\
     & PGD & 0.005 & -0.006 & 0.004 & 0.008 & 0.024 & 0.021 & 0.004 & -0.001 & 0.005 \\
     & Fault Aug. & -0.003 & 0.009 & 0.000 & 0.000 & -0.015 & -0.007 & 0.004 & -0.009 & 0.002 \\
     & ARL & 0.009 & 0.018 & 0.020 & 0.015 & 0.027 & 0.031 & 0.015 & 0.006 & 0.022 \\
     & Ensemble & -0.004 & -0.008 & -0.009 & -0.008 & 0.021 & -0.001 & -0.004 & -0.007 & -0.008 \\
     & Rand. Smooth & -0.001 & -0.004 & -0.003 & 0.001 & 0.006 & 0.003 & 0.001 & -0.004 & 0.000 \\
    \bottomrule
  \end{tabular}
\end{table}

\begin{table}[!htb]
  \centering
  \footnotesize
  \setlength{\tabcolsep}{2.4pt}
  \renewcommand{\arraystretch}{1.06}
 \caption{Complete method-baseline deltas on Traffic for seeds 42, 1001, and 1002.
  Each pair uses its same-seed baseline.
  Negative values improve, and $\Delta\mathrm{MSE}_w$ compares the pair's respective worst scenarios.
  RevIN excludes PatchTST.
  PGD denotes PGD adversarial training, Rand. Train randomized training, Rand. Smooth randomized smoothing, ARL adaptive robust loss, and Fault Aug. fault augmentation.}
  \label{tab:training_seed_pairs_traffic}
  \begin{tabular}{@{}ll*{9}{S[table-format=-1.3]}@{}}
    \toprule
 & & \multicolumn{3}{c}{Seed 42} & \multicolumn{3}{c}{Seed 1001} & \multicolumn{3}{c}{Seed 1002} \\
\cmidrule(lr){3-5}\cmidrule(lr){6-8}\cmidrule(lr){9-11}
Architecture & Method & $\Delta\mathrm{MSE}_c$ & $\Delta\mathcal{D}_w$ & $\Delta\mathrm{MSE}_w$ & $\Delta\mathrm{MSE}_c$ & $\Delta\mathcal{D}_w$ & $\Delta\mathrm{MSE}_w$ & $\Delta\mathrm{MSE}_c$ & $\Delta\mathcal{D}_w$ & $\Delta\mathrm{MSE}_w$ \\
    \midrule
 DLinear & Rand. Train & 0.001 & -0.002 & -0.001 & 0.001 & -0.002 & -0.001 & 0.001 & -0.002 & 0.000 \\
     & PGD & -0.001 & -0.002 & -0.002 & -0.001 & -0.002 & -0.002 & -0.001 & -0.002 & -0.002 \\
     & Fault Aug. & 0.000 & -0.001 & 0.000 & 0.000 & -0.001 & 0.000 & 0.000 & -0.001 & 0.000 \\
     & RevIN & -0.008 & 0.135 & 0.082 & -0.008 & 0.135 & 0.082 & -0.008 & 0.135 & 0.082 \\
     & ARL & 0.103 & 0.101 & 0.201 & 0.101 & 0.105 & 0.201 & 0.103 & 0.101 & 0.201 \\
     & Ensemble & -0.002 & -0.002 & -0.003 & -0.001 & -0.001 & -0.002 & -0.001 & -0.002 & -0.003 \\
     & Rand. Smooth & 0.000 & 0.000 & 0.000 & 0.000 & 0.000 & 0.000 & 0.000 & 0.000 & 0.000 \\
    \addlinespace[3pt]
    GRU & Rand. Train & -0.026 & 0.030 & -0.012 & -0.020 & 0.086 & 0.039 & -0.037 & 0.057 & -0.006 \\
     & PGD & -0.030 & -0.003 & -0.041 & -0.014 & 0.147 & 0.090 & -0.060 & 0.097 & -0.007 \\
     & Fault Aug. & -0.032 & 0.040 & -0.013 & 0.005 & 0.065 & 0.055 & -0.044 & 0.031 & -0.032 \\
     & RevIN & -0.088 & 0.247 & 0.045 & -0.080 & 0.392 & 0.164 & -0.103 & 0.268 & 0.047 \\
     & ARL & 0.049 & -0.050 & 0.024 & 0.084 & -0.026 & 0.078 & 0.053 & -0.054 & 0.022 \\
     & Ensemble & -0.022 & -0.097 & -0.097 & -0.040 & 0.013 & -0.038 & -0.044 & -0.045 & -0.086 \\
     & Rand. Smooth & -0.011 & 0.000 & -0.014 & -0.001 & -0.001 & -0.002 & -0.003 & -0.003 & -0.006 \\
    \addlinespace[3pt]
    ModernTCN & Rand. Train & 0.049 & -0.046 & 0.034 & 0.009 & 0.012 & 0.019 & 0.008 & -0.012 & 0.003 \\
     & PGD & 0.006 & -0.029 & -0.009 & 0.014 & 0.002 & 0.018 & 0.001 & -0.041 & -0.023 \\
     & Fault Aug. & 0.019 & -0.021 & 0.012 & 0.017 & -0.005 & 0.019 & 0.009 & -0.009 & 0.007 \\
     & RevIN & -0.058 & 0.338 & 0.102 & -0.077 & 0.373 & 0.092 & -0.060 & 0.338 & 0.097 \\
     & ARL & 0.093 & -0.028 & 0.100 & 0.002 & 0.016 & 0.011 & 0.003 & 0.005 & 0.007 \\
     & Ensemble & -0.003 & -0.009 & -0.009 & -0.009 & 0.011 & -0.005 & -0.009 & -0.018 & -0.022 \\
     & Rand. Smooth & 0.004 & -0.001 & 0.005 & 0.015 & -0.003 & 0.017 & 0.007 & 0.001 & 0.009 \\
    \addlinespace[3pt]
    TSMixer & Rand. Train & -0.054 & 0.040 & -0.047 & -0.002 & 0.013 & 0.005 & 0.000 & 0.035 & 0.021 \\
     & PGD & -0.061 & 0.083 & -0.031 & -0.010 & 0.058 & 0.019 & -0.040 & 0.075 & -0.009 \\
     & Fault Aug. & -0.052 & 0.030 & -0.050 & -0.001 & -0.018 & -0.012 & -0.017 & 0.005 & -0.019 \\
     & RevIN & -0.101 & 0.346 & 0.054 & -0.035 & 0.266 & 0.099 & -0.082 & 0.355 & 0.082 \\
     & ARL & 0.040 & 0.031 & 0.072 & 0.060 & 0.013 & 0.087 & 0.047 & 0.014 & 0.070 \\
     & Ensemble & -0.050 & -0.021 & -0.076 & -0.011 & -0.027 & -0.030 & -0.040 & -0.035 & -0.073 \\
     & Rand. Smooth & 0.003 & -0.002 & 0.002 & 0.002 & -0.003 & 0.001 & 0.002 & -0.002 & 0.002 \\
    \addlinespace[3pt]
    PatchTST & Rand. Train & -0.003 & 0.032 & 0.010 & 0.008 & -0.047 & -0.008 & 0.010 & -0.089 & -0.022 \\
     & PGD & -0.016 & 0.063 & 0.001 & -0.015 & -0.090 & -0.067 & 0.001 & -0.087 & -0.038 \\
     & Fault Aug. & 0.001 & -0.174 & -0.078 & -0.004 & -0.087 & -0.047 & 0.000 & -0.244 & -0.108 \\
     & ARL & 0.089 & -0.087 & 0.102 & 0.077 & -0.131 & 0.069 & 0.090 & -0.120 & 0.096 \\
     & Ensemble & -0.008 & 0.033 & 0.002 & -0.001 & -0.087 & -0.041 & -0.006 & -0.014 & -0.016 \\
     & Rand. Smooth & 0.059 & -0.104 & 0.046 & 0.038 & -0.107 & 0.016 & 0.032 & -0.080 & 0.020 \\
    \bottomrule
  \end{tabular}
\end{table}

\FloatBarrier

\endgroup

\subsection{Comparison with scenario means, corruption benchmarks, and effective robustness}
\label{app:metric_mapping}

We compare the benchmark's reported quantities with alternative scores from corruption benchmarks and studies of robustness to distribution shift.
As a direct ablation, we replace the maximum over scenarios by the scenario mean.
The corresponding mean-case degradation $\overline{\mathcal{D}}$ keeps the same normalized corruption notion as $\mathcal{D}_w$ and changes only the outer aggregation over scenarios.
It therefore answers the cleanest comparison question for this benchmark design: what changes if models are ranked by average degradation instead of by the worst benchmark scenario?
The detection benchmark of \citet{Michaelis.etal_2019} also averages over corruptions.
It ranks models by mean performance under corruption (mPC) and reports relative performance under corruption (rPC) and clean performance beside it.
rPC would invert the same mean-case ordering as $1/\overline{\mathcal{D}}$.
mPC is recoverable here as the raw mean corrupted MSE, $\mathrm{mPC}=\overline{\mathrm{MSE}}=\overline{\mathcal{D}}\,\mathrm{MSE}_c$.
The score families differ in how they aggregate errors across scenarios and what reference, if any, they compare those errors against (Table~\ref{tab:literature_metric_mapping}).
Certified-robustness evaluations ask a different question again: in randomized smoothing, certified accuracy at radius $r$ is the fraction of the test set classified correctly with a guarantee that remains valid within that radius \citep{cohenCertifiedAdversarialRobustness2019}.
That family therefore serves here as broader metric context rather than as a direct replacement for the scenario-based degradation benchmark.

\begin{table}[!htb]
  \centering
  \small
  \setlength{\tabcolsep}{3.5pt}
  \renewcommand{\arraystretch}{1.12}
  \caption{Uses and limitations of the reported quantities and alternative scores for evaluating forecasting under sensor faults.
  rPC and mPC denote relative and mean performance under corruption, and mCE denotes mean corruption error.}
  \label{tab:literature_metric_mapping}
  \begin{tabularx}{\columnwidth}{@{}
    >{\raggedright\arraybackslash}p{0.17\columnwidth}
    >{\raggedright\arraybackslash}p{0.26\columnwidth}
    >{\raggedright\arraybackslash}X
    >{\raggedright\arraybackslash}X
    @{}}
    \toprule
    Quantity & Definition & Most informative when & Limitation \\
    \midrule
    \multicolumn{4}{@{}l}{\textit{Reported quantities}} \\
    \addlinespace[0.15em]
    Clean error $\mathrm{MSE}_c$ & Clean forecasting error under intact sensing (see Definition~\ref{def:mc_mse_and_degradation_estimators}). & Comparing nominal forecasting accuracy. & Reflects no faults, which $\mathrm{MSE}_w$ and $\mathcal{D}_w$ add. \\
    \addlinespace[2pt]
    Worst-scenario fault-time error $\mathrm{MSE}_w$ & Absolute fault-time error at the model's own worst scenario (see Definition~\ref{def:worst_scenario_fault_time_mse}). & Comparing error magnitude under faults. & Combines clean error with fault sensitivity, which $\mathcal{D}_w$ separates. \\
    \addlinespace[2pt]
    Worst-scenario degradation $\mathcal{D}_w$ & Largest ratio of perturbed to clean MSE across scenarios (see Definition~\ref{def:worst_scenario_degradation}). & Comparing relative fault sensitivity when clean errors differ, including method-baseline comparisons. & Omits the absolute error level, which $\mathrm{MSE}_w$ reports. \\
    \addlinespace[3pt]
    \multicolumn{4}{@{}l}{\textit{Mean-case sensitivity}} \\
    \addlinespace[0.15em]
    Mean degradation $\overline{\mathcal{D}}$ and rPC~\citep{Michaelis.etal_2019} & Mean ratio of perturbed to clean MSE across scenarios (see Definition~\ref{def:mean_case_degradation}). & Comparing relative fault sensitivity when all scenarios are equally relevant. & Can hide a severe failure in one scenario. \\
    \addlinespace[2pt]
    mPC $=\overline{\mathrm{MSE}}$~\citep{Michaelis.etal_2019} & Mean corrupted MSE across scenarios (see Definition~\ref{def:mean_case_degradation}). & Comparing absolute error when all scenarios are equally relevant. & Combines clean error with fault sensitivity and can hide a severe failure in one scenario. \\
    \addlinespace[3pt]
    \multicolumn{4}{@{}l}{\textit{Audits with alternative scores}} \\
    \addlinespace[0.15em]
    mCE and relative mCE~\citep{hendrycksBenchmarkingNeuralNetwork2019} & Mean ratio of perturbed MSE to that of a fixed reference model across scenarios. & Comparing models with a reference across scenarios of different difficulty. & Depends on the reference and does not identify a model's own worst scenario. \\
    \addlinespace[2pt]
    Effective robustness $\overline{\rho}$~\citep{taoriMeasuringRobustnessNatural2020} & Amount by which mean corrupted MSE falls below a baseline frontier fitted on clean MSE. & Separating robustness from clean error across many architectures. & Fragile fit with few architectures per dataset. \\
    \addlinespace[2pt]
    Paired relative robustness $\overline{\tau}$~\citep{taoriMeasuringRobustnessNatural2020} & Reduction in mean corrupted MSE relative to the paired baseline forecasting model. & Testing the absolute benefit of a robustness-improvement method. & Does not separate gains in clean error from gains in robustness. \\
    \bottomrule
  \end{tabularx}
\end{table}

Among these alternatives, $\overline{\mathcal{D}}$ is the cleanest ablation because it leaves the normalization, evaluated scenario set, and loss definition unchanged.
Only the outer aggregation changes, from a maximum over scenarios to a mean over the evaluated scenario set.
This makes $\overline{\mathcal{D}}$ the direct test of whether worst-scenario aggregation materially changes the benchmark interpretation.
For that reason, $\overline{\mathcal{D}}$ is the first mean-case sensitivity quantity to report.
rPC is a monotone inverse of $\overline{\mathcal{D}}$, whereas mPC shares the scenario-mean aggregation but reports raw mean corrupted MSE.
Raw mean corrupted MSE remains useful because it exposes the absolute fault-time cost directly, but it can change rankings when clean MSE differs and therefore does not isolate the aggregation change as cleanly as $\overline{\mathcal{D}}$.

The corruption error (CE) family from ImageNet-C \citep{hendrycksBenchmarkingNeuralNetwork2019} addresses a different comparison question.
It divides a model's perturbed error in each scenario by that of a fixed reference model, here SeasonalNaive.
The mean of these ratios over scenarios is the mean corruption error (mCE).
In ImageNet-C, this division adjusts for the different difficulty of corruptions so that errors become comparable before they are averaged, and relative mCE standardizes the increase over clean error in the same way.
This is scientifically useful, especially because scores normalized by an anchor model can catch regressions that raise clean and perturbed error proportionally, which self-normalized scores may miss.
It is less suitable as the scenario-level degradation score, however, because the fixed reference model can help determine which scenario appears worst.
In exact score-family checks, the worst scenario under the CE family agrees with that of $\mathcal{D}_w$ on only a minority of rows, both on the short-horizon subset and on the four-dataset winner pool.
For that reason, mean degradation is the cleaner first sensitivity analysis, whereas mCE and relative mCE remain reference-normalized comparison scores rather than definitions of the worst benchmark scenario.
The benchmark protocol therefore reports worst-scenario fault-time error separately instead of using an anchor-normalized family to define the scenario identity.

Effective robustness in the sense of \citet{taoriMeasuringRobustnessNatural2020} provides an alternative score from the literature for architectures.
Taori's original construction is defined for standard and shifted accuracies with a fitted baseline frontier.
Taori et al. denote the quantities by $\rho$ and $\tau$ for a standard--shifted test-set pair.
We use overlines for the forecasting adaptation because it uses $\overline{\mathrm{MSE}}$, the scenario-mean corrupted MSE, rather than one shifted accuracy.
For each dataset, the adaptation fits a baseline frontier in log space from clean MSE to mean corrupted MSE over the baseline forecasting architectures with positive clean MSE and mean corrupted MSE.
Concretely, with $\overline{\beta}_d(x)=\exp(a_d+b_d\log x)$ for $x>0$, we define $\overline{\rho}(f)=\overline{\beta}_d(\mathrm{MSE}_c(f))-\overline{\mathrm{MSE}}(f)$.
This keeps the idea of controlling for clean performance with a fitted frontier while adapting it to positive unbounded errors instead of bounded accuracies.
The fitted score is still fragile because each dataset contributes only a handful of architectures with uneven coverage.
The fitted log-space relation remains uneven across datasets.
For method comparisons, the corresponding Taori quantity is relative robustness rather than effective robustness.
This follows Taori et al.'s motivation directly: a positive effective-robustness residual alone does not make an intervention useful, because both standard and shifted performance can still move in the wrong direction.
Each robustness-improved forecasting variant already comes with a canonical local reference, namely the paired baseline forecasting model on the same dataset and architecture.
In the mean case, the corresponding paired relative-robustness analogue is
\begin{equation*}
  \overline{\tau}(f',f_b)
  =
  \overline{\mathrm{MSE}}(f_b)-\overline{\mathrm{MSE}}(f').
\end{equation*}
Positive $\overline{\tau}$ indicates lower mean corrupted MSE than for the paired baseline forecasting model.
Equivalently, $\overline{\tau}=-\Delta\overline{\mathrm{MSE}}$, where $\Delta\overline{\mathrm{MSE}}=\overline{\mathrm{MSE}}(f')-\overline{\mathrm{MSE}}(f_b)$.
Negative $\Delta\mathcal{D}_w$ or $\Delta\overline{\mathcal{D}}$ and positive $\overline{\tau}$ therefore all indicate improvement relative to the paired baseline forecasting model, but on different scales.
The degradation deltas track changes in the self-normalized perturbed-to-clean error ratio, whereas $\overline{\tau}$ tracks raw mean corrupted MSE reduction.
For that reason, $\overline{\tau}$ is treated here as a mean-case audit rather than as a replacement for $\Delta\mathcal{D}_w$.
It plays the same role here that Taori et al.'s relative-robustness quantity plays next to effective robustness: a paired check on whether the intervention actually helps under shift, not only after clean-control adjustment.
It can disagree with worst-scenario degradation when a method lowers mean corrupted MSE yet leaves one exposed worst-case failure, or when clean-error shifts change the normalization in the degradation score without improving raw mean corrupted MSE.

This asymmetry between architectures and methods is deliberate.
A robustness score used in the architecture comparison in the baseline setting should be intrinsic to the evaluated model and should not depend on a chosen reference model or an underpowered cross-model fit.
A pairwise comparison of a robustness-improvement method, in contrast, already comes with a canonical reference, namely the paired baseline forecasting model on the same dataset and architecture.
Accordingly, architecture comparisons report $\mathcal{D}_w$ together with clean and fault-time error measures, whereas the reported method tables use $\Delta\mathcal{D}_w$, $\Delta\mathrm{MSE}_c$, and the worst-scenario fault-time delta $\Delta\mathrm{MSE}_w$.
The corresponding mean-case sensitivity and audits with alternative scores can therefore be read through $\Delta\overline{\mathcal{D}}$, $\Delta\overline{\mathrm{MSE}}$, and the paired relative-robustness analogue $\overline{\tau}$.

\begin{table*}[!tb]
  \centering
  \small
  \setlength{\tabcolsep}{2.8pt}
  \renewcommand{\arraystretch}{1.08}

  \newcommand{\BeijingTiantanRot}{%
    \begin{tabular}{@{}c@{}}
      Beijing Air\\
      Tiantan
    \end{tabular}%
  }

  \newcommand{\PenmanshielWTRot}{%
    \begin{tabular}{@{}c@{}}
      Penmanshiel\\
      WT08
    \end{tabular}%
  }

  \caption{Four-dataset score-family comparison for the architecture comparison in the baseline setting.
  This table follows the same dataset-by-measure grammar as Table~\ref{tab:main_results}, but replaces clean MSE and worst-scenario fault-time MSE with mean-case, anchor-normalized, and fitted clean-controlled alternative scores from the literature.
  Relative performance under corruption is omitted because it is a monotone transform of $\overline{\mathcal{D}}$ and therefore induces the same ordering.
  Mean corruption error and relative mean corruption error use the fixed SeasonalNaive reference on each dataset.
  Lower values are better except for $\overline{\rho}$, where higher is better.}
  \vspace{7pt}
  \label{tab:score_family_architecture_comparison}

  \begin{tabular*}{\textwidth}{@{\extracolsep{\fill}}
    >{\centering\arraybackslash}p{0.055\textwidth}
    l
    *{7}{S[
      table-format=-1.3,
      table-number-alignment=center,
      mode=text,
      detect-weight=true
    ]}
  @{}}
    \toprule
    Data & Measure
      & {\makecell{Seasonal\\Naive}} & {DLinear} & {GRU} & {ModernTCN}
      & {TSMixer} & {PatchTST} & {Chronos-2} \\
    \midrule

    \multirow{6}{*}{\rotatebox[origin=c]{90}{\footnotesize \BeijingTiantanRot}}
      & $\mathcal{D}_w$ & 1.353 & 1.628 & 1.251 & 1.502 & \bfseries 1.153 & 1.376 & 1.230 \\
      & $\overline{\mathcal{D}}$ & 1.145 & 1.206 & 1.125 & 1.186 & \bfseries 1.068 & 1.162 & 1.120 \\
      & {mPC} & 0.404 & 0.397 & \bfseries 0.316 & 0.357 & 0.383 & 0.392 & 0.391 \\
      & {mCE} & 1.000 & 0.978 & \bfseries 0.786 & 0.881 & 0.954 & 0.972 & 0.973 \\
      & {relative mCE} & 1.000 & 0.981 & \bfseries 0.785 & 0.884 & 0.949 & 0.973 & 0.972 \\
      & $\overline{\rho}$ & -0.005 & -0.021 & 0.013 & -0.008 & \bfseries 0.022 & -0.008 & 0.005 \\

    \addlinespace[4pt]
    \multirow{6}{*}{\rotatebox[origin=c]{90}{\footnotesize \PenmanshielWTRot}}
      & $\mathcal{D}_w$ & 1.320 & 1.259 & \bfseries 1.064 & 1.183 & 1.256 & 1.202 & 1.160 \\
      & $\overline{\mathcal{D}}$ & 1.107 & 1.099 & \bfseries 1.027 & 1.071 & 1.095 & 1.102 & 1.052 \\
      & {mPC} & 0.431 & 0.374 & \bfseries 0.355 & 0.360 & 0.386 & 0.390 & 0.422 \\
      & {mCE} & 1.000 & 0.870 & \bfseries 0.831 & 0.838 & 0.898 & 0.909 & 0.987 \\
      & {relative mCE} & 1.000 & 0.870 & \bfseries 0.826 & 0.836 & 0.897 & 0.909 & 0.983 \\
      & $\overline{\rho}$ & -0.011 & -0.007 & \bfseries 0.018 & 0.002 & -0.006 & -0.008 & 0.011 \\

    \addlinespace[4pt]
    \multirow{6}{*}{\rotatebox[origin=c]{90}{\footnotesize ETTh1}}
      & $\mathcal{D}_w$ & 1.288 & 1.251 & \bfseries 1.064 & 1.216 & 1.169 & 1.312 & 1.448 \\
      & $\overline{\mathcal{D}}$ & 1.148 & 1.119 & \bfseries 1.021 & 1.114 & 1.093 & 1.163 & 1.170 \\
      & {mPC} & 0.728 & \bfseries 0.490 & 0.679 & 0.514 & 0.519 & 0.501 & 0.580 \\
      & {mCE} & 1.000 & \bfseries 0.675 & 0.936 & 0.709 & 0.716 & 0.691 & 0.797 \\
      & {relative mCE} & 1.000 & \bfseries 0.674 & 0.926 & 0.707 & 0.713 & 0.692 & 0.798 \\
      & $\overline{\rho}$ & -0.040 & 0.009 & \bfseries 0.038 & 0.008 & 0.016 & -0.009 & -0.024 \\

    \addlinespace[4pt]
    \multirow{6}{*}{\rotatebox[origin=c]{90}{\footnotesize Traffic}}
      & $\mathcal{D}_w$ & 1.210 & \bfseries 1.177 & 1.284 & 1.265 & 1.290 & 1.676 & 1.911 \\
      & $\overline{\mathcal{D}}$ & 1.087 & 1.081 & \bfseries 1.071 & 1.111 & 1.113 & 1.296 & 1.293 \\
      & {mPC} & 1.381 & 0.740 & 0.778 & 0.639 & 0.702 & \bfseries 0.587 & 0.778 \\
      & {mCE} & 1.000 & 0.537 & 0.563 & 0.463 & 0.509 & \bfseries 0.423 & 0.561 \\
      & {relative mCE} & 1.000 & 0.537 & 0.562 & 0.464 & 0.510 & \bfseries 0.430 & 0.569 \\
      & $\overline{\rho}$ & -0.062 & 0.043 & \bfseries 0.045 & 0.037 & 0.028 & -0.034 & -0.076 \\
    \bottomrule
  \end{tabular*}
\end{table*}

\begin{table*}[!htb]
  \centering
  \small
  \setlength{\tabcolsep}{1.6pt}
  \renewcommand{\arraystretch}{1.08}

  \newcommand{\BeijingTiantanRot}{%
    \begin{tabular}{@{}c@{}}
      Beijing Air\\
      Tiantan
    \end{tabular}%
  }

  \newcommand{\PenmanshielWTRot}{%
    \begin{tabular}{@{}c@{}}
      Penmanshiel\\
      WT08
    \end{tabular}%
  }

  \caption{Four-dataset score-family comparison for the method-baseline comparison of robustness-improvement methods.
  This table follows the same dataset-by-measure grammar as Table~\ref{tab:main_results}, but replaces clean and worst-scenario fault-time deltas with mean-case, anchor-normalized, and paired relative-robustness alternatives.
  Values are dataset-level means over the available method-baseline pairs under the clean selector.
  Lower values are better except for $\overline{\tau}$, where higher is better.
  Fault Aug. denotes fault augmentation.}
  \vspace{7pt}
  \label{tab:score_family_method_comparison}

  \begin{tabular*}{\textwidth}{@{\extracolsep{\fill}}
    >{\centering\arraybackslash}p{0.055\textwidth}
    l
    *{7}{S[
      table-format=-1.3,
      table-number-alignment=center,
      mode=text,
      detect-weight=true
    ]}
  @{}}
    \toprule
    Data & Measure
      & {PGD} & {Ensemble}
      & {\makecell{Randomized\\Training}}
      & {\makecell{Randomized\\Smoothing}}
      & {\makecell{Adaptive\\Robust Loss}}
      & {RevIN}
      & {\makecell{Fault\\Aug.}} \\
    \midrule

    \multirow{5}{*}{\rotatebox[origin=c]{90}{\footnotesize \BeijingTiantanRot}}
      & $\Delta\!\mathcal{D}_w$              & -0.136 & 0.007 & \bfseries -0.146 & 0.001 & -0.024 & 0.053 & -0.039 \\
      & $\Delta\!\overline{\mathcal{D}}$     & -0.034 & 0.006 & \bfseries -0.042 & 0.000 & 0.003 & 0.025 & -0.011 \\
      & $\Delta$ mCE                         & -0.012 & \bfseries -0.020 & 0.038 & 0.005 & 0.008 & 0.038 & 0.005 \\
      & $\Delta$ relative mCE                & -0.014 & \bfseries -0.020 & 0.036 & 0.005 & 0.008 & 0.040 & 0.004 \\
      & $\overline{\tau}$                    & 0.006 & \bfseries 0.008 & -0.014 & -0.002 & -0.003 & -0.014 & -0.002 \\

    \addlinespace[4pt]
    \multirow{5}{*}{\rotatebox[origin=c]{90}{\footnotesize \PenmanshielWTRot}}
      & $\Delta\!\mathcal{D}_w$              & \bfseries -0.059 & -0.023 & -0.036 & -0.002 & 0.144 & 0.110 & -0.019 \\
      & $\Delta\!\overline{\mathcal{D}}$     & \bfseries -0.016 & -0.007 & -0.012 & -0.001 & 0.015 & 0.043 & -0.007 \\
      & $\Delta$ mCE                         & \bfseries -0.025 & -0.022 & -0.016 & 0.006 & 0.112 & 0.033 & -0.008 \\
      & $\Delta$ relative mCE                & \bfseries -0.026 & -0.022 & -0.017 & 0.006 & 0.113 & 0.037 & -0.008 \\
      & $\overline{\tau}$                    & \bfseries 0.012 & 0.010 & 0.007 & -0.003 & -0.049 & -0.014 & 0.004 \\

    \addlinespace[4pt]
    \multirow{5}{*}{\rotatebox[origin=c]{90}{\footnotesize ETTh1}}
      & $\Delta\!\mathcal{D}_w$              & 0.000 & \bfseries -0.005 & -0.004 & 0.000 & 0.001 & 0.124 & -0.001 \\
      & $\Delta\!\overline{\mathcal{D}}$     & -0.004 & -0.004 & -0.002 & 0.002 & \bfseries -0.011 & 0.045 & -0.002 \\
      & $\Delta$ mCE                         & -0.010 & -0.010 & 0.012 & 0.000 & 0.039 & \bfseries -0.036 & 0.003 \\
      & $\Delta$ relative mCE                & -0.010 & -0.010 & 0.012 & 0.000 & 0.038 & \bfseries -0.033 & 0.002 \\
      & $\overline{\tau}$                    & 0.007 & 0.007 & -0.009 & 0.000 & -0.028 & \bfseries 0.026 & -0.002 \\

    \addlinespace[4pt]
    \multirow{5}{*}{\rotatebox[origin=c]{90}{\footnotesize Traffic}}
      & $\Delta\!\mathcal{D}_w$              & 0.023 & -0.019 & 0.011 & -0.021 & -0.007 & 0.267 & \bfseries -0.025 \\
      & $\Delta\!\overline{\mathcal{D}}$     & -0.002 & -0.004 & 0.000 & -0.009 & 0.005 & 0.102 & \bfseries -0.011 \\
      & $\Delta$ mCE                         & \bfseries -0.018 & -0.015 & -0.006 & 0.008 & 0.066 & -0.010 & -0.013 \\
      & $\Delta$ relative mCE                & \bfseries -0.018 & -0.016 & -0.006 & 0.007 & 0.067 & -0.006 & -0.013 \\
      & $\overline{\tau}$                    & \bfseries 0.024 & 0.022 & 0.008 & -0.010 & -0.091 & 0.011 & 0.018 \\
    \bottomrule
  \end{tabular*}
\end{table*}

Tables~\ref{tab:score_family_architecture_comparison} and~\ref{tab:score_family_method_comparison} provide the full four-dataset score-family surfaces for the models of Table~\ref{tab:main_results}, selected on clean validation data at one training and search seed.
On the architecture side, $\mathcal{D}_w$ and $\overline{\mathcal{D}}$ agree on three of the four datasets and differ only on Traffic, whereas mPC and the CE family shift \BeijingTiantan{}, ETTh1, and Traffic toward architectures with lower raw or reference-normalized mean corrupted MSE.
The fitted effective-robustness score $\overline{\rho}$ aligns most closely with mean degradation, matching the $\overline{\mathcal{D}}$ leader on all four datasets.
On the method side, the worst-versus-mean degradation swap changes only the ETTh1 winner: ensemble is lowest under $\Delta\mathcal{D}_w$, whereas adaptive robust loss is lowest under $\Delta\overline{\mathcal{D}}$.
The larger disagreements come from $\overline{\tau}$ and the CE-family deltas, which reward mean corrupted MSE reduction and therefore favor ensemble on \BeijingTiantan{}, RevIN on ETTh1, and PGD adversarial training on Traffic even when worst-scenario self-normalized degradation points elsewhere.
Thus, the direct aggregation ablation largely preserves the degradation-based interpretation, while the broader audits with alternative scores identify which conclusions change when the evaluation question changes from worst-scenario sensitivity to average or reference-normalized corrupted error.

Prior work motivates worst-case evaluation under shift and related uniform-performance or distributionally robust viewpoints \citep{liEvaluatingModelPerformance2021, Subbaswamy.etal_2021, Duchi.Namkoong_2021, Dziugaite.etal_2020}.
Together with the score-family comparisons above, this supports using worst-scenario degradation as the relative-robustness score for architecture comparisons in this benchmark.
Worst-scenario degradation asks whether any scored sensor-fault scenario substantially increases error relative to nominal performance.
Mean-case quantities such as $\overline{\mathcal{D}}$ and mPC are useful sensitivity views, but they change the question to average degradation or average corrupted error and can hide concentrated failures.
\citet{hendrycksBenchmarkingNeuralNetwork2019} define corruption robustness as average-case performance over a corruption distribution meant to approximate how often each corruption occurs in practice.
The benchmark does not estimate how often each scenario occurs in deployment, so a mean over scenarios would weight all scenarios equally by assumption, whereas the maximum needs no such weights.
The remaining alternative scores additionally depend on an external anchor or a small fitted baseline frontier.

\clearpage
\section{Extended related work and taxonomy of robustness-improvement methods}
\label{sec:taxonomy}

The taxonomy has two roles: it states how the evaluated method set covers distinct intervention cells, and it grounds that non-exhaustive set by showing where adjacent mechanisms would enter the same benchmark protocol.

\subsection{Robustness-improvement taxonomy}
\label{subsec:robustness_improvement_taxonomy}

The benchmarked robustness-improvement methods differ in loss design, perturbation model, wrapper logic, timing, and required access to the forecaster.
These differences determine which comparisons are meaningful: retraining a model with adversarial inner loops is not interchangeable with wrapping a frozen predictor, even when both are evaluated under the same sensor-fault protocol.
The taxonomy organizes these distinctions along three operational axes: stage of intervention, model access, and target of intervention.
Table~\ref{tab:xai_axes_robustness} defines the axes, and Table~\ref{tab:implemented_method_taxonomy} assigns the evaluated robustness-improvement methods.

\begin{table}[!htb]
  \centering
  \small
  \setlength{\tabcolsep}{4pt}
  \renewcommand{\arraystretch}{1.20}
  \caption{Operational axes for organizing robustness-improvement methods for forecasting under sensor faults.}
  \label{tab:xai_axes_robustness}
    \begin{tabularx}{\columnwidth}{@{}
    >{\RaggedRight\arraybackslash}p{0.22\columnwidth}
    X
    @{}}
    \toprule
    Taxonomy element & Description \\
    \midrule
    \multicolumn{2}{@{}l}{\textit{Stage of intervention}} \\
    \addlinespace[0.15em]
    Ante-hoc & Robustness is built into preprocessing, model design, or training before deployment. \\
    Post-hoc & Robustness is added after the base predictor has been trained. \\
    \addlinespace[3pt]
    \multicolumn{2}{@{}l}{\textit{Model access}} \\
    \addlinespace[0.15em]
    Black-box & The method uses query access to a trained predictor and wraps inputs or outputs without retraining or modifying that predictor. \\
    Grey-box & The method uses portable training or model-composition hooks without relying on hidden states, architecture-specific modules, or adversarial gradients of the current model. \\
    White-box & The method uses gradients, hidden states, architecture-specific modules, or other model-specific internals of the current model. \\
    \addlinespace[3pt]
    \multicolumn{2}{@{}l}{\textit{Target of intervention}} \\
    \addlinespace[0.15em]
    Input level & The method acts on observed inputs, repaired signals, derived features, or derived training examples. \\
    Model level & The method acts on the forecasting model itself, including its architecture, representation, normalization behavior, or optimization objective. \\
    Output level & The method acts around the deployed model at the level of predictions, predictive distributions, prediction sets, or downstream decisions. \\
    \bottomrule
  \end{tabularx}
\end{table}

\subsection{Benchmarked robustness-improvement methods within the taxonomy}
\label{subsec:implemented_method_taxonomy}

Table~\ref{tab:implemented_method_taxonomy} places the seven evaluated robustness-improvement methods within the taxonomy.
Assignments refer to the evaluated implementation or variant.
The methods span four cells of the taxonomy: ante-hoc and grey-box at the input level, ante-hoc and grey-box at the model level, ante-hoc and white-box at the model level, and post-hoc and black-box at the output level.
The method-level summaries average paired rows over DLinear, GRU, ModernTCN, PatchTST, and TSMixer.
SeasonalNaive and Chronos-2 enter only the architecture comparison.
RevIN has four paired architectures because PatchTST already applies per-window instance normalization and de-normalization.

\begin{table}[!htb]
  \centering
  \small
  \setlength{\tabcolsep}{4pt}
  \renewcommand{\arraystretch}{1.14}
  \caption{Assignment of the seven robustness-improvement methods evaluated in the benchmark to the taxonomy. Assignments refer to the evaluated implementation. The source column names method-origin papers where applicable, while fault augmentation is from this work.}
  \label{tab:implemented_method_taxonomy}
  \begin{tabular}{@{}lllll@{}}
    \toprule
    Method & Source & Stage & Access & Target \\
    \midrule
    Randomized training & \citet{yoonRobustProbabilisticTime2022} & Ante-hoc & Grey-box & Input level \\
    Fault augmentation & This work & Ante-hoc & Grey-box & Input level \\
    RevIN & \citet{Kim.etal_2022} & Ante-hoc & Grey-box & Model level \\
    Adaptive robust loss & \citet{barronGeneralAdaptiveRobust2019} & Ante-hoc & Grey-box & Model level \\
    PGD adversarial training & \citet{madry2018towards} & Ante-hoc & White-box & Model level \\
    Ensemble aggregation & \citet{barrowEvaluationNeuralNetwork2010} & Post-hoc & Black-box & Output level \\
    Randomized smoothing & \citet{rekavandiCertifiedAdversarialRobustness2024} & Post-hoc & Black-box & Output level \\
    \bottomrule
  \end{tabular}
\end{table}

The access and target assignments follow the operational definitions in Table~\ref{tab:xai_axes_robustness}.
Randomized training and fault augmentation are input-level because they alter the training input distribution.
Fault augmentation uses the disjoint transfer suite in Table~\ref{tab:holdout_suite}.
Adaptive robust loss, RevIN, and PGD adversarial training are model-level because they change the loss, predictor computation path, or training objective.
Ensemble aggregation and randomized smoothing are post-hoc, black-box methods at the output level because they operate on predictions from fixed base predictors.

\subsection{Benchmarks and stress tests}
\label{subsec:benchmark_stress_predecessors}

Forecasting benchmarks make model comparisons reusable by fixing datasets, splits, metrics, and baselines.
The Monash Forecasting Repository provides a multi-dataset forecasting archive with baseline results \citep{godahewaMonashTimeSeries2021}, and TFB broadens benchmark coverage across domains, methods, forecasting strategies, and metrics \citep{qiuTFBComprehensiveFair2024}.
Long-horizon re-evaluations show that baseline choice can change empirical conclusions \citep{zengAreTransformersEffective2023}, and forecast-evaluation surveys emphasize split design, scaling, and metric choice \citep{hewamalageForecastEvaluationData2023}.
These works provide nominal forecasting-comparison precedent, but not injected sensor-fault scenarios, transfer-fault splits, or separate degradation-oriented measures.

Adjacent robustness and stress-test studies cover single stress or defense settings.
Work on adversarial forecasting studies adversarial input perturbations \citep{liu2023robust}, work on robust probabilistic forecasting covers randomized smoothing and randomized training \citep{yoonRobustProbabilisticTime2022}, and RobustTSF studies learning from training series contaminated by anomalies and robust objectives \citep{chengRobustTSFTheoryDesign2024}.
Pretraining that accounts for sensor failures targets representation learning for time series of vehicle dynamics with sensor faults \citep{brandtFaultsFeaturesPretraining2025}.
TSRBench evaluates spikes and level shifts with controllable severity \citep{kimLocalGeometryAttention2026}, Variable Subset Forecast studies unavailable variables at inference \citep{chauhanMultiVariateTimeSeries2022}, and a synthetic testbed for multivariate long-horizon forecasting controls signal components and noise processes \citep{janssenBenchmarkingMLTSFFrequency2026}.

Industrial and CPS studies add data-quality and sensor-disturbance precedents.
Prior work has studied stress tests with data-quality perturbations for industrial time-series classification \citep{Dix.etal_2023} and severity-controlled disturbance benchmarks for CPS forecasting architectures \citep{windmannQuantifyingRobustnessBenchmarking2025}.
These studies motivate the positioning axes in Table~\ref{tab:benchmark_positioning}: nominal benchmarks standardize forecasting comparison, robustness and stress-test studies define stress or defense settings, and industrial and CPS studies provide the context of stress tests on data quality and sensor disturbances.
SensorFault-Bench combines these roles through sensor faults grounded in CPS practice, a broad comparison of architectures, paired evaluation of robustness-improvement methods, disjoint transfer faults used only in training, and separate measures of clean error, fault-time error, and degradation.

\subsection{Relation to existing taxonomies}
\label{subsec:assessment_taxonomy}

Existing robustness, time-series, and resilience taxonomies motivate lifecycle and pipeline views.
Robust AI work has been organized by pipeline phase, architecture, task, system, and assessment methodology \citep{tocchettiAIRobustnessHumanCentered2025}.
Time-series out-of-distribution work organizes the area around data distribution, representation learning, and evaluation \citep{Wu.etal_2026}.
System-resilience work distinguishes threat sources, resilience factors, and activities such as preparation, disturbance absorption, recovery, and adaptation \citep{moskalenkoResilienceResilientSystems2023}.
For method comparison under a shared sensor-fault benchmark, these lifecycle views leave implementation access, retraining requirements, architectural dependence, and deployment portability to be specified separately.

Assessment-oriented frameworks describe how robustness or data quality is evaluated rather than how a forecasting method is altered.
The ISO/IEC 24029 series separates overview guidance, formal-method methodology, and statistical-method methodology for robustness assessment \citep{ISOIEC2021_TR24029_1,ISOIEC2023_24029_2,ISOIEC2026_DIS24029_3}.
ISO/IEC 5259 covers overview, terminology, examples, data-quality models, measurable characteristics, and data-quality reporting for analytics and machine learning \citep{ISOIEC2024_5259_1,ISOIEC2024_5259_2}.
Local robustness properties concern a specific sample, forecast window, or trajectory segment, while global robustness concerns a broader input space or, in probabilistic relaxations, a data distribution \citep{Leino2021_ICML_GloballyRobustNN,Blohm2025_ICML_PAGR}.
Even local robustness comparisons can depend on the perturbation parameter and the summary of probability-perturbation curves \citep{BuSun2023_ICANN_LocalRobustnessComparison}.

Studies of industrial data quality and sensor data quality document missing data, noisy measurements, device faults, synchronization differences, outliers, drift, and related provenance problems \citep{hubauerAnalysisDataQuality2013,jesusSurveyDataQuality2017,tehSensorDataQuality2020}.
Adjacent studies of industrial time series classification and CPS forecasting use stress tests based on perturbations and aggregate degradation at the dataset level \citep{Dix.etal_2023,windmannQuantifyingRobustnessBenchmarking2025}.
Adversarial robustness studies construct perturbations under explicit threat models \citep{madry2018towards,Blohm2025_ICML_PAGR}, whereas corruption-oriented robustness evaluates degradation under fixed, non-adversarial corruption families and severity levels \citep{hendrycksBenchmarkingNeuralNetwork2019}.
For industrial forecasting under sensor faults, measurement imperfections, synchronization effects, device faults, and drift align with the corruption-oriented setting because they are recurring non-adversarial input-side disruptions.
These assessment dimensions annotate the disturbance model, evidence basis, and claim scope.
They do not identify whether a robustness-improvement method changes inputs, the trained model, or the prediction interface.

Explainable AI (XAI) provides adjacent vocabulary for naming the operational distinctions in Table~\ref{tab:xai_axes_robustness}.
XAI surveys organize heterogeneous methods by pipeline location \citep{Ali.etal_2023}, ante-hoc versus post-hoc timing \citep{Retzlaff.etal_2024}, and model-specific versus model-agnostic access, portability, and local versus global scope \citep{Schwalbe.Finzel_2024,Mersha.etal_2024}.
Recent XAI work also studies explanations as tools for model debugging and model improvement rather than only retrospective inspection \citep{Weber.etal_2023}.
Explanation methods and robustness-improvement methods have different objectives, so the shared distinctions concern only how a method enters a prediction pipeline: the stage of intervention, the required model access, and the target of intervention at the input, model, or output level.
Local versus global remains a property of the robustness claim and evaluation protocol rather than a separate intervention axis.

\subsection{Literature-grounded coverage and extension map}
\label{subsec:broader_lit_map}

This coverage map grounds the taxonomy by placing the evaluated method set next to adjacent mechanisms from forecasting and related time-series work.
It is not an exhaustive survey.
Its purpose is to show that the operational axes in Table~\ref{tab:xai_axes_robustness} describe real method choices and to mark which unevaluated mechanisms and cells are natural extensions of the benchmark without implying that the evaluated set exhausts them.
Assessment, diagnostics, monitoring, and engineering processes describe how robustness evidence is produced, reported, or used to trigger later action.
Because those activities do not alter a forecaster, they remain outside this coverage map.
Operational forecasting systems can also combine cells.
One production neural forecasting system uses periodic retraining, native missing-value handling, and anomaly filtering, spanning model updates and input-side handling \citep{Bohlke-Schneider.etal_2020}.

\paragraph{Ante-hoc input level.}
Input-level methods alter the signal or training distribution before the fitted predictor is fixed.
This cell contains the evaluated randomized-training and fault-augmentation families.
Adjacent mechanisms include soft sensors that recover unavailable variables of a wind sensor board \citep{Kumar.etal_2023}, generative imputation for missing values in multivariate time series \citep{luoE2GANEndtoEndGenerative2019}, and RobustSTL decomposition under anomalies and abrupt changes \citep{wenRobustSTLRobustSeasonalTrend2019}.
Broader surveys of time series augmentation cover transformations in time, in frequency, by decomposition, and by generative models, while emphasizing that the validity of augmentation depends on task and data \citep{wenTimeSeriesData2021,iglesiasDataAugmentationTechniques2023}.
Randomized perturbation training in probabilistic forecasting is a narrower training-distribution intervention in the same cell \citep{yoonRobustProbabilisticTime2022}.

\paragraph{Ante-hoc model level: objectives and training.}
Interventions on the objective and on training change how the predictor is fitted.
The evaluated method set covers this cell through adaptive robust loss and PGD adversarial training.
Related mechanisms include adaptive robust losses for generic regression objectives \citep{barronGeneralAdaptiveRobust2019}, robust forecasting designs for training data contaminated by anomalies \citep{chengRobustTSFTheoryDesign2024}, and sparse additive forecasting models with Huber-type losses for heavy-tailed or nonstationary time series \citep{Wang.etal_2022}.
Adversarial perturbation training optimizes a min-max objective against the current model \citep{madry2018towards}.
Variants for forecasting study perturbations that seek high loss and defenses for multivariate probabilistic forecasting \citep{liu2023robust}, while adversarial training with transferred perturbations provides context on source models outside forecasting \citep{Tramer.etal_2018}.
Attack papers mainly delimit threat-model boundaries rather than add separate intervention cells.
Examples include attacks on solar forecasts in white-box and transferred black-box settings \citep{Tang.etal_2021}, targeted attacks on wind power forecasting \citep{Heinrich.etal_2024}, and white-box attacks on time series forecasting that preserve temporal characteristics \citep{shenTemporalCharacteristicsbasedAdversarial2025}.

\paragraph{Ante-hoc model level: architecture and representation.}
Model-level interventions can also change the predictor computation path or representation.
The evaluated method set includes RevIN as a normalization module inside the forecaster \citep{Kim.etal_2022}.
Unevaluated extension candidates include observation-aware architectures for missingness or irregular timing.
Heterogeneous graph-attention recurrent models avoid a separate imputation stage for sensor-network series with missing windows and unequal sampling rates \citep{DeBarros.etal_2023}.
ChannelTokenFormer uses dynamic patching and mask-guided attention for asynchronous and partially missing channels \citep{Jang.etal_2026}, while RFN models irregular and partially observed multivariate inputs probabilistically \citep{Li.etal_2025}.
Other examples at the architecture level target adjacent mechanisms of robustness or generalization: ANA-LSTM adds recurrent structure against noise for inputs to battery prognosis \citep{Wang.etal_2023}, modular forecasters encode assumed dependencies between states and controls for out-of-distribution generalization \citep{Bansal.etal_2021b}, ReTimeCausal combines imputation over irregular time with learning of lagged causal graphs \citep{Li.etal_2026}, and DeepRRTime regularizes a forecaster with implicit representations for missing lookback values \citep{Sastry.etal_2025}.
Approaches based on representation learning include pretraining aimed at sensor failures \citep{brandtFaultsFeaturesPretraining2025}, multi-view learning under varying missing rates \citep{yuMerlinMultiViewRepresentation2025}, forecasting with domain adaptation through shared attention representations \citep{Jin.etal_2022}, and universal forecasting in the style of foundation models with transfer across datasets \citep{Darlow.etal_2024}.

\paragraph{Post-hoc input level.}
Post-hoc input-level interventions preserve the trained forecaster but alter or recover inputs at inference.
Variable-subset forecasting wraps an existing multivariate forecaster when only some variables are available \citep{chauhanMultiVariateTimeSeries2022}.
ROBUSTTS instead performs test-time recovery or denoising before downstream time-series recovery and classification \citep{Jeon.etal_2023}.
These methods are extension candidates for the benchmark when the repaired input stream can be paired with the same no-intervention forecaster and scored under the same scenario set.

\paragraph{Post-hoc model level.}
Post-hoc model-level methods update, recalibrate, adapt, or replace model behavior after an initial fit.
TAFAS adapts forecasting models at test time with scheduling and calibration modules while preserving the source forecaster \citep{kimBattlingNonstationarityTime2025}.
Other examples include drift-triggered retraining of selected regressors \citep{Pawar.etal_2025}, closed-loop drift detection and model updates in industrial CPS \citep{Jayaratne.etal_2021}, and adaptive soft-sensor ensembles over nonstationary process streams \citep{Bakirov.etal_2017}.
These methods lie outside the evaluated comparison, which selects on clean data and keeps weights fixed, but fit the taxonomy as benchmark extensions when their adaptation interface is fixed before scoring.

\paragraph{Post-hoc output level.}
Output-level interventions act through predictions rather than through the fitted model or input stream.
The evaluated method set covers this cell through ensemble aggregation and randomized smoothing.
Forecast ensembles combine fitted models through selected-member mean or median schemes \citep{barrowEvaluationNeuralNetwork2010}, and related ensemble work compares aggregation operators such as mean, median, and mode across initialization and bagging variants \citep{kourentzesNeuralNetworkEnsemble2014}.
Randomized smoothing builds a smoothed predictor by aggregating noisy queries around a fixed model in probabilistic forecasting \citep{yoonRobustProbabilisticTime2022}.
For regression, alpha-trimmed aggregation gives a black-box randomized-smoothing wrapper around a fixed regressor \citep{rekavandiCertifiedAdversarialRobustness2024}.
Adjacent output-interface mechanisms include adaptive conformal intervals for dependent time series \citep{Zaffran.etal_2022}, conformalized randomized smoothing for time-series classification \citep{francoGuaranteeingRobustnessRealWorld2024}, and uncertainty-aware abstention, handoff, or fallback decisions around trained networks \citep{gawlikowskiSurveyUncertaintyDeep2023}.

The operational taxonomy structures the controlled method-baseline comparison for the evaluated method set and identifies where methods specialized for sensor faults or missing data, adaptation methods, and wrappers can enter future benchmark extensions.

\end{document}